\documentclass[10pt,journal,compsoc]{IEEEtran}

\usepackage[pagebackref,breaklinks,colorlinks]{hyperref}
\usepackage{graphicx}
\usepackage{amsmath}
\usepackage{amssymb}
\usepackage{multirow}
\usepackage{booktabs}
\usepackage[normalem]{ulem}
\usepackage{color, colortbl}
\definecolor{Gray}{gray}{0.9}

\ifCLASSOPTIONcompsoc
  \usepackage[nocompress]{cite}
\else
  \usepackage{cite}
\fi

\ifCLASSOPTIONcompsoc
  \usepackage[caption=false,font=footnotesize,labelfont=sf,textfont=sf]{subfig}
\else
  \usepackage[caption=false,font=footnotesize]{subfig}
\fi

\begin{document}

\title{Learning Optical Flow and Scene Flow with Bidirectional Camera-LiDAR Fusion}

\author{Haisong~Liu,
        Tao~Lu,
        Yihui~Xu,
        Jia~Liu, \textit{Member,~IEEE}
        and Limin~Wang, \textit{Member,~IEEE}
\IEEEcompsocitemizethanks{
\IEEEcompsocthanksitem Haisong Liu, Tao Lu, Yihui Xu, Jia Liu, and Limin Wang are with
the State Key Laboratory for Novel Software Technology, Nanjing
University, Nanjing, China, 210023 (E-mail: \{liuhs, taolu, yhxu\}@smail.nju.edu.cn;
\{jialiu, lmwang\}@nju.edu.cn) (Corresponding author: Limin Wang.)
}
}

\IEEEtitleabstractindextext{%
\begin{abstract}
  In this paper, we study the problem of jointly estimating the optical flow and scene flow from synchronized 2D and 3D data. Previous methods either employ a complex pipeline that splits the joint task into independent stages, or fuse 2D and 3D information in an ``early-fusion'' or ``late-fusion'' manner. Such one-size-fits-all approaches suffer from a dilemma of failing to fully utilize the characteristic of each modality or to maximize the inter-modality complementarity. To address the problem, we propose a novel end-to-end framework, which consists of 2D and 3D branches with multiple bidirectional fusion connections between them in specific layers. Different from previous work, we apply a point-based 3D branch to extract the LiDAR features, as it preserves the geometric structure of point clouds. To fuse dense image features and sparse point features, we propose a learnable operator named bidirectional camera-LiDAR fusion module (Bi-CLFM). We instantiate two types of the bidirectional fusion pipeline, one based on the pyramidal coarse-to-fine architecture (dubbed CamLiPWC), and the other one based on the recurrent all-pairs field transforms (dubbed CamLiRAFT). On FlyingThings3D, both CamLiPWC and CamLiRAFT surpass all existing methods and achieve up to a 47.9\% reduction in 3D end-point-error from the best published result. Our best-performing model, CamLiRAFT, achieves an error of 4.26\% on the KITTI Scene Flow benchmark, ranking 1st among all submissions with much fewer parameters. Besides, our methods have strong generalization performance and the ability to handle non-rigid motion. Code is available at \url{https://github.com/MCG-NJU/CamLiFlow}.
\end{abstract}

}

\maketitle

\IEEEraisesectionheading{\section{Introduction}}
Optical flow represents the pixel motion of adjacent frames, while scene flow is a 3D motion field of the dynamic scene. Through them, we can gain insights into the dynamics of the scene, which is critical to some high-level scene understanding tasks. In this work, we focus on the joint estimation of optical flow and scene flow, which addresses monocular camera frames with sparse point clouds from LiDAR (or dense depth maps from stereo cameras).

\begin{figure}[t]
  \centering
  \captionsetup[subfloat]{captionskip=3pt}
  \subfloat[Early Fusion]{%
    \includegraphics[page=1,width=0.8\linewidth]{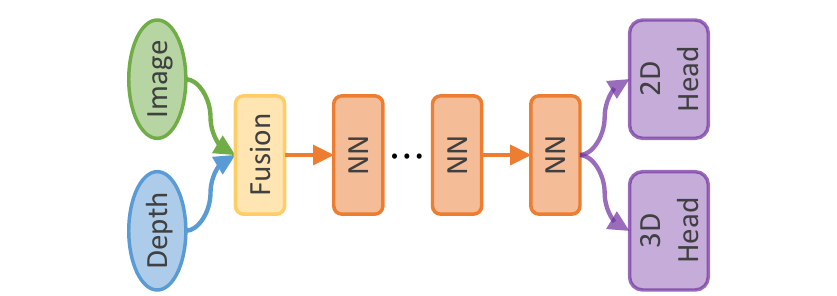}%
    \label{fig:fusion-early}}%
  \vspace{-6pt}
  \subfloat[Late Fusion]{%
    \includegraphics[page=2,width=0.8\linewidth]{images/fusion-comparison.pdf}%
    \label{fig:fusion-late}}%
  \vspace{-6pt}
  \subfloat[Our Fusion]{%
    \includegraphics[page=3,width=0.8\linewidth]{images/fusion-comparison.pdf}%
    \label{fig:fusion-ours}}%
  \caption{Architectures for feature-level fusion. Different from previous methods which adopt an early/late fusion manner, we propose a multi-stage and bidirectional fusion pipeline.}
\end{figure}

Previous methods \cite{yang2021rigidmask, ma2019drisf, yang2020opticalexp, behl2017isf} have constructed modular networks that break down the estimation of scene flow into several subtasks. Despite achieving impressive results, these submodules operate independently, preventing the utilization of their complementary nature. Furthermore, any limitations in a submodule can negatively impact the overall performance, as the entire pipeline relies on its results. On the other hand, some researchers \cite{teed2021raft3d, rishav2020deeplidarflow} employ feature-level fusion and construct end-to-end architectures based on well-known optical flow architectures. RAFT-3D \cite{teed2021raft3d}, which is built upon RAFT \cite{teed2020raft}, merges images and dense depth maps into RGB-D frames and feeds them into a unified 2D network to predict pixel-wise 3D motion. However, this type of ``early fusion'' (Fig. \ref{fig:fusion-early}) results in 2D CNNs failing to leverage most of the 3D structural information provided by depth. DeepLiDARFlow \cite{rishav2020deeplidarflow} incorporates images and LiDAR point clouds as input. In this process, points are projected onto the image plane for densification and subsequently fused with images using a ``late-fusion'' approach (Fig. \ref{fig:fusion-late}). However, errors generated in the early stages lack the opportunity for correction by the other modality and are accumulated into subsequent stages. As a result, the potential for complementarity between the two modalities is not fully realized.

\begin{figure}[t]
  \hspace*{0.46cm}
  \includegraphics[width=0.81\linewidth]{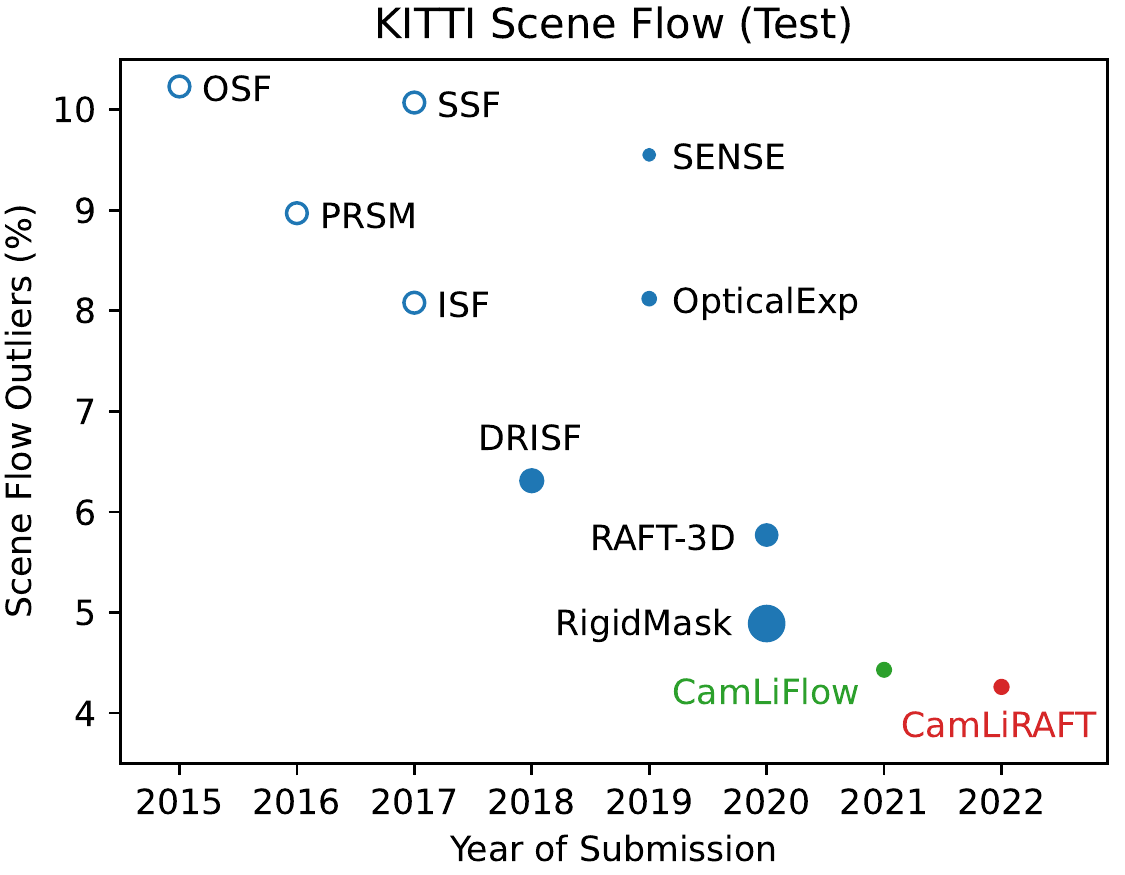}
  \caption{Leaderboard of the KITTI Scene Flow Benchmark. Marker size indicates model size. Models with unknown sizes and conventional approaches are marked as hollow. Our method outperforms all existing methods \cite{menze2015osf,ren2017ssf,vogel2015prsm,behl2017isf,ma2019drisf,jiang2019sense,yang2020opticalexp,teed2021raft3d,yang2021rigidmask} with much fewer parameters.}
  \label{fig:kitti-leaderboard}
  \vspace{-10pt}
\end{figure}

Generally, single-stage fusion encounters a dilemma: it fails to fully leverage the characteristics of each modality or to maximize the complementarity between modalities. To overcome this challenge, we introduce a \textbf{multi-stage} and \textbf{bidirectional} fusion pipeline (refer to Fig. \ref{fig:fusion-ours}). This approach not only enhances performance but also reduces the number of parameters. Within each stage, the two modalities are processed in separate branches using modality-specific architectures. At the end of each stage, a learnable bidirectional bridge interconnects the two branches, facilitating the exchange of complementary information.
The full network can be trained in an end-to-end manner with a multi-task loss. Our fusion scheme is general and can be applied to different architectures, such as PWC-Net \cite{sun2018pwc} and RAFT \cite{teed2020raft}. Moreover, recent point-based methods \cite{qi2017pointnet, qi2017pointnet++, liu2019flownet3d, wu2019pointpwc} achieve remarkable progress for LiDAR scene flow. This inspires us to process the LiDAR point clouds with a point-based branch, which can extract fine 3D geometric information without any voxelization or projection. Thus, our point branch directly consumes point clouds and predicts point-wise scene flow, while our image branch estimates dense pixel-wise optical flow from input images.

It is worth noting that there are three kinds of mismatches for the fusion of the image branch and the point branch. \textit{First}, the data structure of the image feature and the point feature do not match. Specifically, image features are organized in a dense grid structure, while point clouds do not conform to the regular grid and are sparsely distributed in the continuous domain. As a result, there is no guarantee of one-to-one correspondence between pixels and points. This problem becomes more intractable when the features need to be fused in a bidirectional manner. Here, we propose a new learnable fusion operator, named \textit{bidirectional camera-LiDAR fusion module} (Bi-CLFM), in which features are fused for both directions: 2D-to-3D and 3D-to-2D. The 2D-to-3D process can be understood as dense-to-sparse since points are projected to the image plane to retrieve the corresponding 2D feature with bilinear grid sampling. For the 3D-to-2D process which is sparse-to-dense, we propose \textit{learnable interpolation} to densify the sparse 3D features. 

\textit{Second}, the performance of the two branches does not match. For the image branch, we can easily adopt existing optical flow methods (e.g. RAFT \cite{teed2020raft}). However, building a strong point branch that matches the architecture and performance of the image branch is non-trivial. Although there are some off-the-shelf networks \cite{kittenplon2021flowstep3d, wei2021pvraft} which extend the idea of RAFT to 3D space, it is difficult to adopt them as our point branch due to poor performance, heterogenous architecture, and heavy computation. To tackle this problem, we build our point branch from scratch, which is completely homogeneous to RAFT. The key contribution is the point-based correlation pyramid, which can capture both small and large motion. Equipped with better design and optimized implementation, our point branch substantially outperforms the competitors while running much faster.

\textit{Third}, the gradient scale of the two branches does not match. As we find in our experiments, fusing two branches belonging to different modalities may encounter scale-mismatched gradients, making the training unstable and dominated by one modality. In this paper, we propose a simple but effective strategy by detaching the gradient from one branch to the other, so that each branch focuses on its task. Experiments demonstrate that our gradient detaching strategy makes the training more stable and significantly boosts performance.



We instantiate our fusion scheme on two popular architectures, the recurrent all-pairs field transforms (RAFT) \cite{teed2020raft}, and the pyramidal coarse-to-fine strategy (PWC) \cite{sun2018pwc}. The two models are named CamLiRAFT and CamLiPWC respectively. Experiments demonstrate that our approach achieves better performance with much fewer parameters. On FlyingThings3D \cite{mayer2016things3d}, both CamLiPWC and CamLiRAFT surpass all existing methods and achieve up to a 47.9\% reduction in end-point-error over RAFT-3D with only 1/5 parameters. On KITTI \cite{menze2015osf}, even the non-rigid CamLiRAFT performs on par with the previous state-of-the-art method \cite{yang2021rigidmask} which leverages strict rigid-body assumptions (SF-all: 4.97\% vs. 4.89\%). By refining the background scene flow with rigid priors, CamLiRAFT further achieves an error of 4.26\%, outperforming all previous methods by a large margin (the leaderboard is shown in Fig. \ref{fig:kitti-leaderboard}). Our methods also have strong generalization performance and can handle non-rigid motion. Without finetuning on Sintel \cite{sintel}, CamLiRAFT achieves 2.38 AEPE on the final pass of the Sintel training set, reducing the error by 12.2\% and 18.2\% over RAFT and RAFT-3D respectively. Moreover, the LiDAR-only variants of our method, dubbed CamLiPWC-L and CamLiRAFT-L, also outperform all previous LiDAR-only scene flow methods in terms of both accuracy and speed.

To sum up, we make \textbf{four major contributions}: 

\begin{itemize}
  \item We propose a bidirectional and multi-stage camera-LiDAR fusion pipeline for optical flow and scene flow estimation. Our pipeline is universal and can be applied to various network architectures.
  \item We instantiate two types of the bidirectional fusion pipeline, one based on the pyramidal coarse-to-fine architecture (dubbed CamLiPWC), and the other based on the recurrent all-pairs field transforms (dubbed CamLiRAFT).
  \item We design a learnable fusion operator (Bi-CLFM) to align and fuse the image and point features in a bidirectional manner via learnable interpolation and bilinear sampling. A gradient detaching strategy is also introduced to prevent one modality from dominating the training.
  \item On FlyingThings3D and KITTI, Our method achieves state-of-the-art performance in both camera-LiDAR and LiDAR-only settings. Experiments on Sintel also demonstrate its strong generalization performance and ability to handle non-rigid motion.
\end{itemize}

A preliminary version of this work, CamLiFlow \cite{liu2022camliflow}, has been accepted to CVPR 2022 as an \textbf{oral} presentation. In this journal version, we extend our previous work in a number of aspects:
(1) We instantiate two types of the bidirectional fusion pipeline, the original CamLiPWC based on the pyramidal coarse-to-fine architecture, and the new CamLiRAFT based on the recurrent all-pairs field transforms. CamLiRAFT obtains consistent performance improvements over the original CamLiPWC and sets a new state-of-the-art record on various datasets.
(2) For the point branch of CamLiRAFT (dubbed CamLiRAFT-L), we optimize the implementation and propose a point-based correlation pyramid to capture both small and large motion. Compared with other RAFT-based LiDAR scene flow methods, CamLiRAFT-L achieves superior performance while being $5\times$ more efficient.
(3) We improve the bidirectional camera-LiDAR fusion module (Bi-CLFM) with channel attention, enabling the model to adaptively select features to fuse.
(4) We validate our methods on Sintel without finetuning. Experiments demonstrate that our methods have good generalization performance and can handle non-rigid motion.


\section{Related Work}

\subsection{Optical Flow}

Optical flow estimation aims to predict dense 2D motion for each pixel from a pair of frames. We roughly categorize the related optical flow estimation approaches into two types: \textit{traditional methods} and those based on \textit{convolutional neural network (CNN)}.

\vspace{5pt} \noindent
\textbf{Traditional Methods.} Traditional methods \cite{horn1981determining, black1996robust, zach2007duality, brox2009large, weinzaepfel2013deepflow, brox2004warping, bruhn2005lucas} formulate optical flow estimation as an energy minimization problem. Horn and Schunck \cite{horn1981determining} propose the variational approach to estimate optical flow by imposing a tradeoff between a data term and a regularization term. Black and Anandan \cite{black1996robust} introduce a robust framework to address the problem of over-smoothing and noise sensitivity. Other methods improve the data terms \cite{zach2007duality} and the matching costs \cite{brox2009large, weinzaepfel2013deepflow}.

\vspace{5pt} \noindent
\textbf{CNN-based Methods.} Since Krizhevsky \textit{et al.} \cite{krizhevsky2012imagenet} demonstrate that convolutional neural network performs well on large-scale image classification, many researchers start to explore CNN-based approaches for various computer vision tasks. FlowNet \cite{dosovitskiy2015flownet} is the first end-to-end trainable CNN for optical flow estimation, which adopts an encoder-decoder architecture. FlowNet2 \cite{ilg2017flownet2} stacks several FlowNets into a larger one. PWC-Net \cite{sun2018pwc} and some other methods \cite{ranjan2017spynet, hur2019iterative, hui2018liteflownet, yang2019volumetric} apply iterative refinement using coarse-to-fine pyramids. These coarse-to-fine approaches tend to miss small and fast-moving objects which disappear at coarse levels. To solve this, RAFT \cite{teed2020raft} constructs 4D cost volumes for all pairs of pixels and updates the flow iteratively at high resolution. In this work, we implement our bidirectional fusion pipeline based on two representative optical flow architectures: PWC-Net \cite{sun2018pwc}, and RAFT \cite{teed2020raft}.

\subsection{Scene Flow}

Scene flow is similar to optical flow, except that scene flow is the motion field in 3D space, while optical flow is defined in 2D space. Some approaches estimate dense pixel-wise scene flow from monocular or stereo images, while others focus on estimating sparse scene flow from point clouds.

\vspace{5pt} \noindent
\textbf{Scene Flow from RGB Images.} This line of work estimates dense 3D motion for each pixel from a pair of monocular \cite{yang2020opticalexp, hur2020self, hur2021self, bayramli2023raft} or stereo \cite{menze2015osf, quiroga2014dense, jaimez2015primal, jaimez2015motion, behl2017isf, ma2019drisf, yang2021rigidmask, aleotti2020learning, teed2021raft3d} frames. Here, we mainly talk about stereo scene flow, since our method belongs to this category. Like optical flow, traditional methods \cite{menze2015osf, quiroga2014dense, jaimez2015primal, jaimez2015motion} explore variational optimization and discrete optimization and treat scene flow as an energy minimization problem. Recent methods \cite{behl2017isf, ma2019drisf, yang2021rigidmask} divide scene flow estimation into multiple subtasks and build a modular network with one or more submodules for each subtask. Specifically, DRISF \cite{ma2019drisf} estimates optical flow, depth, and segmentation from two consecutive stereo images and employs a Gaussian-Newton solver to find the best 3D rigid motion. RigidMask \cite{yang2021rigidmask} predicts segmentation masks for the background and multiple rigidly moving objects, which are then parameterized by 3D rigid transformation. Although achieving remarkable progress, their submodules are independent of each other, which can not exploit the complementary characteristics of different modalities. DWARF \cite{aleotti2020learning} presents a lightweight architecture that infers scene flow by jointly reasoning about depth and optical flow, leveraging knowledge learned by networks specialized in stereo or flow to distill proxy annotations. RAFT-3D \cite{teed2021raft3d} explores feature-level fusion and concatenates images and depth maps to RGB-D frames at an early stage, followed by a unified 2D network that iteratively updates a dense field of pixel-wise SE3 motion. However, this kind of ``early fusion'' makes it hard for 2D CNNs to take advantage of the 3D structural information.

\begin{figure*}
  \centering
  \includegraphics[width=0.95\linewidth]{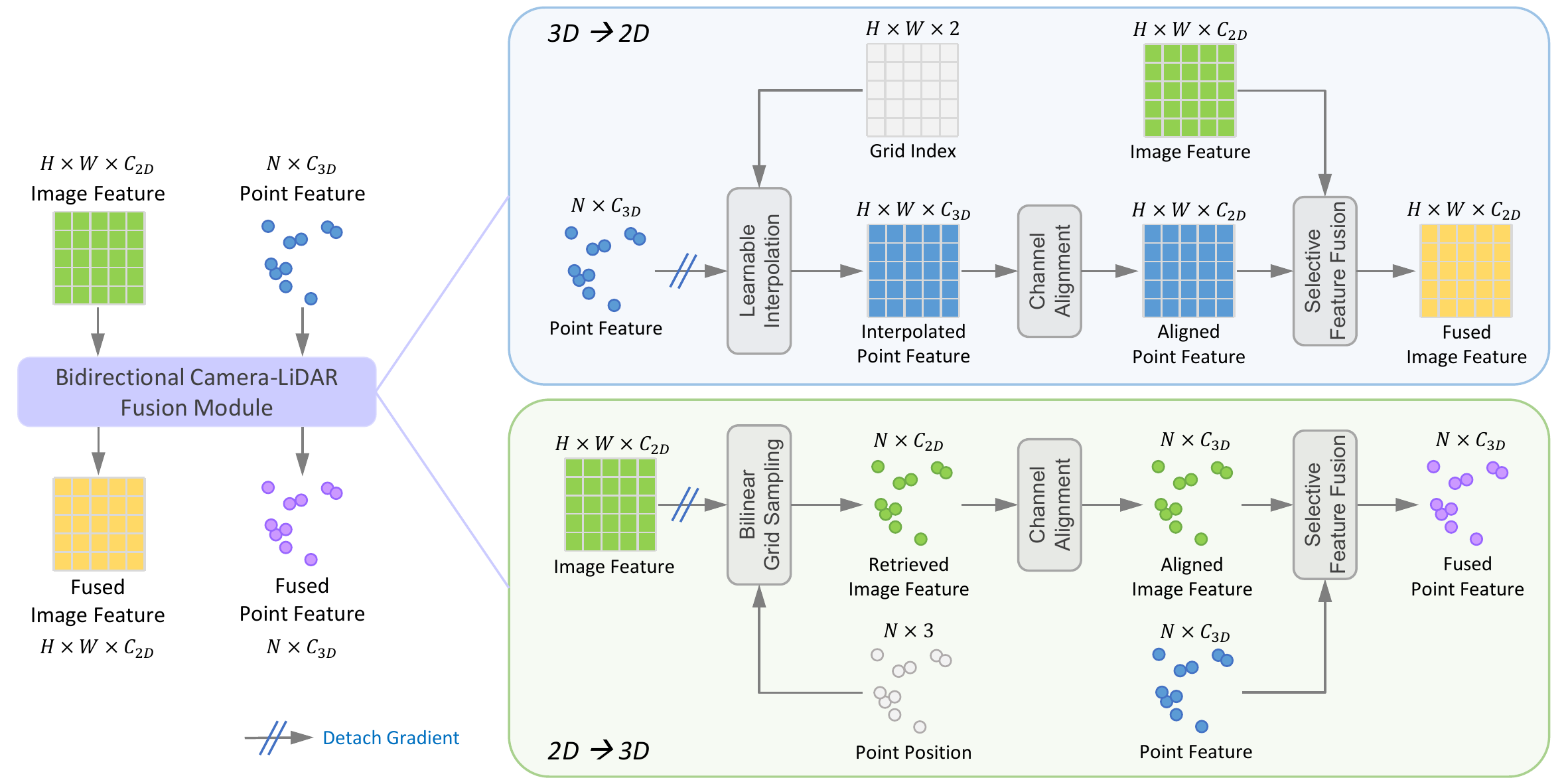}
  \vspace{-3pt}
  \caption{Details of Bidirectional Camera-LiDAR Fusion Module (Bi-CLFM). Features from two different modalities are fused in a bidirectional way, so that both modalities can benefit each other. We detach the gradient from one branch to the other to prevent one modality from dominating.}
  \label{fig:clfm}
\end{figure*}

\vspace{5pt} \noindent
\textbf{Scene Flow from Point Clouds.} PointNet is a pioneering effort that studies deep learning on point sets and directly processes 3D points (e.g. from LiDAR). Since then, researchers \cite{liu2019flownet3d, wang2020flownet3d++, gu2019hplflownet, wu2019pointpwc, puy2020flot, liu2019meteornet, wei2021pvraft, kittenplon2021flowstep3d} start to explore point-based approaches for scene flow estimation. Based on PointNet++ \cite{qi2017pointnet++}, FlowNet3D \cite{liu2019flownet3d} uses a flow embedding layer to represent the motion of points. FlowNet3D++ \cite{wang2020flownet3d++} achieves better performance by adding geometric constraints. Inspired by Bilateral Convolutional Layers, HPLFlowNet \cite{gu2019hplflownet} projects the points onto a permutohedral lattice. PointPWC-Net \cite{wu2019pointpwc} introduces a learnable cost volume for point clouds and estimates scene flow in a coarse-to-fine fashion. FLOT \cite{puy2020flot} addresses the scene flow estimation as a graph matching problem between corresponding points in the adjacent frames and solves it using optimal transport. PV-RAFT \cite{wei2021pvraft} proposes point-voxel correlation fields to capture both local and long-range dependencies of point pairs. FlowStep3D \cite{kittenplon2021flowstep3d} designs a recurrent architecture that learns to iteratively refine scene flow predictions. However, these methods do not exploit the color features provided by images, which limits the performance.

\subsection{Camera-LiDAR Fusion}

Cameras and LiDARs have complementary characteristics, facilitating many computer vision tasks, such as depth estimation \cite{ma2018sparse, you2019pseudo, feng2021advancing}, scene flow estimation \cite{battrawy2019lidarflow, rishav2020deeplidarflow, poggi2021sensor}, and 3D object detection \cite{qi2018frustum, liang2018continuous, xu2018pointfusion, vora2020pointpainting, liu2022bevfusion, bai2022transfusion, yan2023cross}. These methods can be divided into \textit{result-level} and \textit{feature-level} fusion.

\vspace{5pt} \noindent
\textbf{Result-level Fusion.} Some researchers \cite{yang2018ipod, qi2018frustum, vora2020pointpainting} build a modular network and perform result-level fusion. F-PointNet \cite{qi2018frustum} uses off-the-shelf 2D object detectors to limit the 3D search space for 3D object detection, which significantly reduces computation and improves run-time. IPOD \cite{yang2018ipod} replaces the 2D object detector with 2D semantic segmentation and uses point-based proposal generation. PointPainting \cite{vora2020pointpainting} projects LiDAR point clouds into the output of an image-only semantic segmentation network and appends the class scores to each point. However, the performance of result-level fusion is limited by each submodule, since the whole network depends on its results. In contrast, our method exploits feature-level fusion and can be trained in an end-to-end manner.

\vspace{5pt} \noindent
\textbf{Feature-level Fusion.} Another approach is feature-level fusion. PointFusion \cite{xu2018pointfusion} utilizes a 2D object detector to generate 2D boxes, followed by a CNN-based and a point-based network to fuse image and point features for 3D object detection. MVX-Net \cite{sindagi2019mvx} uses a pretrained 2D Faster R-CNN to extract the image features and a VoxelNet to generate final boxes. Points or voxels are projected to the image plane and the corresponding features are fused with 3D features. Liang \textit{el al.} \cite{liang2018continuous} exploit continuous convolutions to fuse the image features onto the BEV feature maps at different levels of resolution. BEVFusion \cite{liu2022bevfusion} uses a lift-splat-shoot operation to transform the camera features into BEV space and fuse the two modalities with a BEV Encoder. TransFusion \cite{bai2022transfusion} follows a two-stage pipeline: the queries are generated from the LiDAR features and interact with 2D and 3D features separately. CMT \cite{yan2023cross} explores cross modal transformer and aligns multi-modal features implicitly with coordinate encoding. Different from previous work, we propose a multi-stage and bidirectional fusion pipeline, which not only fully utilizes the characteristic of each modality, but maximizes the inter-modality complementarity as well.

\section{Approach}

In the following sections, we first introduce the bidirectional camera-LiDAR fusion module (Bi-CLFM). Next, we implement the bidirectional fusion pipeline on two popular architectures, the pyramidal coarse-to-fine strategy \cite{sun2018pwc} and the recurrent all-pairs field transforms \cite{teed2020raft}.

\subsection{Bidirectional Camera-LiDAR Fusion Module}

In this section, we introduce \textit{bidirectional camera-LiDAR fusion module} (Bi-CLFM), which can fuse dense image features and sparse point features in a bidirectional manner (2D-to-3D and 3D-to-2D).

As illustrated in Fig. \ref{fig:clfm}, Bi-CLFM takes image features $F_{2D} \in \mathbb{R}^{H \times W \times C_{2D}}$, point features $G_{3D} = \{g_i | i = 1, ..., N\} \in \mathbb{R}^{N \times C_{3D}}$ and point positions $P = \{ p_i | i = 1, ..., N \} \in \mathbb{R}^{N \times 3}$ as input, where $N$ denotes the number of points. The output contains the fused image and point features. Thus, Bi-CLFM performs bidirectional fusion without altering the spatial structure of the input features, which can be easily plugged into any point-image fusion architecture.

For each direction (2D-to-3D or 3D-to-2D), the features are first aligned to have the same spatial structure using bilinear grid sampling and learnable interpolation. Next, the aligned features are fused adaptively based on selective kernel convolution \cite{li2019sknet}. Besides, we introduce our gradient detaching strategy which solves the problem of scale-mismatched gradients.

\subsubsection{Feature Alignment}

Since image features are dense while point features are sparse, we need to align the features of the two modalities before the fusion. Specifically, the image features need to be sampled to be sparse, while the point features need to be interpolated to become dense.

\vspace{5pt} \noindent
\textbf{2D $\Rightarrow$ 3D.} First, points are projected to the image plane (denoted as $X = \{ x_i | i = 1, ..., N \} \in \mathbb{R}^{N \times 2}$) to retrieve the corresponding 2D feature:
\begin{equation}
    G_{2D} = \{ F_{2D}(x_i) | i = 1, ..., N \} \in \mathbb{R}^{N \times C_{2D}},
\end{equation}
where $F_{2D}(x)$ denotes the image feature at $x$ and can be retrieved by bilinear interpolation to cope with non-integer coordinates. Next, the retrieved feature $H$ is aligned to have the same shape as the input point feature ($N \times C_{3D}$) by a $1 \times 1$ convolution.

\vspace{5pt} \noindent
\textbf{3D $\Rightarrow$ 2D.} Similarly, points are first projected to the image plane (denoted as $X = \{ x_i | i = 1, ..., N \} \in \mathbb{R}^{N \times 2}$). Since point clouds are sparse, we propose a learnable interpolation module (detailed in the next paragraph) to create a dense feature map $F_{3D} \in \mathbb{R}^{H \times W \times C_{3D}}$ from sparse point features. Next, the ``interpolated'' point feature $F_{3D}$ is aligned to have the same shape of the input image feature ($H \times W \times C_{2D}$) by a $1 \times 1$ convolution.

\begin{figure}[t]
  \centering
  \includegraphics[width=0.9\linewidth]{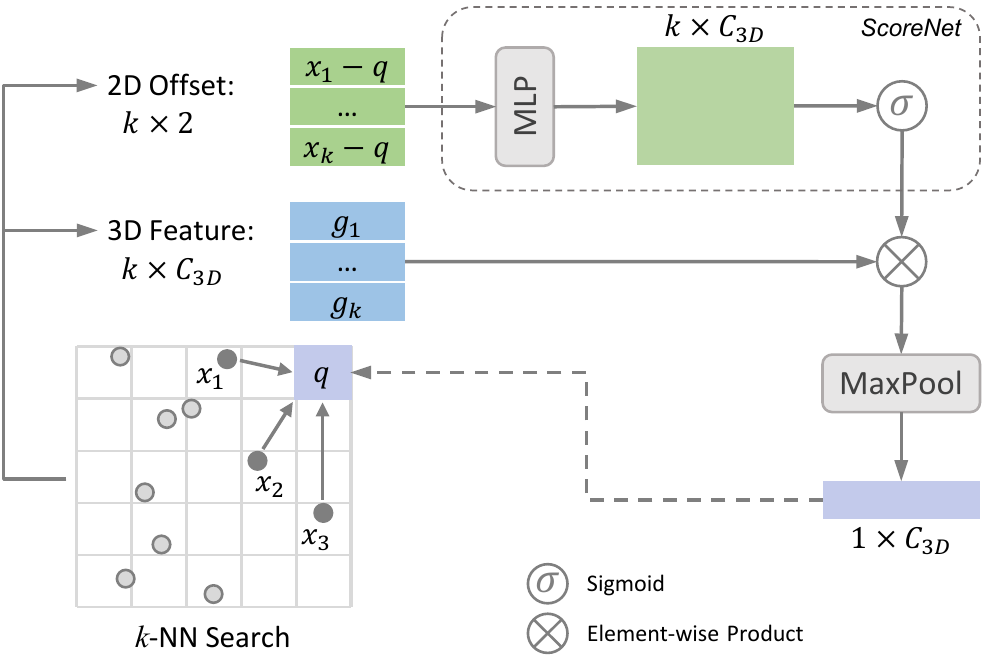}
  \vspace{-5pt}
  \caption{Details of Learnable Interpolation. For each target pixel, we find the $k$ nearest points around it. A lightweight MLP followed by a sigmoid activation is used to weigh the neighboring features.}
  \label{fig:learnable-interp}
  \vspace{-5pt}
\end{figure}

\vspace{5pt} \noindent
\textbf{Learnable Interpolation.} To solve the problem of fusing sparse point features into dense image features, we propose a learnable interpolation module. As illustrated in Fig. \ref{fig:learnable-interp}, for each target pixel $q$ in the dense map, we find its $k$ nearest neighbors among the projected points over the image plane. Next, we use a ScoreNet to weigh the neighboring features according to the coordinate offset. The ScoreNet is a lightweight MLP followed by a sigmoid activation to give scores ranging in $(0, 1)$. Finally, the neighboring features are weighted by the scores and aggregated with a symmetric operation (e.g. \texttt{max} or \texttt{sum}).

Note that the interpolation module here is different from the one in the conference version \cite{liu2022camliflow}. The modifications lie in two aspects: 1) instead of processing the concatenation of the neighboring offsets and the features with a heavy MLP, we explicitly build a lightweight ScoreNet to generate weights for the neighboring features; 2) we remove the 2D similarity measurements between $q$ and its neighbors, as we empirically find that it no longer improves the performance.

\subsubsection{Adaptive Feature Fusion}
\label{sec:feature-fusion}

In the section above, 2D and 3D features are aligned to have the same spatial shape and are ready for feature fusion. Common feature fusion operations include addition and concatenation, but both of them can be influenced by massive useless features. Inspired by Selective Kernel Networks (SKNet)\cite{li2019sknet}, we propose to fuse them adaptively based on channel attention.

As shown in Fig. \ref{fig:clfm}, Bi-CLFM contains two selective fusion modules with the same architectural design. The only difference is that one of them is responsible for dense feature fusion (3D-to-2D), while the other takes sparse features as input (2D-to-3D).

Take the dense one as an example. Given the image feature $F_{2D}$ and the ``interpolated'' point feature $F_{3D}$ (both of them have the shape of $H \times W \times C$), we first generate channel-wise statistics $z \in \mathbb{R}^{C}$ from the inputs:
\begin{equation}
  z = \mathbf{P}(F_{2D} + F_{3D}),
\end{equation}
where $\mathbf{P}$ denotes global average pooling. A fully connected layer further reduces the dimension of $z$ from $C$ to $C/r$ to create a compact feature $\mathbf{z'}$, which enables the guidance for more precise and adaptive selections.

Next, the attention scores $A \in \mathbb{R}^{C \times 2}$ are predicted using the compact channel statistics $\mathbf{z'}$:
\begin{equation}
  A = \text{Softmax}(\text{Linear}_{C/r \rightarrow C \times 2}(\mathbf{z'})),
\end{equation}
where the softmax operator is applied on the channel-wise digits to ensure that $A_{i,0} + A_{i,1} = 1$ holds for any $i \in \{1, 2, ..., C\}$. Finally, the fused feature map $\hat{F}$ is obtained with by a weighted sum of $F_{2D}$ and $F_{3D}$:
\begin{align}
  &\hat{F}_i = A_{i,0} \cdot F_{2D} + A_{i,1} \cdot F_{3D}, \\
  &\hat{F} = [\hat{F}_1, \hat{F}_2, ..., \hat{F}_C],
\end{align}
where $[\cdot]$ denotes concatenation along the channel axis.

\begin{figure}[t]
  \centering
  \subfloat{%
    \includegraphics[width=0.5\linewidth]{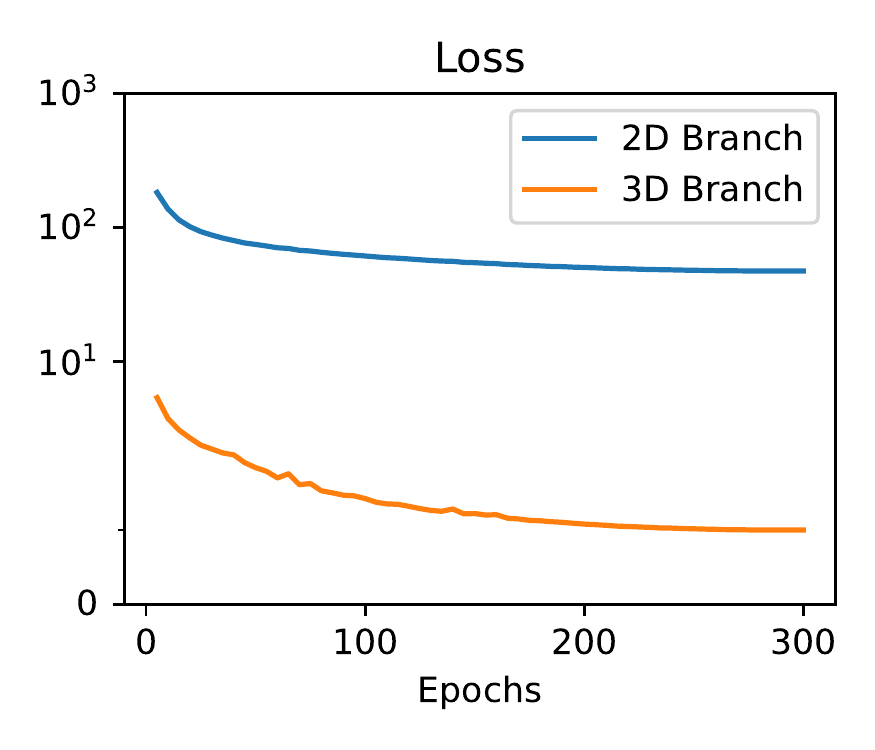}}%
  \hfill
  \subfloat{%
    \includegraphics[width=0.5\linewidth]{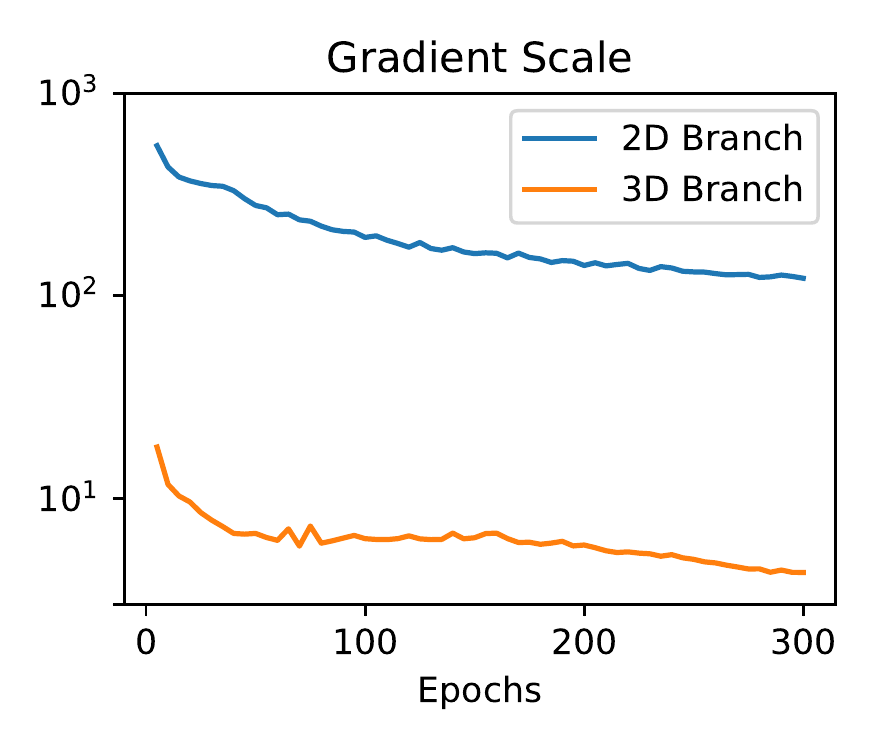}}%
  \vspace{-10pt}
  \caption{The loss and gradient scale of the two branches in CamLiPWC. The 2D gradients are $\sim$40x larger than 3D gradients and the gap does not shrink as the number of training epochs increases.}
  \label{fig:grad_norm}
  \vspace{-5pt}
\end{figure}

\begin{figure*} 
  \centering
  \includegraphics[width=\linewidth]{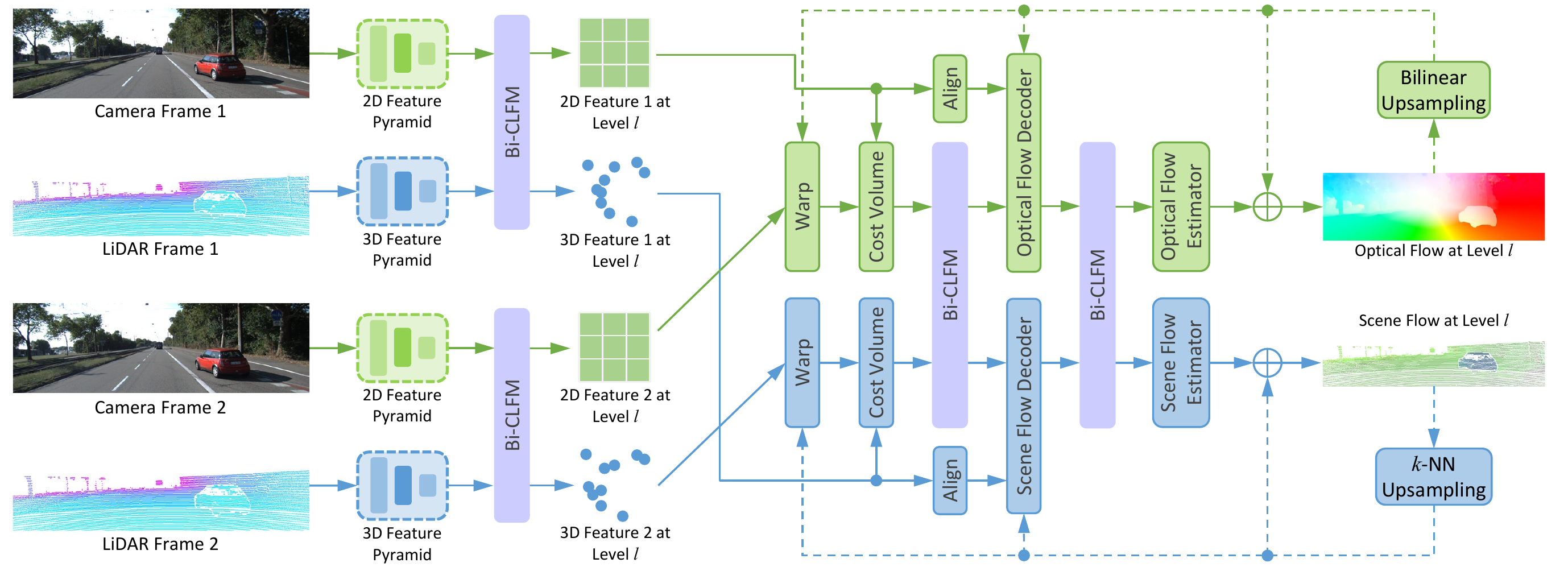}
  \vspace{-15pt}
  \caption{The architecture of CamLiPWC. Synchronized camera and LiDAR frames are taken as input, from which dense optical flow and sparse scene flow are estimated respectively. Built on top of the PWC architecture, CamLiPWC is a two-branch network with multiple bidirectional fusion connections (Bi-CLFM) between them.}
  \label{fig:camlipwc}
  \vspace{-3pt}
\end{figure*}

\subsubsection{Gradient Detaching}

As mentioned before, multi-modal fusion may encounter scale-mismatched gradients, making the training unstable and dominated by one modality. To solve the problem, we propose to detach the gradient from one branch to the other in Bi-CLFM, as shown in Fig. \ref{fig:clfm}.

In this section, we first perform statistical analysis on the gradient scale, which is obtained by calculating the $L2$-norm of the gradient. As we can see from Fig. \ref{fig:grad_norm}, the loss and gradient scale of the two branches are very different. The 2D gradients are $\sim$40x larger than 3D gradients and the gap does not shrink as the number of training epochs increases. This may be due to the large difference in the data range and feature space of the two modalities. In such a case, directly fusing the features from different modalities can encounter scale-mismatched gradients, which motivates us to detach the gradient from one branch to the other.

We also conduct an ablation to demonstrate the effect of gradient detaching, as shown in Fig. \ref{fig:grad_detach}. Without gradient detaching, the 2D branch dominates the training and hurts the performance of the 3D branch. By detaching the gradient from one branch to the other, we prevent one modality from dominating so that each branch focuses on its task.

\subsection{Pyramidal Coarse-to-fine Fusion Pipeline}

PWC-Net \cite{sun2018pwc} is designed according to simple and well-established principles, including pyramid processing, warping, and the use of a cost volume. Flow computed at the coarse level is upsampled and warped to a finer level. As shown in Fig. \ref{fig:camlipwc}, we introduce CamLiPWC, a two-branch network based on the PWC architecture.

\vspace{5pt} \noindent
\textbf{Basic Architecture.} We use the IRR-PWC \cite{hur2019iterative} as the image branch. The only difference is that we replace the bilinear upsampling with a learnable convex upsampling module \cite{teed2020raft} to produce the final prediction from the desired level. The point branch is based on PointPWC-Net \cite{wu2019pointpwc} with two major modifications. \textit{First, we increase the level of the pyramid to match the image branch.} Thus, the point pyramid has 6 levels with 8192, 4096, 2048, 1024, 512, and 256 points respectively. \textit{Second, the weights of the decoder are shared across all pyramid levels.} According to IRR-PWC, iterative residual refinements with weight sharing can reduce the number of parameters and increase the accuracy.

\vspace{5pt} \noindent
\textbf{Fusion Locations.} The two branches are connected at three locations: (1) \textit{Feature pyramid.} The image pyramid encodes rich textural information, while the point pyramid encodes geometric information. Thus, features are fused at multiple levels to merge textural and structural information. (2) \textit{Correlation feature.} The pixel-based 2D correlation maintains a fixed range of neighborhoods, while the point-based 3D correlation searches for a dynamic range. Thus, the receptive fields can be complementary. (3) \textit{Flow decoder.} Features from the second last layer of the flow decoder are fused to exchange information for final predictions.

\vspace{5pt} \noindent
\textbf{Loss Functions.} Although the estimation of optical flow and scene flow is highly relevant (the projection of scene flow onto the image plane becomes optical flow), we formulate them as two different tasks. We supervise the 2D and 3D branches respectively and design a multi-task loss for joint optimization. Let $\mathbf{o}_l^{gt}$ and $\mathbf{s}_l^{gt}$ be the ground truth optical flow and scene flow at the $l$-th level respectively. The regression loss for each branch is defined as follows:
\begin{align}
    \mathcal{L}_{2D} &= \sum_{l=l_0}^{L} \alpha_l \Vert \mathbf{o}_l - \mathbf{o}_l^{gt} \Vert_2,\\
    \mathcal{L}_{3D} &= \sum_{l=l_0}^{L} \alpha_l \Vert \mathbf{s}_l - \mathbf{s}_l^{gt} \Vert_2,
\end{align}
where $\Vert \cdot \Vert_2$ computes the $L_2$ norm. The loss weights are set to $\alpha_0=8$, $\alpha_1=4$, $\alpha_2=2$, $\alpha_3=1$, and $\alpha_4=0.5$. The final loss is a sum of the losses defined above:
\begin{equation}
    \mathcal{L} = \mathcal{L}_{2D} + \mathcal{L}_{3D}.
\end{equation}

\subsection{Recurrent All-pairs Field Fusion Pipeline}
\label{sec:camliraft}

\begin{figure*} 
  \centering
  \includegraphics[width=\linewidth]{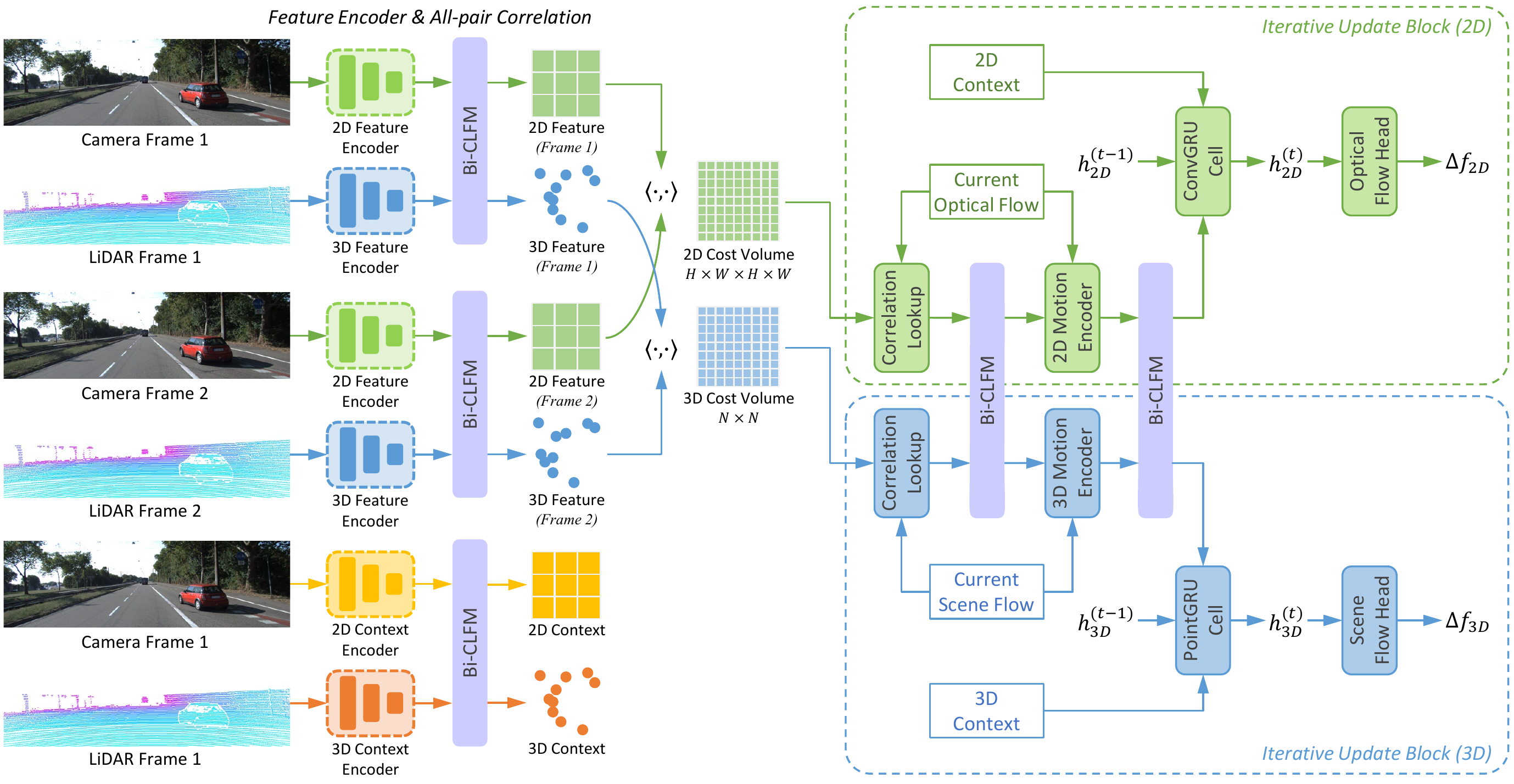}
  \vspace{-15pt}
  \caption{The architecture of CamLiRAFT. Built on top of the RAFT architecture, we perform four-stage feature fusion: features from the feature encoder, the context encoder, the correlation lookup operation, and the motion encoder are fused to pass complementary information.}
  \vspace{-3pt}
  \label{fig:camliraft}
\end{figure*}

Recurrent All-pairs Field Transforms (RAFT) \cite{teed2020raft} is the state-of-the-art algorithm for optical flow estimation, which constructs 4D correlation volumes for all pairs of pixels and updates the flow iteratively using the gated mechanism. In this section, we extend our fusion pipeline to the RAFT architecture and build a model named CamLiRAFT.

Similarly, CamLiRAFT has two homogeneous branches: an image branch and a point branch. The image branch is identical to RAFT. For the point branch, however, it is difficult to adopt the off-the-shelf networks \cite{teed2021raft3d, kittenplon2021flowstep3d, wei2021pvraft} due to the inhomogeneous architecture, complicated training schedule, and heavy computation (see Sec. \ref{sec:lidar-only-cmp} for detailed discussion). To overcome these challenges, we build our point branch from scratch, which is completely homogeneous to RAFT and is suitable for multi-stage fusion. Next, we connect the two branches with multiple Bi-CLFMs to build the complete fusion pipeline, as shown in Fig. \ref{fig:camliraft}.

\vspace{5pt} \noindent
\textbf{Feature Extraction.} We build a feature encoder for each branch. The image encoder consists of 6 residual blocks and outputs features at $1/8$ resolution. The point encoder downsamples the point clouds to $1/4$ using furthest point sampling and aggregates the features using PointConv \cite{wu2019pointconv}.

\vspace{5pt} \noindent
\textbf{Context Network.} Following RAFT \cite{teed2020raft}, we build a context network for each branch to extract semantic and contextual information from the first frame. The architecture of the context network and the feature encoder are the same, but the weights are not shared.

\vspace{5pt} \noindent
\textbf{All-pair Correlation.} The correlation volume for the image branch $\mathbf{V}_{2D} \in \mathbb{R}^{H \times W \times H \times W}$ is defined to be the dot product between all pairs of image features. To capture long-range dependencies, we construct a 4-layer correlation pyramid $\{\mathbf{V}_{2D}^{(1)}, \mathbf{V}_{2D}^{(2)}, \mathbf{V}_{2D}^{(3)}, \mathbf{V}_{2D}^{(4)}\}$ by applying average pooling to the last two dimensions of $\mathbf{V}_{2D}$ with kernel sizes 1, 2, 4 and 8. For the point branch, the correlation volume $\mathbf{V}_{3D} \in \mathbb{R}^{N \times N}$ is defined to be the dot product between all pairs of point features. Similarly, we build a 4-layer correlation pyramid $\{\mathbf{V}_{3D}^{(1)}, \mathbf{V}_{3D}^{(2)}, \mathbf{V}_{3D}^{(3)}, \mathbf{V}_{3D}^{(4)}\}$ using point-based average pooling. The average pooling operation for points is similar with the one for images, except that we downsample the points using \textit{furthest point sampling} and define the neighborhood to be the $k$ nearest neighbors. 

\vspace{5pt} \noindent
\textbf{Correlation Lookup.} In each iteration, we perform lookups on all levels of the correlation pyramid. For the image branch, the $i$th-level correlation feature $\mathbf{C}_{2D}^{(i)}(q)$ is built by indexing a $d \times d$ neighborhood around each query $q$ from $\mathbf{V}_{2D}^{(i)}$, where $d$ is set to 9. The final correlation feature is formed by concatenating the values from each level: $\mathbf{C}_{2D}(q) = [\mathbf{C}^{(1)}_{2D}(q), \mathbf{C}^{(2)}_{2D}(q), \mathbf{C}^{(3)}_{2D}(q), \mathbf{C}^{(4)}_{2D}(q)]$. For the point branch, since the cost volume is sparse, we search the neighborhood around each query using the $k$-nearest neighbor algorithm. Suppose the $i$-th nearest neighbor of the query $q$ in $\mathbf{V}_{3D}^{(l)}$ is $p_i$ and the correlation value is $\mathbf{V}_{3D}^{(l)}(p_i)$, we compute the matching cost between $q$ and $p_i$ in a learnable way:
\begin{equation}
  \mathbf{C}^{(l)}_{3D}(q, p_i) = \text{MLP}([q - p_i, \mathbf{V}^{(l)}_{3D}(p_k)]),
\end{equation}
where $[\cdot]$ denotes concatenation. Next, the $l$-th level correlation feature for $q$ is defined to be the maximum over its $k$ nearest neighbors:
\begin{equation}
  \mathbf{C}^{(l)}_{3D}(q) = \text{Max}_{i=1}^{k} \mathbf{C}^{(l)}_{3D}(q, p_i).
\end{equation}
Finally, we concatenate the correlation features at all four levels to form the output:
\begin{equation}
  \mathbf{C}_{3D}(q) = [\mathbf{C}^{(1)}_{3D}(q), \mathbf{C}^{(2)}_{3D}(q), \mathbf{C}^{(3)}_{3D}(q), \mathbf{C}^{(4)}_{3D}(q)].
\end{equation}

\vspace{5pt} \noindent
\textbf{Iterative Updates.} Given the current estimation of optical flow and scene flow, we use them to retrieve respective correlation features from the cost volume as described in the paragraph above. Next, we follow the implementation of RAFT to build a motion encoder and a GRU cell for the image branch. The motion encoder produces motion features from the correlation features and the current flow estimate. The GRU cell then updates the hidden status from the motion and context features. Finally, we predict the residual optical flow from the updated hidden status using two 3 $\times$ 3 convolutions.

For the point branch, we build the same pipeline but replace the standard 2D convolutions with point-based convolutions.
However, using the original version of the PointConv operator here would significantly increase computational overhead when unrolling many iterations (e.g. 20 $\times$). Thus, we build a depth-wise separable version of PointConv (dubbed PointConvDW), which is inspired by the depth-wise convolution \cite{howard2017mobilenets}. Formally, PointConvDW aggregates features for point $i$ from its neighbors denoted as $\{j : (i, j) \in \mathcal{N} \}$:
\begin{align}
  F'_j &= \text{MLP}_{C_{in} \rightarrow C_{out}} (F_j) \in \mathbb{R}^{C_{out}}, \\
  G_i &= \underset{j}{\mbox{Max}} \: \{ \text{MLP}_{3 \rightarrow C_{out}} (p_j - p_i) \cdot F'_j \: | \: (i,j) \in \mathcal{N} \},
\end{align}
where $p_j \in \mathbb{R}^{3}$ and $F_j \in \mathbb{R}^{C_{in}}$ are the coordinates and features of the neighborhoods. In this way, the depth and spatial dimension of the filter is separated, making PointConvDW run much faster than PointConv. To enable bidirectional feature fusion, we unroll the same number of iterations for the image and point branch.

\vspace{5pt} \noindent
\textbf{Fusion Locations.} For CamLiRAFT, we perform four-stage feature fusion: features from the feature encoder, the context encoder, the correlation lookup operation, and the motion encoder are fused to pass complementary information. Each fusion connection plays its unique role and contributes to the final performance.

\vspace{5pt} \noindent
\textbf{Loss Functions.} Similar to CamLiPWC, we give supervision to each branch and design a multi-task loss. Formally, let $\mathbf{o}_{gt}$ and $\mathbf{s}_{gt}$ be the ground truth optical flow and scene flow respectively and let $\mathbf{o}_{i}$ and $\mathbf{s}_{i}$ be the predicted optical flow and scene flow at the $i$-th iteration. The loss for each branch is defined as follows:
\begin{align}
    \mathcal{L}_{2D} &= \sum_{i=1}^{N} \alpha^{N-i} \Vert \mathbf{o}_{i} - \mathbf{o}_{gt} \Vert_2,\\
    \mathcal{L}_{3D} &= \sum_{i=1}^{N} \alpha^{N-i} \Vert \mathbf{s}_{i} - \mathbf{s}_{gt} \Vert_2,
\end{align}
where $\Vert \cdot \Vert_2$ computes the $L_2$ norm, $N$ is the number of iterations, and $\alpha$ is set to 0.8. The final loss is a sum of the losses defined above:
\begin{equation}
    \mathcal{L} = \mathcal{L}_{2D} + \mathcal{L}_{3D}.
\end{equation}

\subsection{Implementation Details}

\vspace{5pt} \noindent
\textbf{CamLiPWC.} We build a 6-level pyramid for both the image and point branch. We start processing from the top level and perform the coarse-to-fine estimation scheme until level 2. Hence, CamLiPWC outputs optical flow at 1/4 resolution and scene flow at 1/2 resolution. We obtain full-scale optical flow and scene flow by convex upsampling \cite{teed2020raft} and $k$-NN upsampling respectively.

\vspace{5pt} \noindent
\textbf{CamLiRAFT.} The initial optical flow and scene flow are both set to 0. We unroll 10 iterations during training and evaluate after 20 iterations. CamLiRAFT outputs optical flow at 1/8 resolution and scene flow at 1/4 resolution. Similar to CamLiPWC, the full-scale optical flow and scene flow are respectively obtained by convex upsampling and $k$-NN upsampling.

\vspace{5pt} \noindent

\textbf{Inverse Depth Scaling.} LiDAR point clouds exhibit an imbalanced distribution, with a higher density in the nearby region compared to the farther-away region. To tackle this issue, we propose a transformation for point clouds, termed as \textit{inverse depth scaling} (IDS). Formally, let $(P_x, P_y, P_z)$ denote the original coordinate of a point and $(P_x', P_y', P_z')$ denote the coordinate after transformation. All three dimensions are scaled equally by the inverse depth $\frac{1}{P_z}$:
\begin{equation}
    \frac{\delta P_x'}{\delta P_x} = \frac{\delta P_y'}{\delta P_y} = \frac{\delta P_z'}{\delta P_z} = \frac{1}{P_z}.
\end{equation}

The transformed coordinate $(P_x', P_y', P_z')$ can be calculated by integrating the above equation:
\begin{align}
    \label{eq:ids_x}
    P_x' &= \int \frac{1}{P_z} dP_x = \frac{P_x}{P_z} + C_x, \\
    \label{eq:ids_y}
    P_y' &= \int \frac{1}{P_z} dP_y = \frac{P_y}{P_z} + C_y, \\
    \label{eq:ids_z}
    P_z' &= \int \frac{1}{P_z} dP_z = \log{P_z} + C_z,
\end{align}
where $C_x$ and $C_y$ are both set to 0, and $C_z$ is set to 1 to avoid zero depth. In this paper, point clouds undergo IDS before being fed into the neural network.

\section{Experiments}

\begin{table}[t]
  \small
  \renewcommand{\arraystretch}{1.15}
  \caption{Training setting of CamLiPWC and CamLiRAFT on FlyingThings3D.}
  \label{tab:training-things}
  \centering
  \begin{tabular}{l|c|c}
    \hline
    Hyperparameters & CamLiPWC & CamLiRAFT \\
    \hline
    Epochs & 600 & 150 \\
    Batch Size & 32 & 8 \\
    Optimizer & AdamW & AdamW \\
    Weight Decay & 1e-6 & 1e-6 \\
    LR\textsubscript{2D} & 3e-4 & 2e-4 \\
    LR\textsubscript{3D} & 6e-4 & 2e-3 \\
    LR Schedule & Cosine & Cosine \\
    \hline
  \end{tabular}
  \vspace{-5pt}
\end{table}

\begin{table*}
  \caption{Performance comparison on the ``val'' split of the FlyingThings3D subset. ``RGB'' and ``XYZ'' denote the image and point cloud respectively.\\ Underlined results are reported from \cite{ouyang2021ogsf}. $\dag$ pretrained on ImageNet-1k.}
  \vspace{-5pt}
  \label{tab:main-things}
  \centering
  \small
  \renewcommand{\arraystretch}{1.15}
  \begin{tabular}{l|c|cc|cc|c}
  \hline
  \multirow{2}{*}{Method} & \multirow{2}{*}{Input} & \multicolumn{2}{c|}{2D Metrics} & \multicolumn{2}{c|}{3D Metrics} & \multirow{2}{*}{Parameters} \\
  & & EPE\textsubscript{2D} & ACC\textsubscript{1px} & EPE\textsubscript{3D} & ACC\textsubscript{.05} \\
  \hline
  FlowNet2 \cite{ilg2017flownet2}   & RGB & 5.05 & 72.8\% & - & - & 162.5M \\
  PWC-Net \cite{sun2018pwc}         & RGB & 6.55 & 64.3\% & - & - & 9.4M \\
  RAFT \cite{teed2020raft}          & RGB & 3.12 & 81.1\% & - & - & 5.3M \\
  \hline
  FlowNet3D \cite{liu2019flownet3d} & XYZ & - & - & 0.214 & 18.2\% & 1.2M \\
  PointPWC \cite{wu2019pointpwc}    & XYZ & - & - & \underline{0.195}  & - & 7.7M \\
  OGSF-Net \cite{ouyang2021ogsf}    & XYZ & - & - & \underline{0.163} & - & 7.7M \\
  \hline
  RAFT-3D $\dag$ \cite{teed2021raft3d} & RGB+Depth & 2.37 & 87.1\% & 0.094 & 80.6\% & 45M \\
  CamLiFlow \cite{liu2022camliflow}    & RGB+XYZ & 2.18 & 84.3\% & 0.061 & 85.6\% & 7.7M \\
  \hline
  \textbf{CamLiPWC}  & RGB+XYZ & 2.03 & 85.9\% & 0.057 & 86.3\% & 8.5M \\
  \textbf{CamLiRAFT} & RGB+XYZ & 1.76 & 87.1\% & 0.050 & \textbf{88.4\%} & 7.4M\\
  \textbf{CamLiRAFT} $\dag$ & RGB+XYZ & \textbf{1.73} & \textbf{87.5\%} & \textbf{0.049} & \textbf{88.4\%} & 8.4M\\
  \hline  
  \end{tabular}
  \vspace{-12pt}
\end{table*}

\begin{figure*}
    \centering
    \captionsetup[subfigure]{labelformat=empty,position=top}
    \captionsetup[subfloat]{captionskip=1pt}

    \subfloat[Reference Frame]{\includegraphics[width=0.198\linewidth]{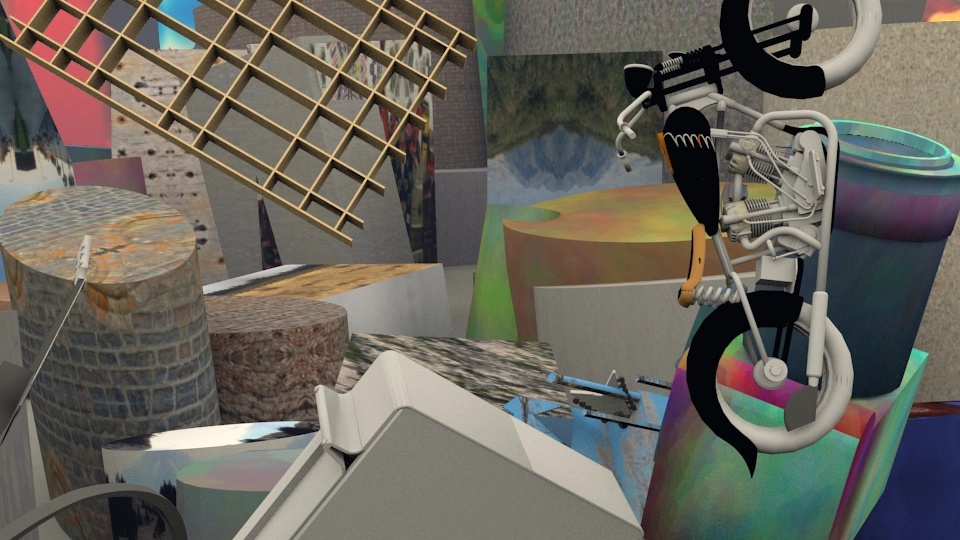}}
    \hfill
    \subfloat[RAFT]{\includegraphics[width=0.198\linewidth]{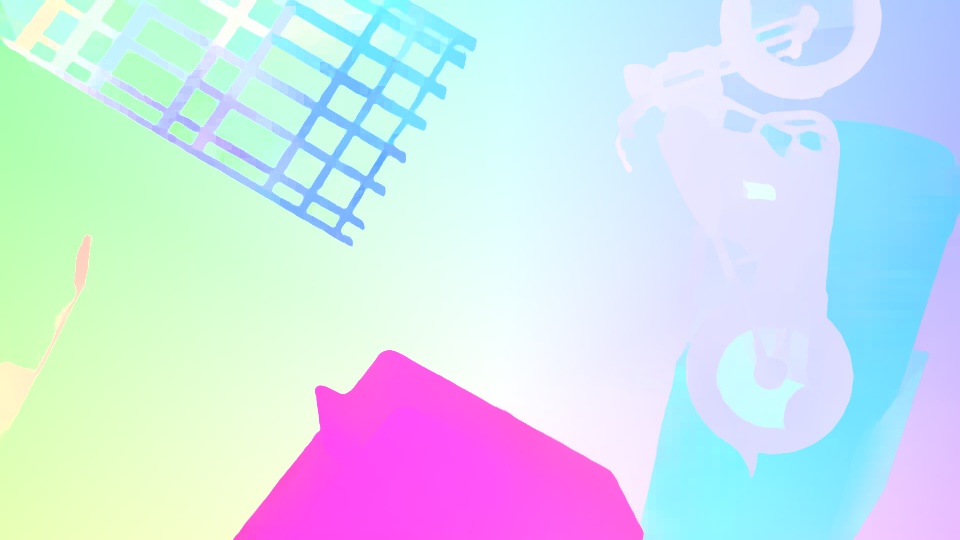}}
    \hfill
    \subfloat[RAFT-3D]{\includegraphics[width=0.198\linewidth]{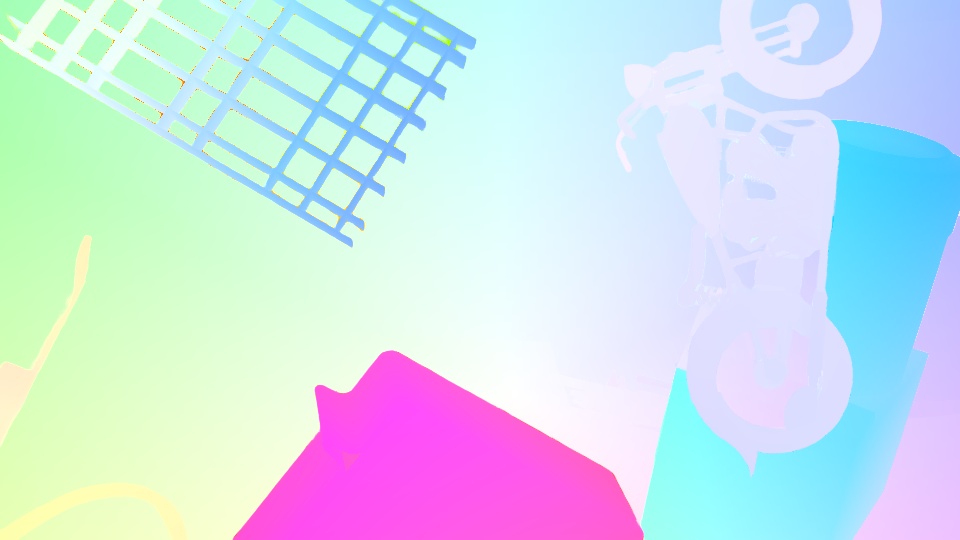}}
    \hfill
    \subfloat[CamLiRAFT (Ours)]{\includegraphics[width=0.198\linewidth]{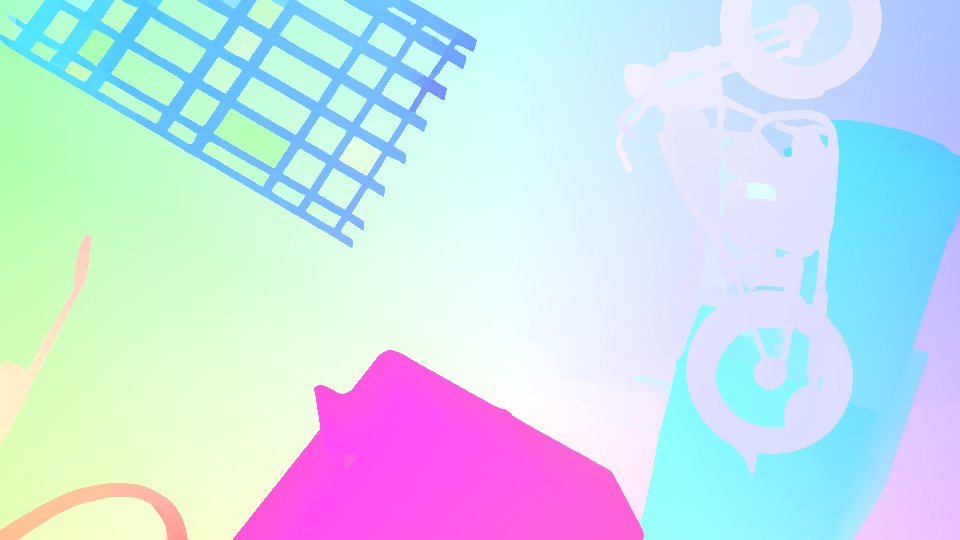}}
    \hfill
    \subfloat[Ground Truth]{\includegraphics[width=0.198\linewidth]{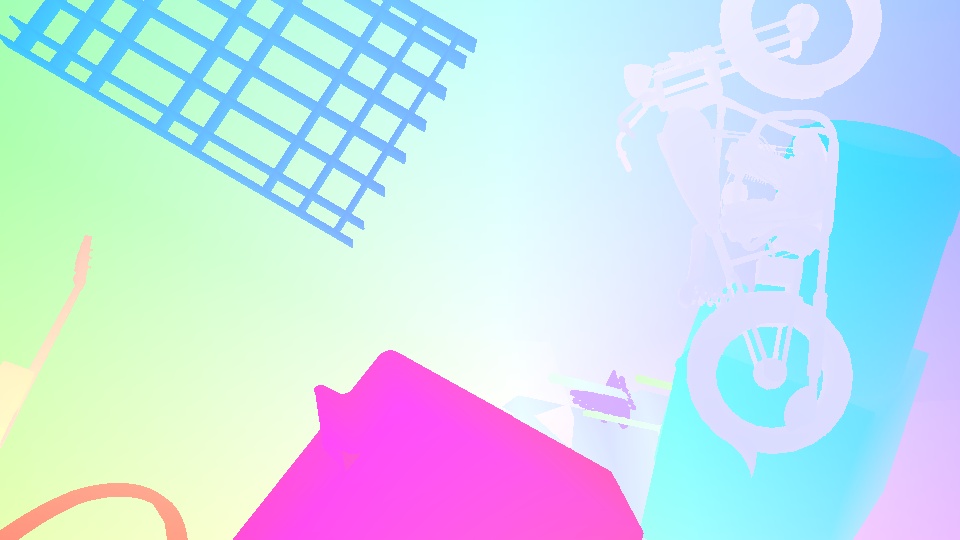}}\\
    \vspace{-9pt}

    \subfloat{\includegraphics[width=0.198\linewidth]{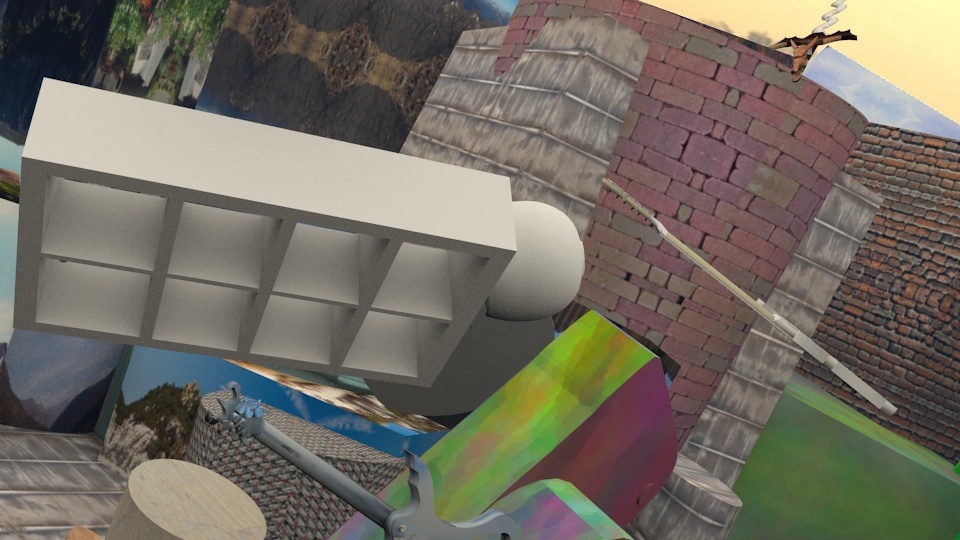}}
    \hfill
    \subfloat{\includegraphics[width=0.198\linewidth]{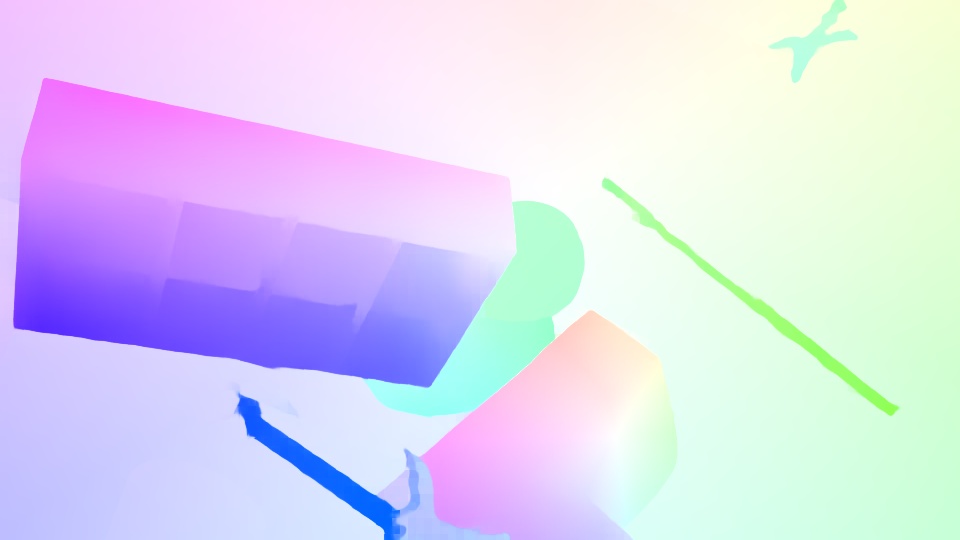}}
    \hfill
    \subfloat{\includegraphics[width=0.198\linewidth]{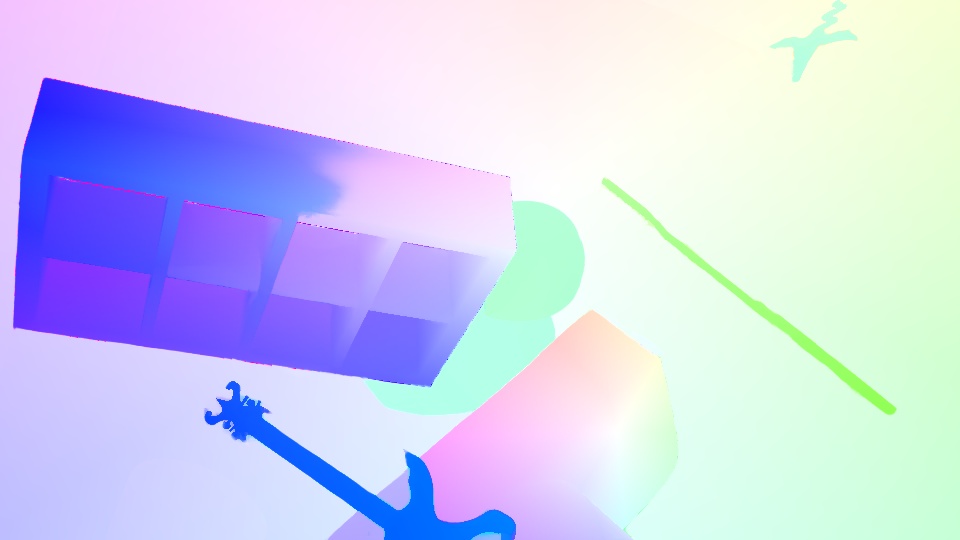}}
    \hfill
    \subfloat{\includegraphics[width=0.198\linewidth]{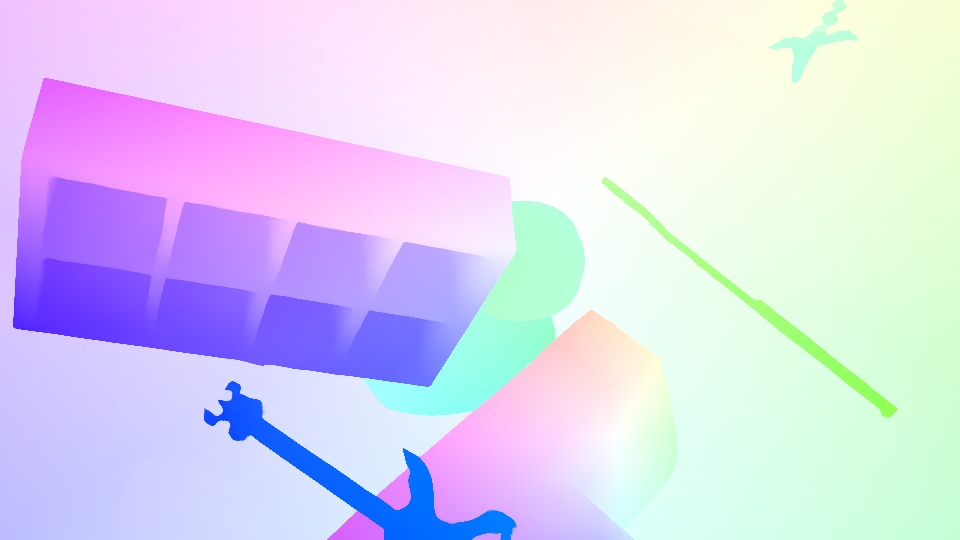}}
    \hfill
    \subfloat{\includegraphics[width=0.198\linewidth]{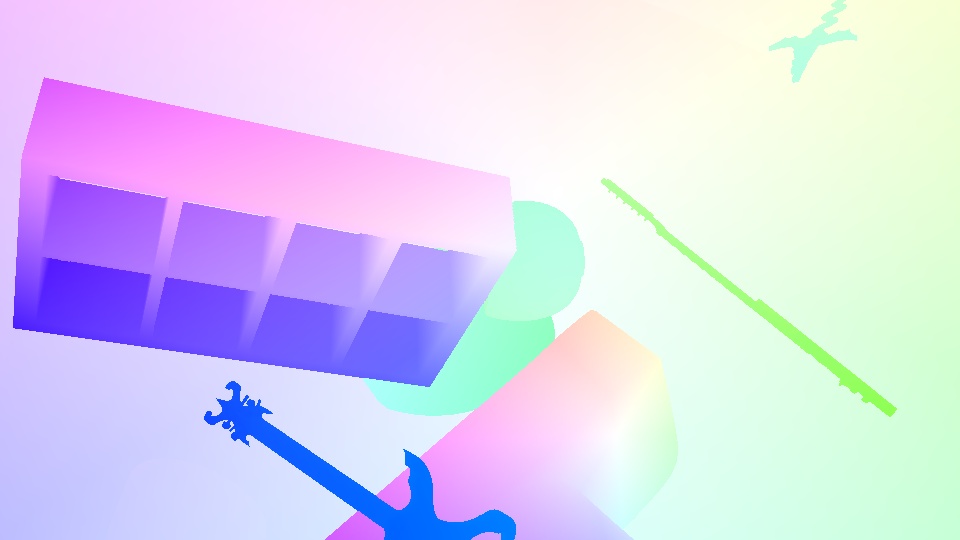}}\\
    \vspace{-4pt}

    \subfloat[PC1 + PC2]{\includegraphics[width=0.198\linewidth]{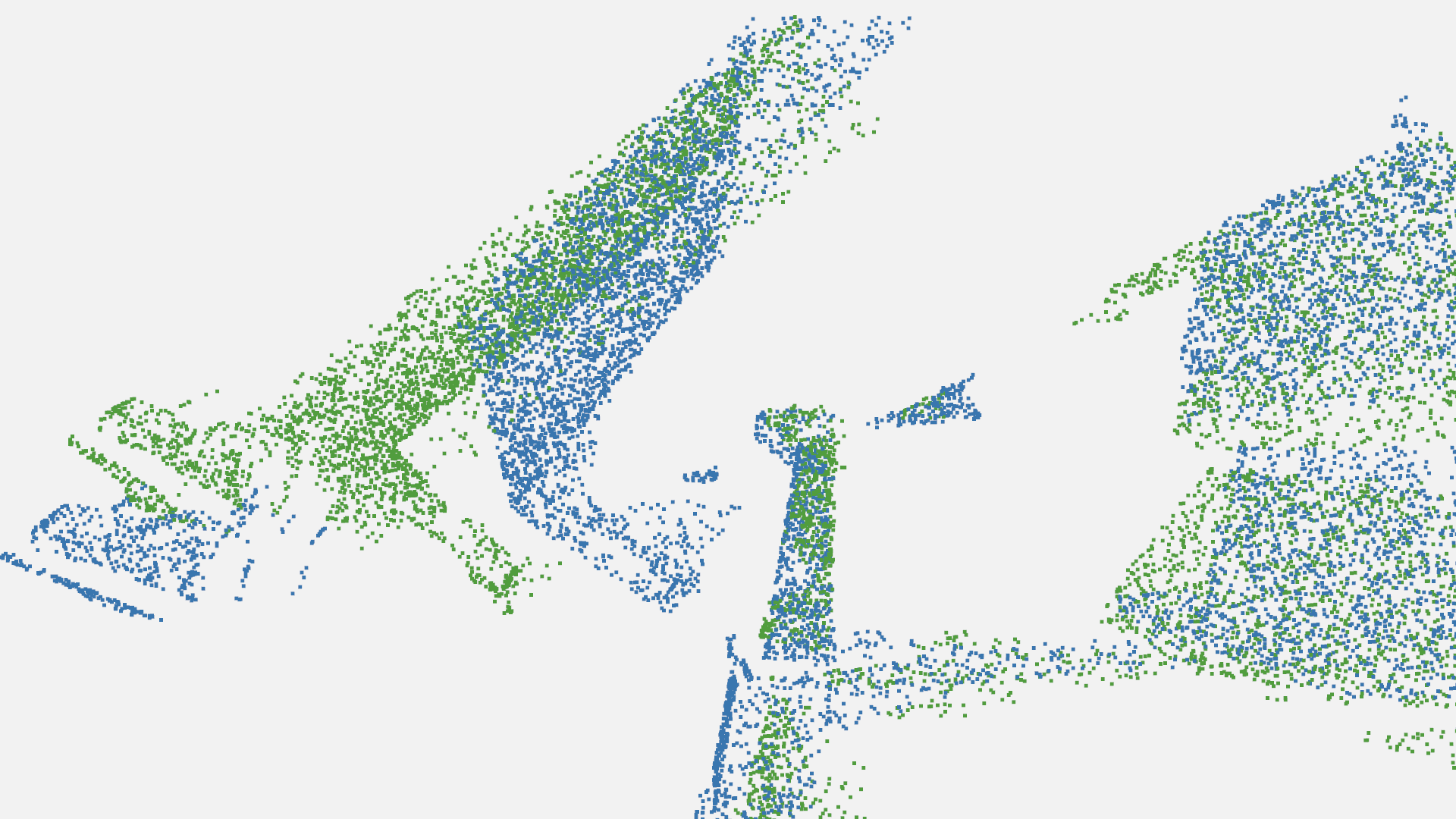}}
    \hfill
    \subfloat[FlowNet3D]{\includegraphics[width=0.198\linewidth]{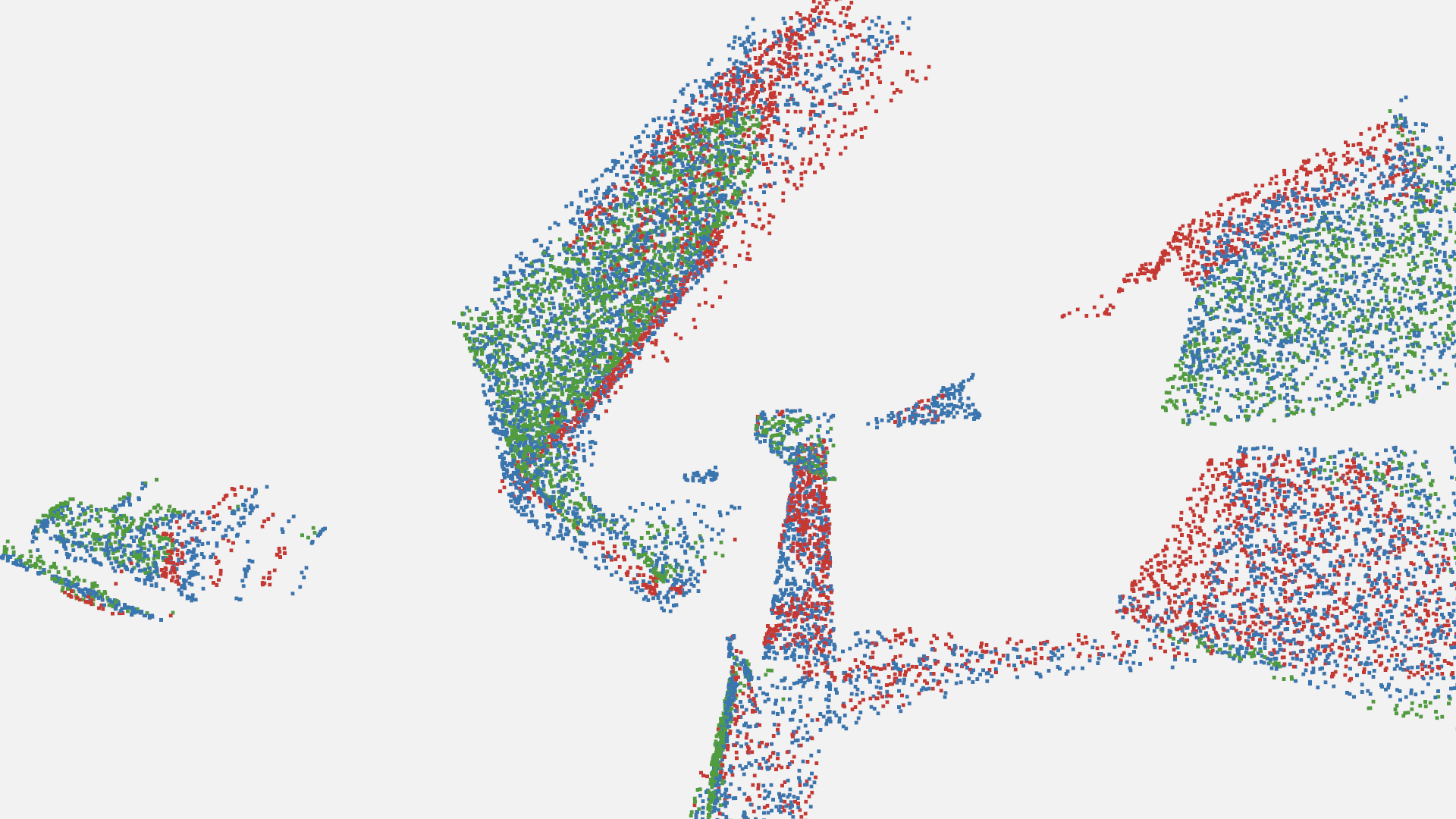}}
    \hfill
    \subfloat[RAFT-3D]{\includegraphics[width=0.198\linewidth]{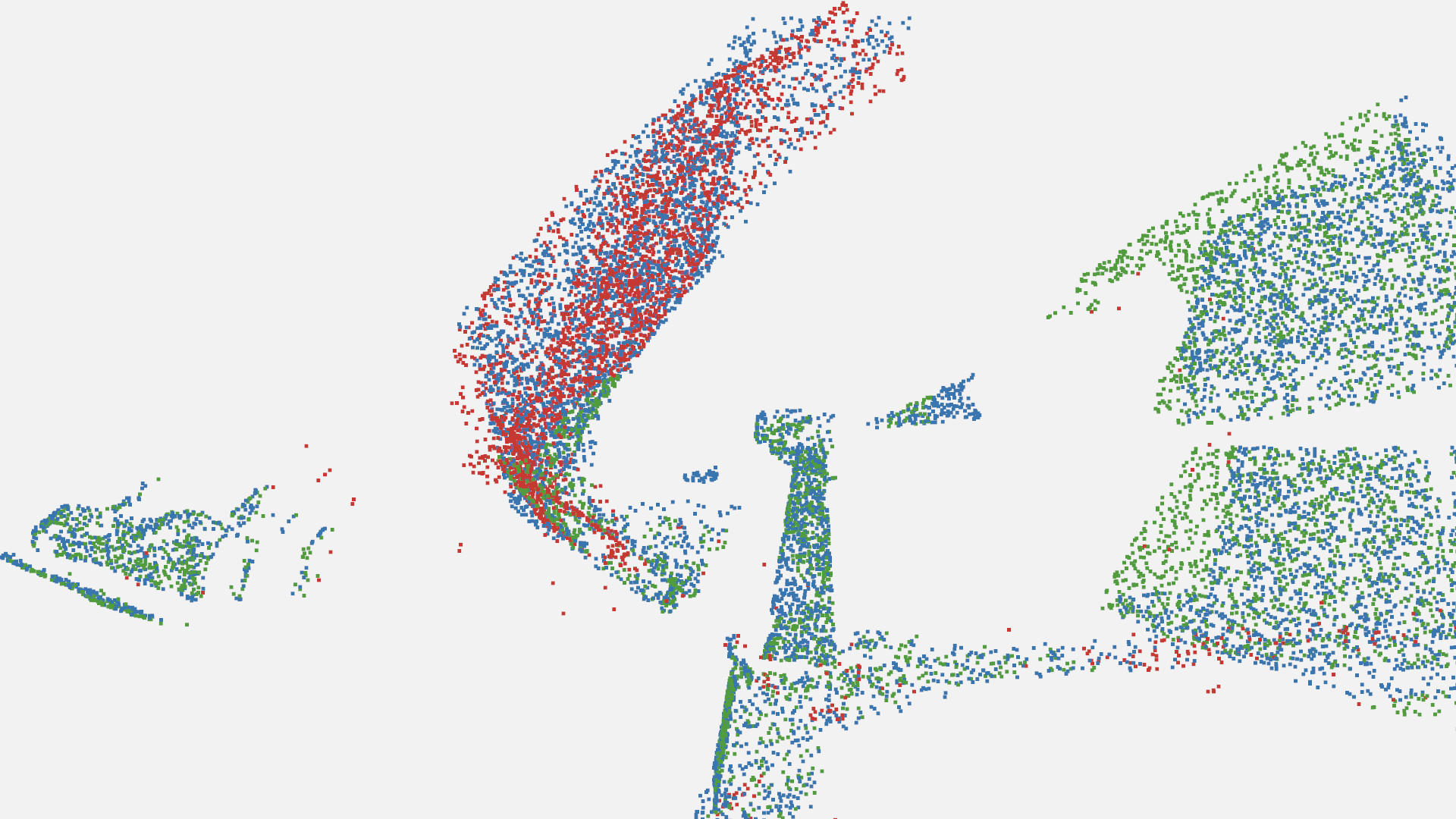}}
    \hfill
    \subfloat[CamLiRAFT (Ours)]{\includegraphics[width=0.198\linewidth]{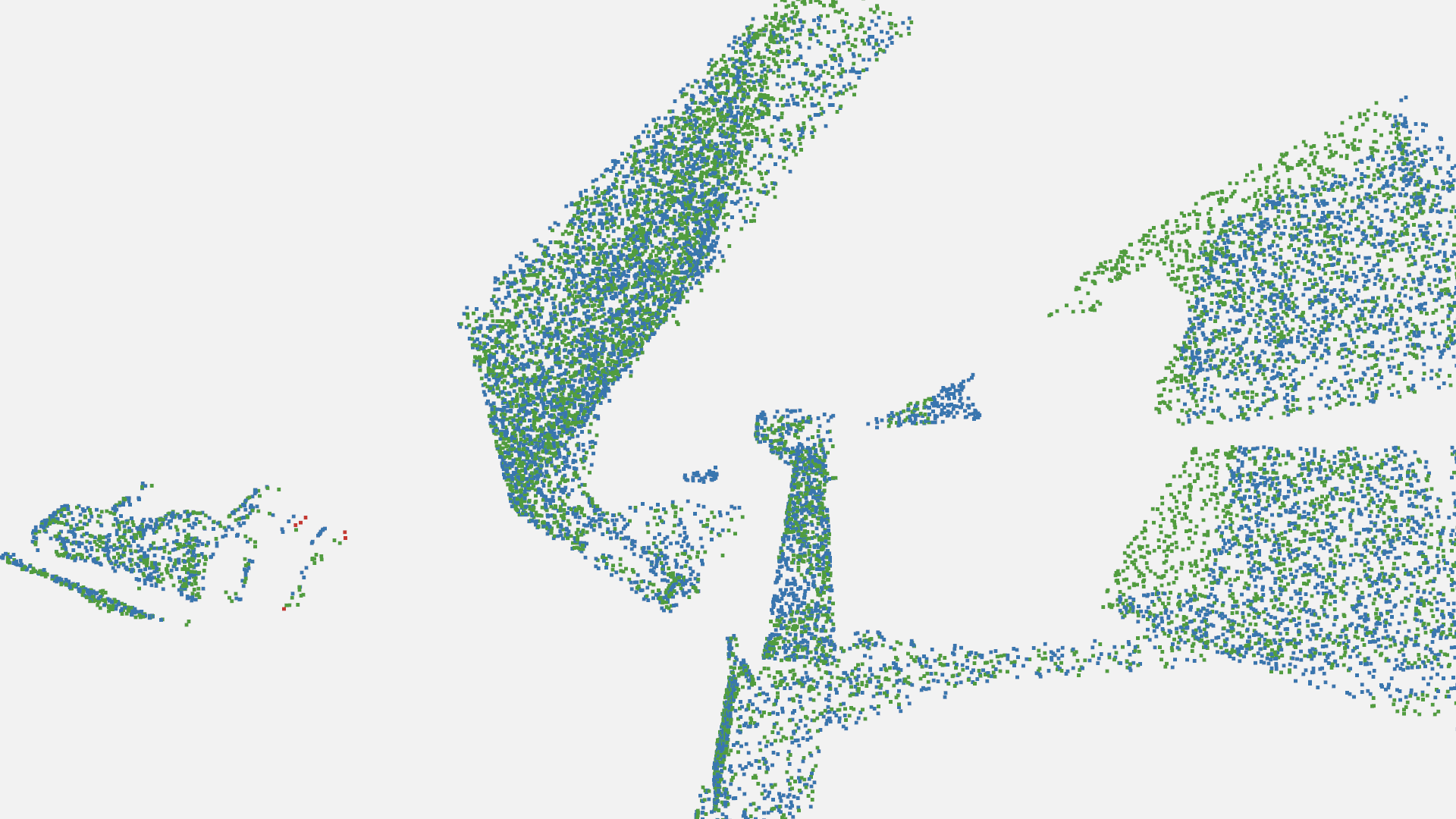}}
    \hfill
    \subfloat[Ground Truth]{\includegraphics[width=0.198\linewidth]{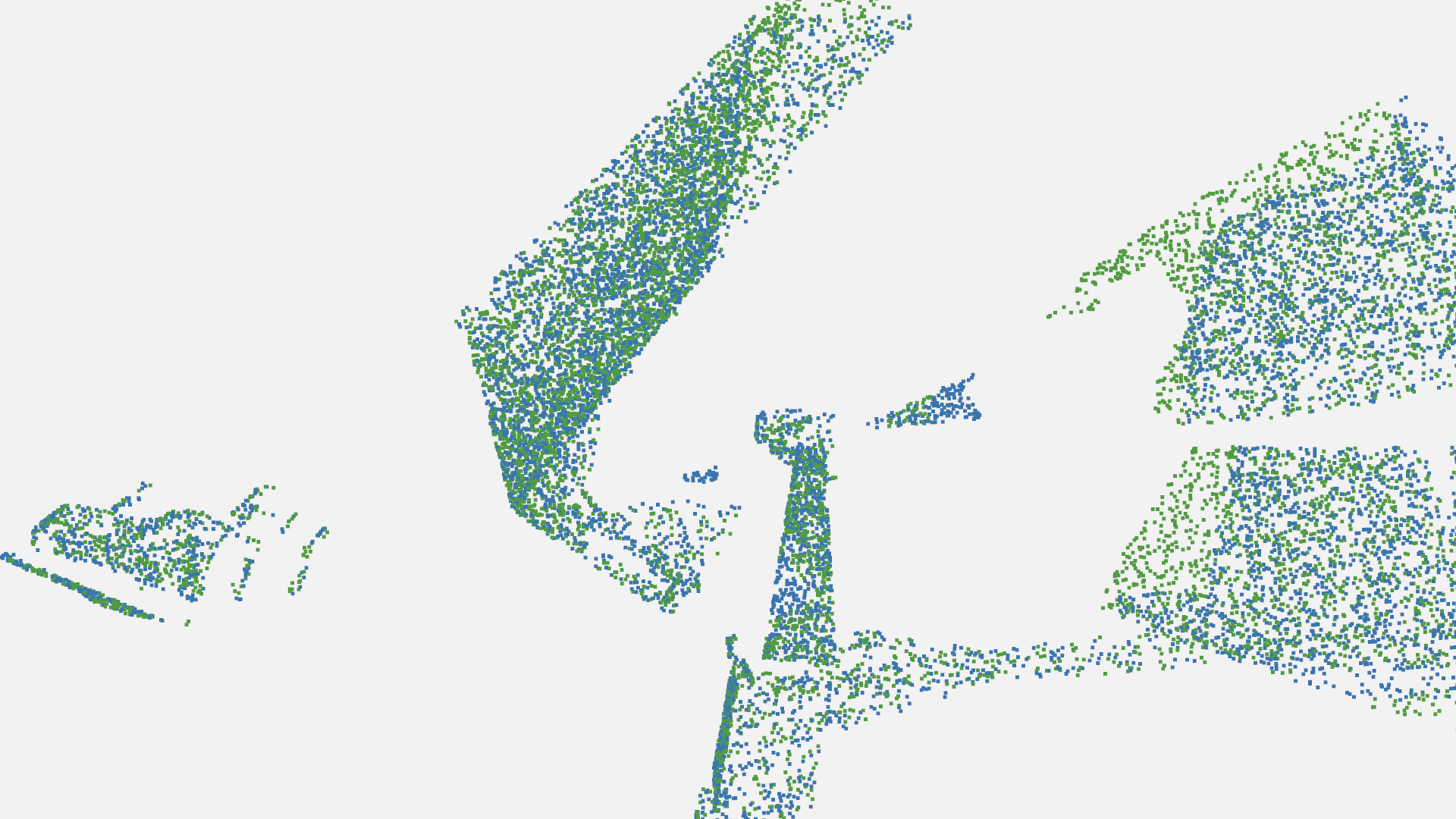}}\\
    \vspace{-9pt}

    \subfloat{\includegraphics[width=0.198\linewidth]{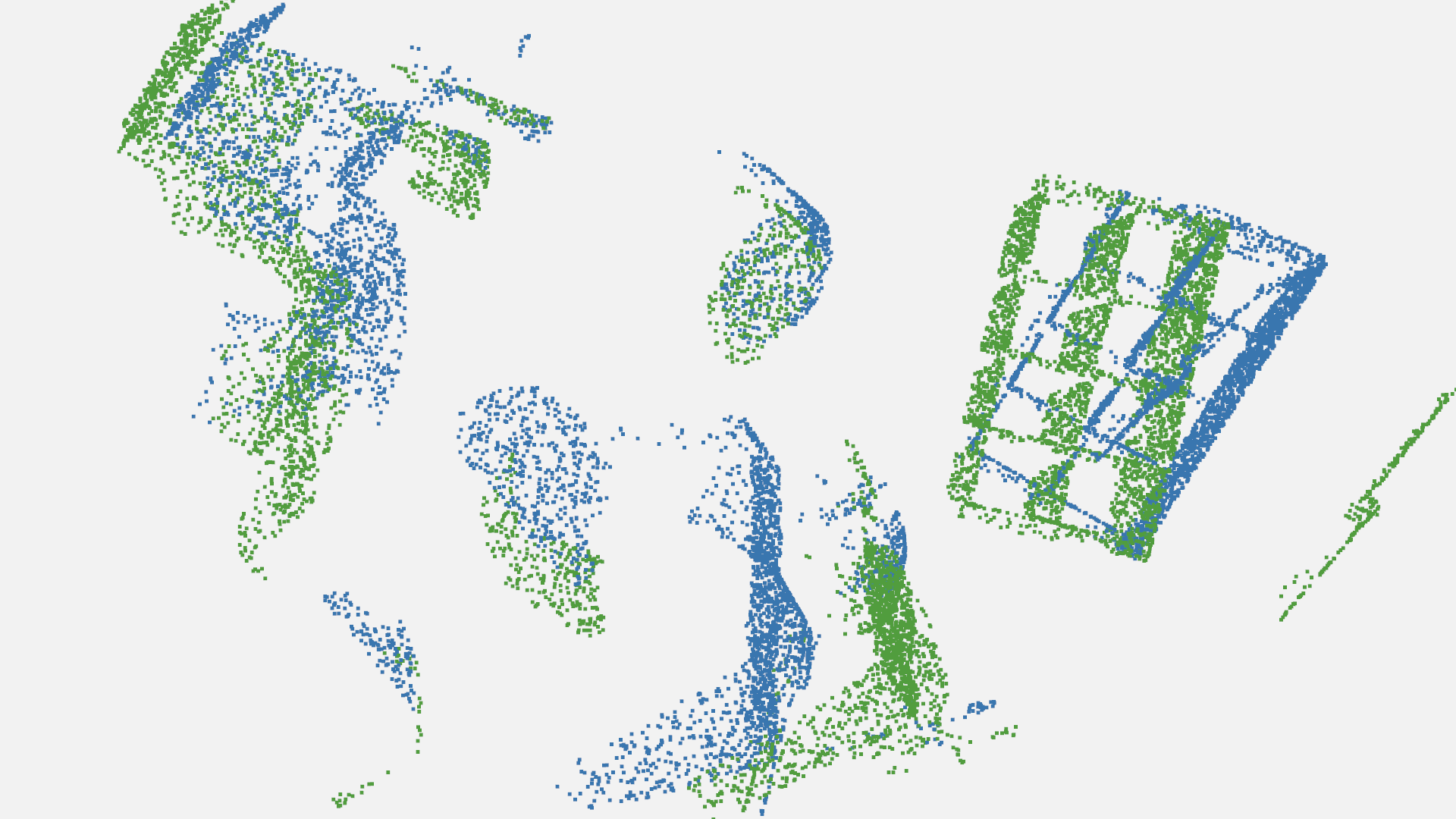}}
    \hfill
    \subfloat{\includegraphics[width=0.198\linewidth]{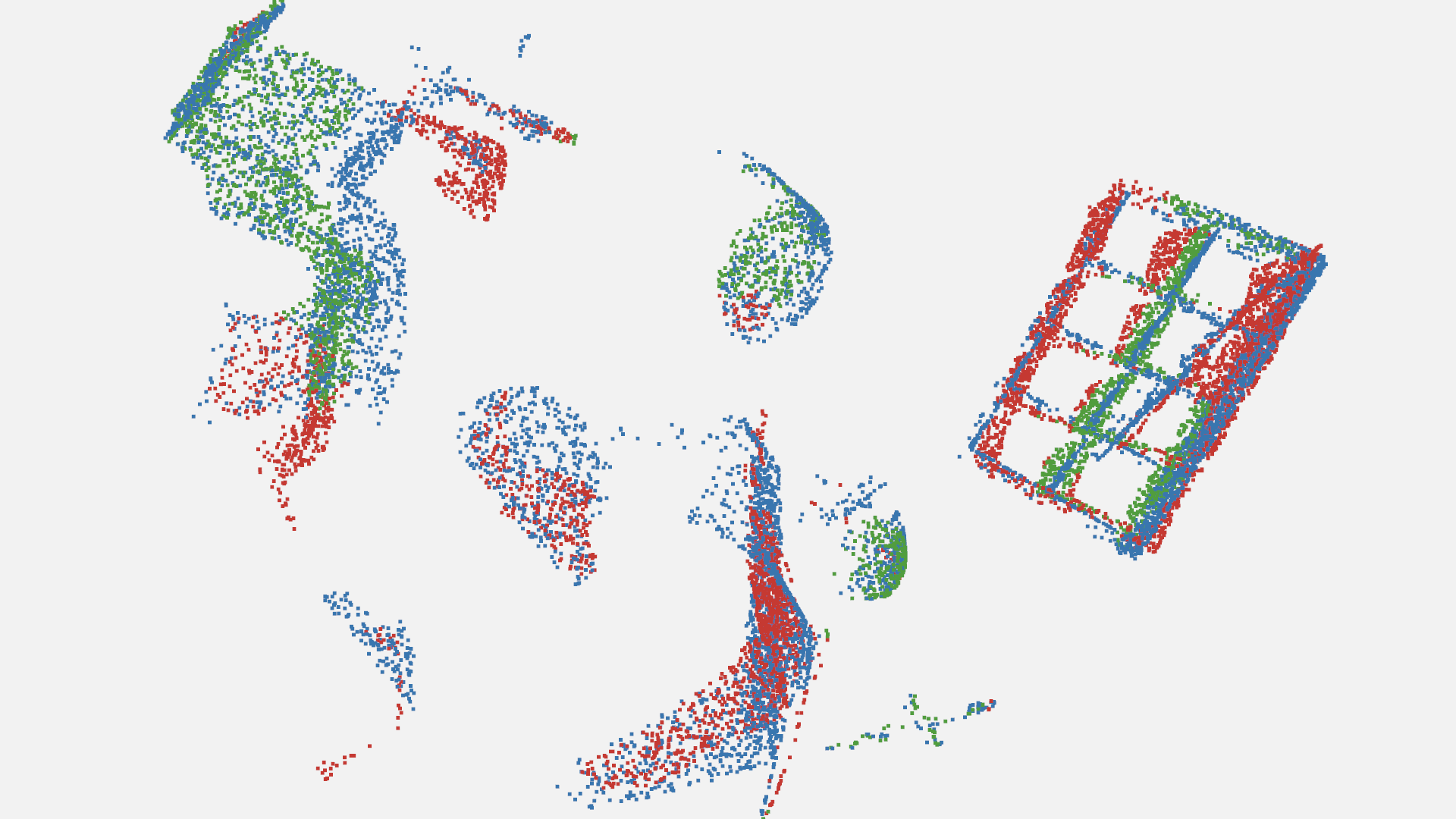}}
    \hfill
    \subfloat{\includegraphics[width=0.198\linewidth]{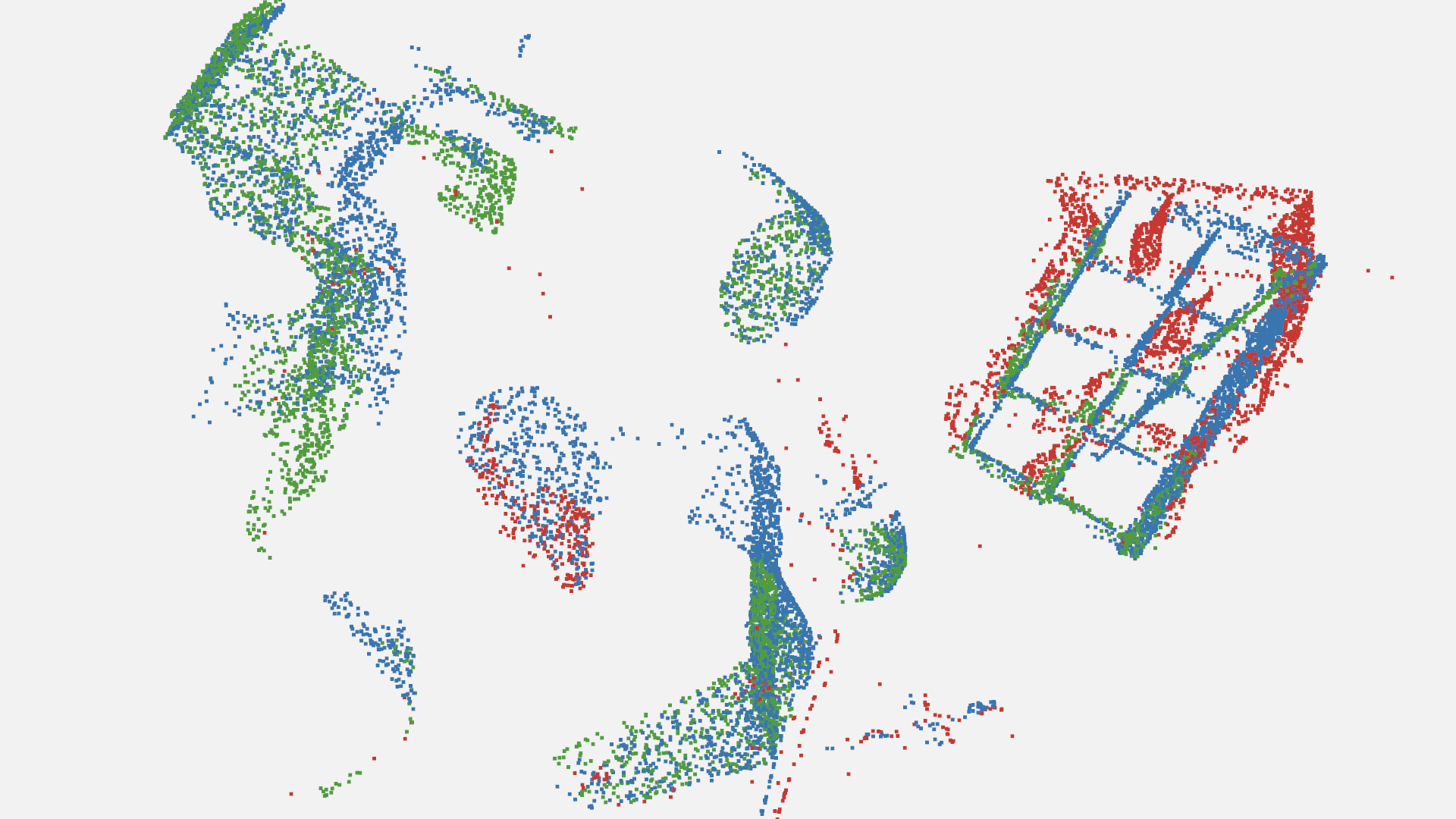}}
    \hfill
    \subfloat{\includegraphics[width=0.198\linewidth]{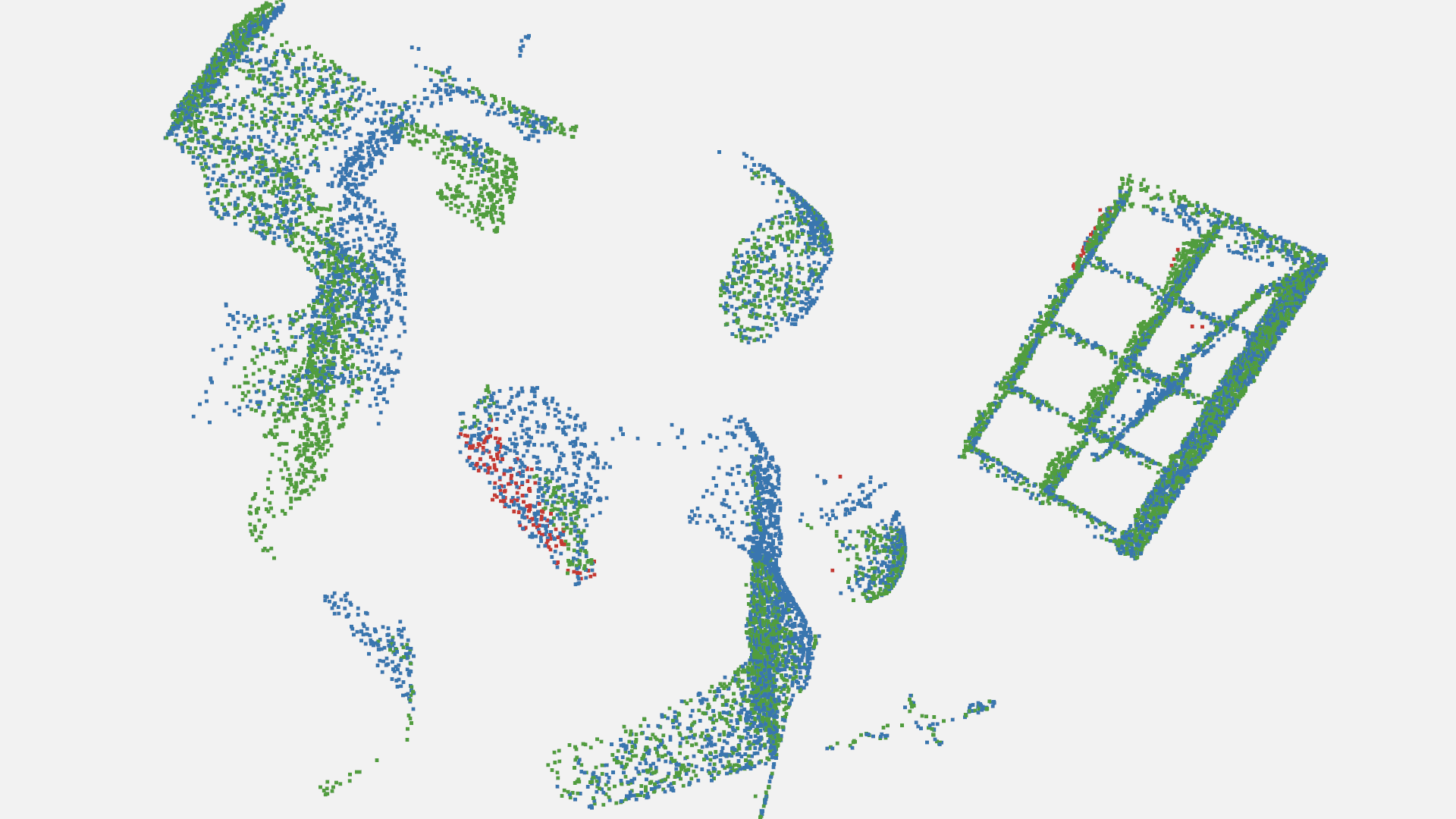}}
    \hfill
    \subfloat{\includegraphics[width=0.198\linewidth]{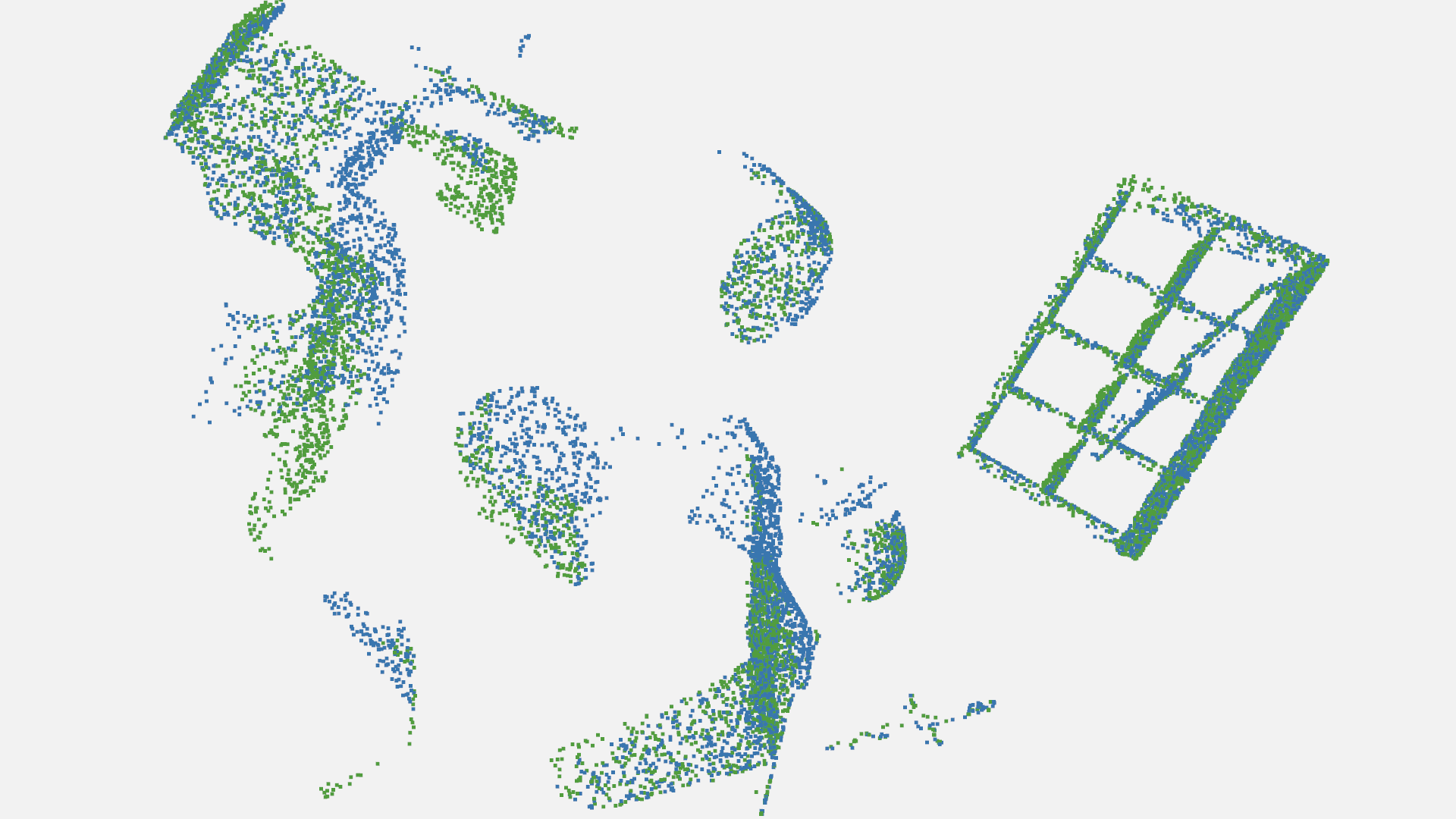}}\\
  
    \caption{Visualized optical flow and scene flow estimation on the ``val'' split of the FlyingThings3D subset. The outliers are marked as \textcolor{red}{red} for scene flow estimation. Our model better handles objects with repetitive structures and texture-less regions.}
    \label{fig:main-things}
    \vspace{-5pt}
\end{figure*}

\subsection{Main Results}

We evaluate our method on the synthetic dataset FlyingThings3D \cite{mayer2016things3d} and the real-world dataset KITTI \cite{menze2015osf}. FlyingThings3D consists of stereo and RGB-D images rendered with multiple randomly moving objects from ShapeNet \cite{chang2015shapenet}, which is large-scale and challenging. KITTI Scene Flow is a real-world benchmark for autonomous driving, consisting of 200 training scenes and 200 test scenes. We also evaluate the generalization performance on MPI Sintel \cite{sintel}, which mainly consists of non-rigid motion.

\subsubsection{FlyingThings3D}

\vspace{4pt} \noindent
\textbf{Data Preprocessing.} Following previous work \cite{ilg2017flownet2, gu2019hplflownet, wu2019pointpwc}, we use the subset of FlyingThings3D. The training and validation set respectively contains 19640 and 3824 pairs of camera-LiDAR frames. We follow FlowNet3D \cite{liu2019flownet3d} instead of HPLFlowNet \cite{gu2019hplflownet} to lift the depth images to point clouds, since HPLFlowNet only keeps non-occluded points which oversimplify the problem. 

\vspace{4pt} \noindent
\textbf{Training.} In Tab. \ref{tab:training-things}, we provide the hyper-parameters and training recipes for CamLiPWC and CamLiRAFT. Here, we use different learning rates for the image and point branch due to the nature of the modality. Cosine decay is adopted as the learning rate schedule. For data augmentation, we only perform random flipping (for both horizontal and vertical directions) since the number of training samples is enough.

\vspace{4pt} \noindent
\textbf{Evaluation Metrics.} We evaluate our network using 2D and 3D end-point error (EPE), as well as threshold metrics (ACC\textsubscript{1px} and ACC\textsubscript{.05}), which measure the portion of error within a threshold. For 2D metrics, we follow RAFT-3D \cite{teed2021raft3d} to evaluate on full images excluding extremely fast-moving regions with flow $>$ 250px. For 3D metrics, we evaluate on all points (including the occluded ones) with depth $<$ 35m. We also include model size in comparison.

\begin{table*}
  \caption{Leaderboard of KITTI Scene Flow. Even the non-rigid CamLiRAFT performs on par with the previous state-of-the-art method. By refining the scene flow with the priors of rigidity (denoted by $\dagger$), CamLiRAFT achieves an error of {\bf4.26\%}, ranking 1st on the leaderboard.}
  \vspace{-5pt}
  \label{tab:main-kitti}
  \centering
  \small
  \renewcommand{\arraystretch}{1.15}
  \begin{tabular}{l|c|cc|cc|cc|cc|c}
  \hline
  \multirow{2}{*}{Method} & \multirow{2}{*}{Rigidity} & \multicolumn{2}{c|}{D1 (\%)} & \multicolumn{2}{c|}{D2 (\%)} & \multicolumn{2}{c|}{OF (\%)} & \multicolumn{2}{c|}{SF (\%)} & \multirow{2}{*}{Parameters} \\
  & & noc & all & noc & all & noc & all & noc & all & \\
  \hline
  SENSE $\dagger$ \cite{jiang2019sense} & Background & 2.05 & 2.22 & 3.87 & 5.89 & 5.98 & 7.64 & 7.30 & 9.55 & 13.4M \\
  Binary TTC \cite{badki2021binaryttc} & None & \textbf{1.63} & \textbf{1.81} & 2.72 & 4.76 & 3.89 & 6.31 & 5.29 & 8.50 & - \\
  OpticalExp \cite{yang2020opticalexp} & None & \textbf{1.63} & \textbf{1.81} & 2.62 & 4.25 & 3.89 & 6.30 & 5.21 & 8.12 & 18.5M \\
  ISF $\dagger$ \cite{behl2017isf} & Full Image & 4.02 & 4.46 & 4.95 & 5.95 & 4.69 & 6.22 & 6.45 & 8.08 & - \\
  ACOSF $\dagger$ \cite{li2021acosf} & Full Image & 3.35 & 3.58 & 4.26 & 5.31 & 4.51 & 5.79 & 6.40 & 7.90 & - \\
  DRISF $\dagger$ \cite{ma2019drisf} & Full Image & 2.35 & 2.55 & 3.14 & 4.04 & 3.58 & 4.73 & 4.99 & 6.31 & 58.9M \\
  RAFT-3D \cite{teed2021raft3d} & Full Image & \textbf{1.63} & \textbf{1.81} & 2.67 & 3.67 & 3.23 & 4.29 & 4.53 & 5.77 & 51.3M \\
  RigidMask $\dagger$ \cite{yang2021rigidmask} & Full Image & 1.70 & 1.89 & 2.47 & 3.23 & 2.54 & 3.50 & 3.73 & 4.89 & 145.3M \\
  \hline
  CamLiFlow \cite{liu2022camliflow} & None & \textbf{1.63} & \textbf{1.81} & 2.39 & 3.19 & 2.77 & 4.05 & 4.03 & 5.62 & 14.0M \\
  CamLiFlow $\dagger$ \cite{liu2022camliflow} & Background & \textbf{1.63} & \textbf{1.81} & {2.37} & {2.95} & {2.40} & {3.10} & {3.55} & {4.43} & 19.7M \\
  \hline
  \textbf{CamLiRAFT} & None & \textbf{1.63} & \textbf{1.81} & 2.35 & 3.02 & 2.38 & 3.43 & 3.70 & 4.97 & 14.7M \\
  \textbf{CamLiRAFT} $\dagger$ & Background & \textbf{1.63} & \textbf{1.81} & \textbf{2.33} & \textbf{2.94} & \textbf{2.18} & \textbf{2.96} & \textbf{3.33} & \textbf{4.26} & 20.4M \\
  \hline
  \end{tabular}
  \vspace{-12pt}
\end{table*}

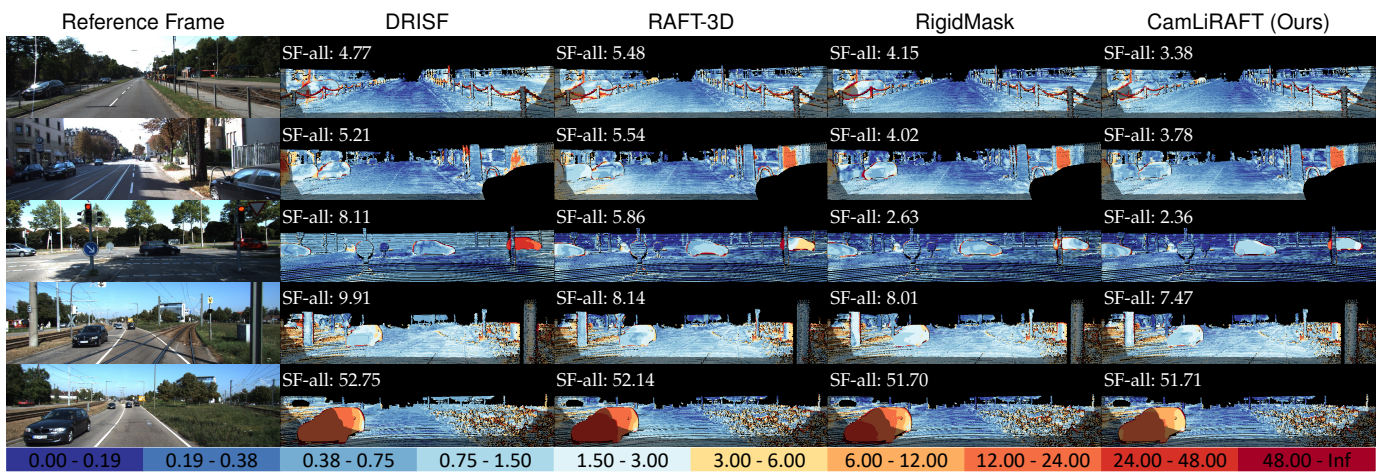
\begin{figure*}
  \centering
  \captionsetup[subfigure]{labelformat=empty,position=top}
  \captionsetup[subfloat]{captionskip=1.5pt}

  \subfloat[Reference Frame]{\begin{picture}(0.2\linewidth,30)%
    \includegraphics[width=0.2\linewidth]{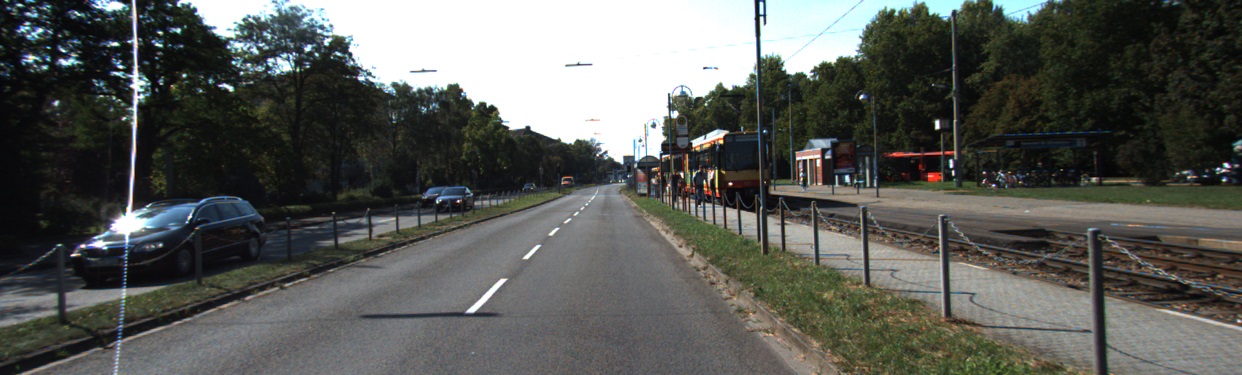}%
  \end{picture}}
  \subfloat[DRISF]{\begin{picture}(0.2\linewidth,30)
    \put(0,0){\includegraphics[width=0.2\linewidth]{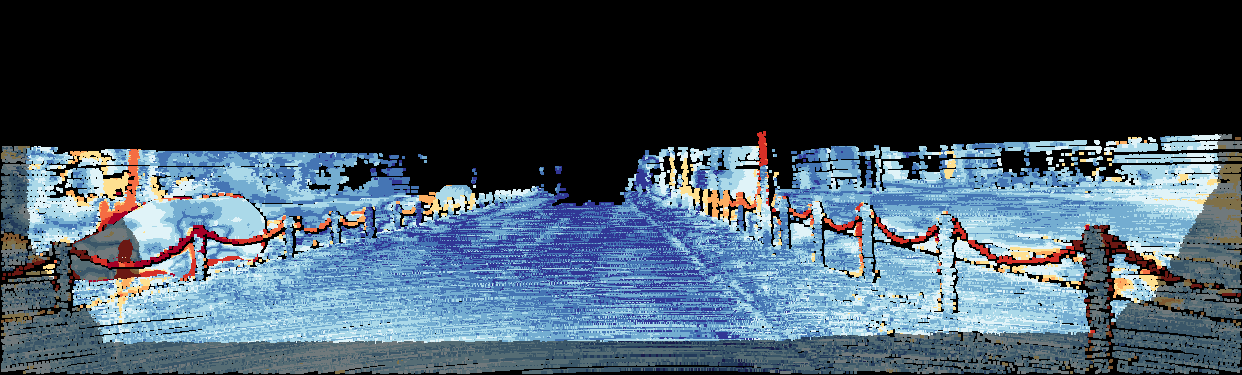}}
    \put(1,23){\scriptsize \textcolor{white}{SF-all: 4.77}}
  \end{picture}}
  \subfloat[RAFT-3D]{\begin{picture}(0.2\linewidth,30)
    \put(0,0){\includegraphics[width=0.2\linewidth]{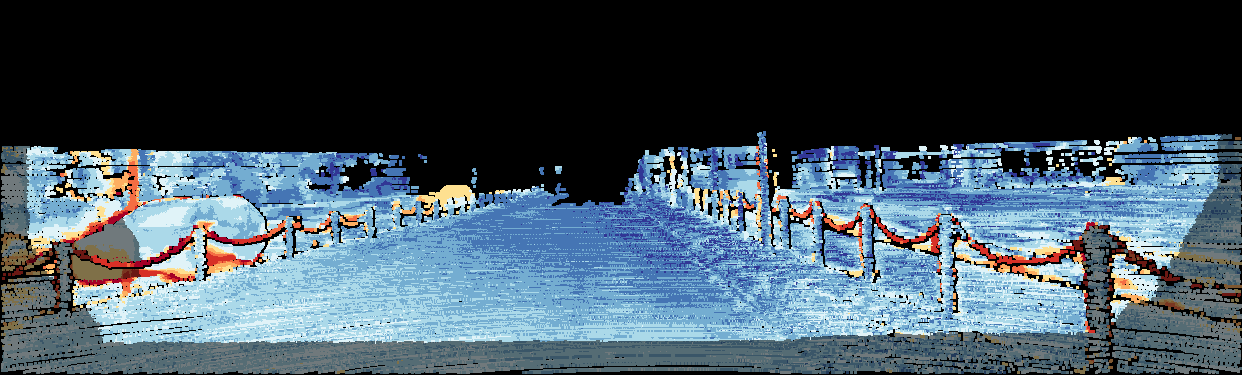}}
    \put(1,23){\scriptsize \textcolor{white}{SF-all: 5.48}}
  \end{picture}}
  \subfloat[RigidMask]{\begin{picture}(0.2\linewidth,30)
    \put(0,0){\includegraphics[width=0.2\linewidth]{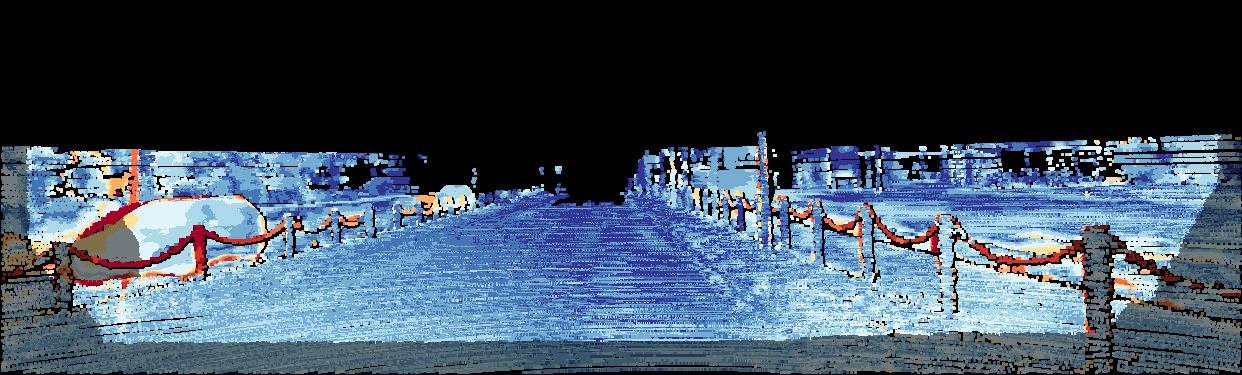}}
    \put(1,23){\scriptsize \textcolor{white}{SF-all: 4.15}}
  \end{picture}}
  \subfloat[CamLiRAFT (Ours)]{\begin{picture}(0.2\linewidth,30)
    \put(0,0){\includegraphics[width=0.2\linewidth]{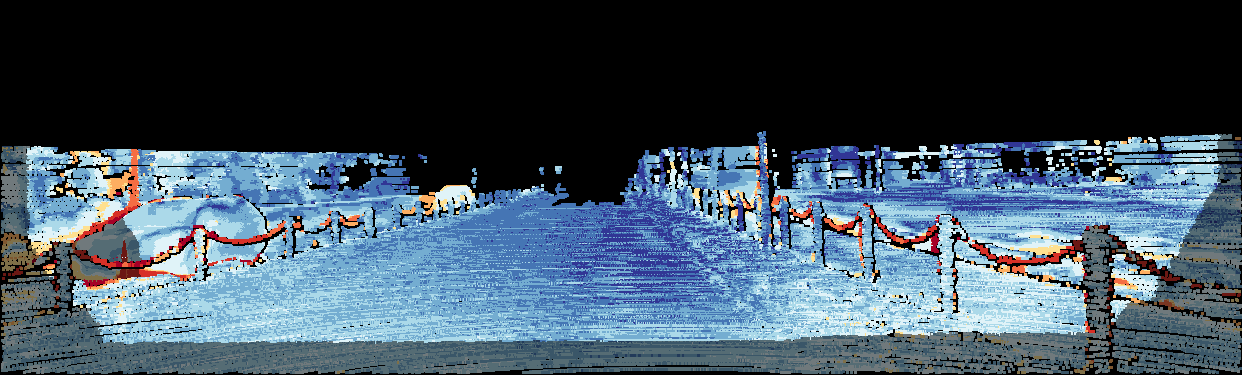}}
    \put(1,23){\scriptsize \textcolor{white}{SF-all: 3.38}}
  \end{picture}}\\
  \vspace{-9pt}

  \subfloat{\begin{picture}(0.2\linewidth,30)
    \includegraphics[width=0.2\linewidth]{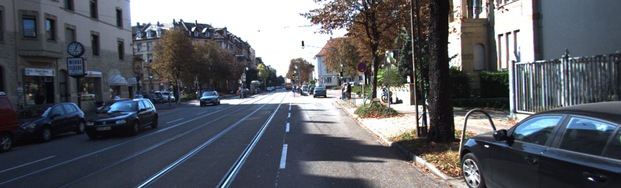}
  \end{picture}}
  \subfloat{\begin{picture}(0.2\linewidth,30)
    \put(0,0){\includegraphics[width=0.2\linewidth]{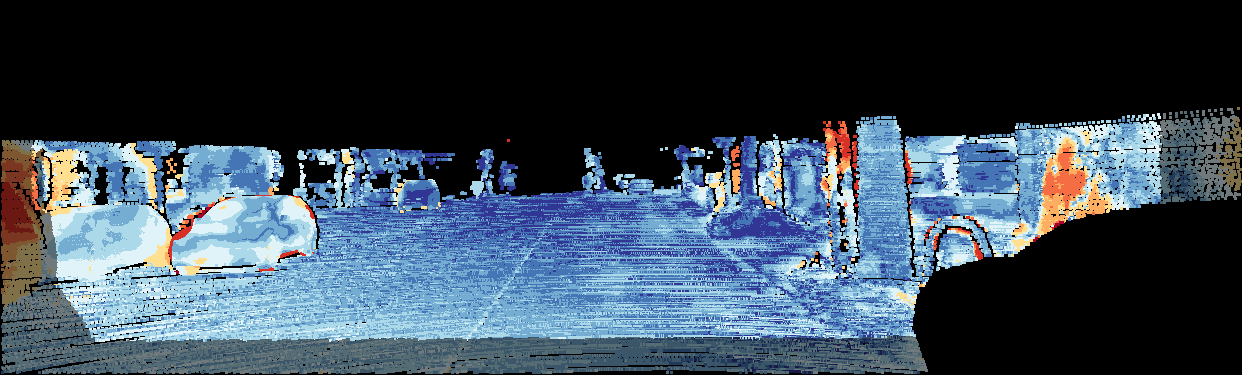}}
    \put(1,23){\scriptsize \textcolor{white}{SF-all: 5.21}}
  \end{picture}}
  \subfloat{\begin{picture}(0.2\linewidth,30)
    \put(0,0){\includegraphics[width=0.2\linewidth]{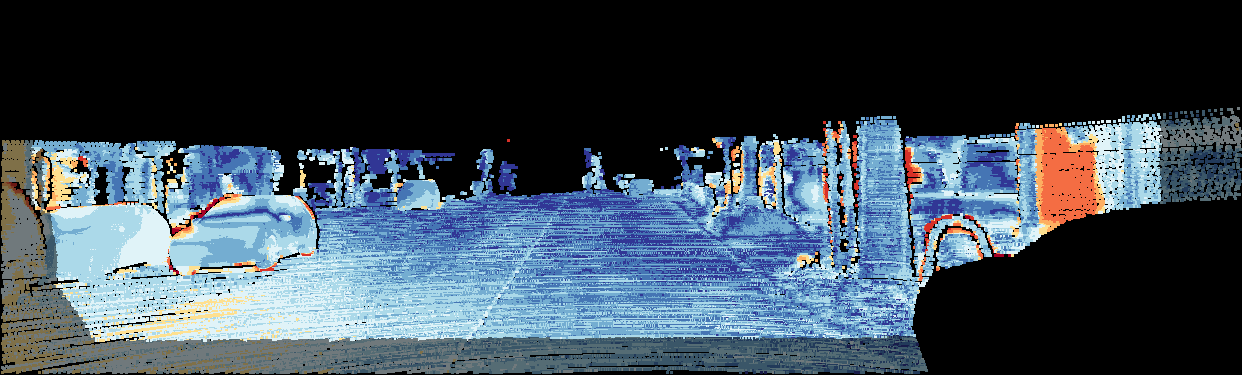}}
    \put(1,23){\scriptsize \textcolor{white}{SF-all: 5.54}}
  \end{picture}}
  \subfloat{\begin{picture}(0.2\linewidth,30)
    \put(0,0){\includegraphics[width=0.2\linewidth]{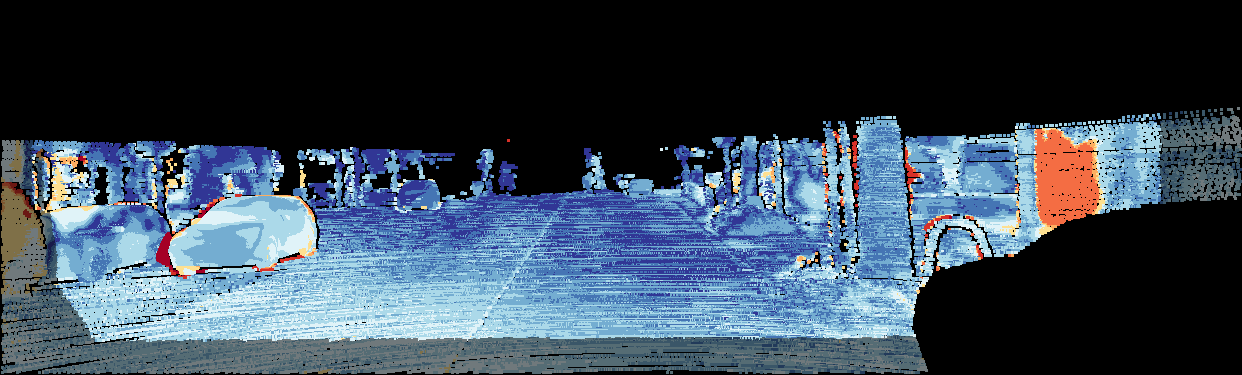}}
    \put(1,23){\scriptsize \textcolor{white}{SF-all: 4.02}}
  \end{picture}}
  \subfloat{\begin{picture}(0.2\linewidth,30)
    \put(0,0){\includegraphics[width=0.2\linewidth]{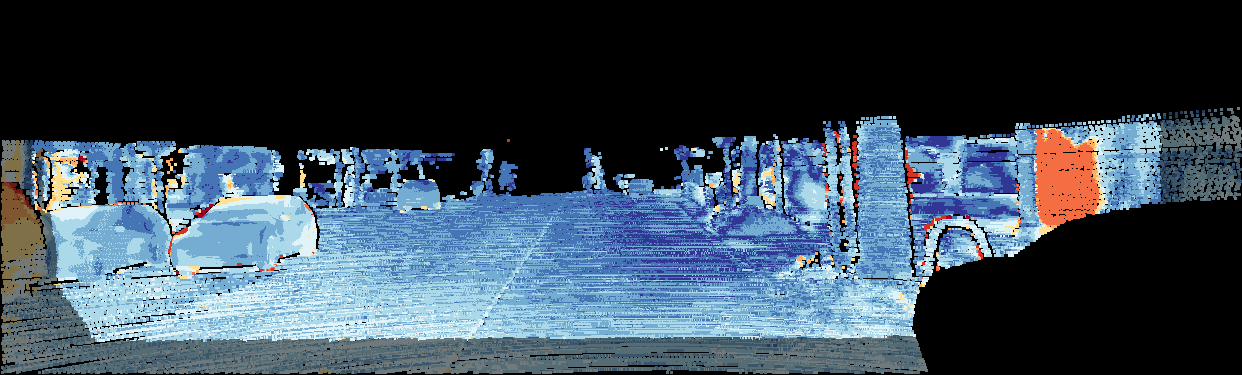}}
    \put(1,23){\scriptsize \textcolor{white}{SF-all: 3.78}}
  \end{picture}}\\
  \vspace{-9pt}

  \subfloat{\begin{picture}(0.2\linewidth,30)
    \includegraphics[width=0.2\linewidth]{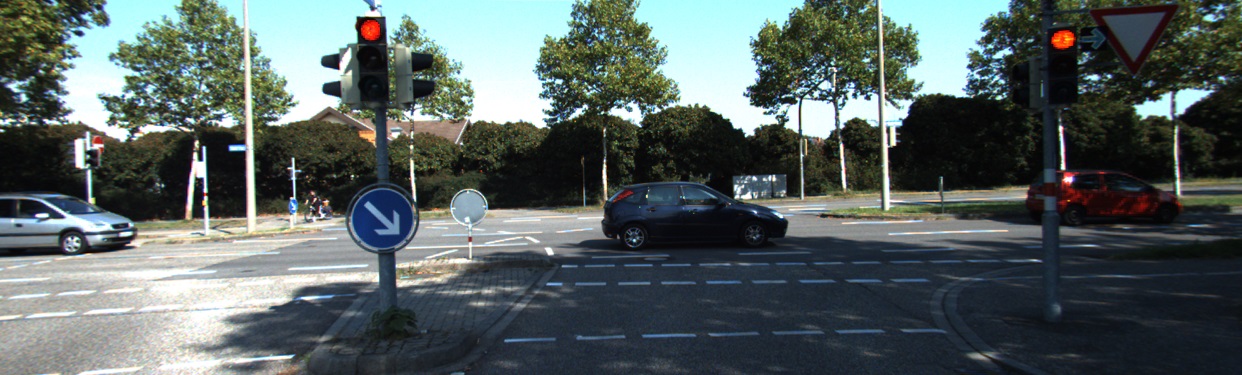}
  \end{picture}}
  \subfloat{\begin{picture}(0.2\linewidth,30)
    \put(0,0){\includegraphics[width=0.2\linewidth]{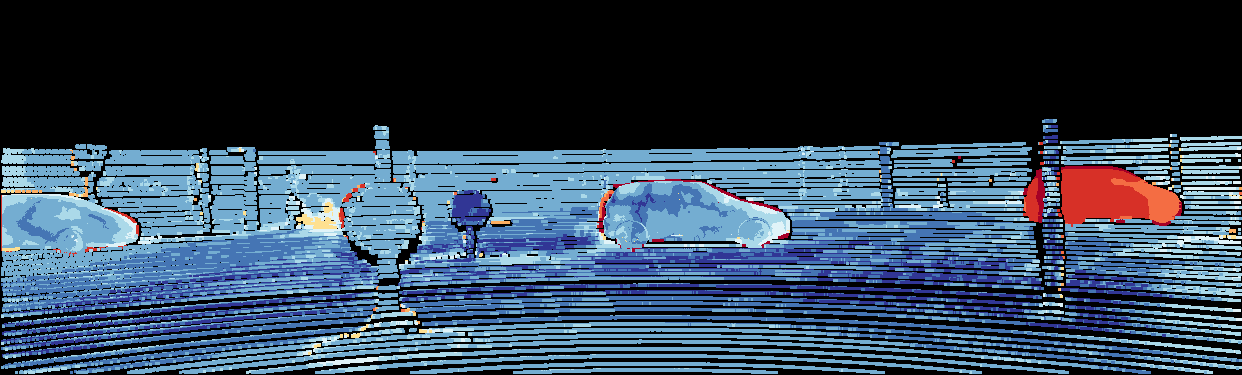}}
    \put(1,23){\scriptsize \textcolor{white}{SF-all: 8.11}}
  \end{picture}}
  \subfloat{\begin{picture}(0.2\linewidth,30)
    \put(0,0){\includegraphics[width=0.2\linewidth]{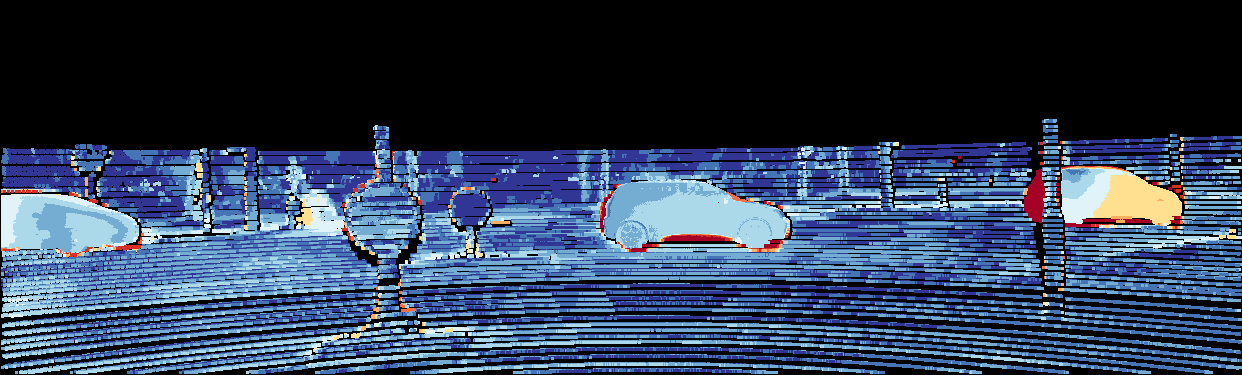}}
    \put(1,23){\scriptsize \textcolor{white}{SF-all: 5.86}}
  \end{picture}}
  \subfloat{\begin{picture}(0.2\linewidth,30)
    \put(0,0){\includegraphics[width=0.2\linewidth]{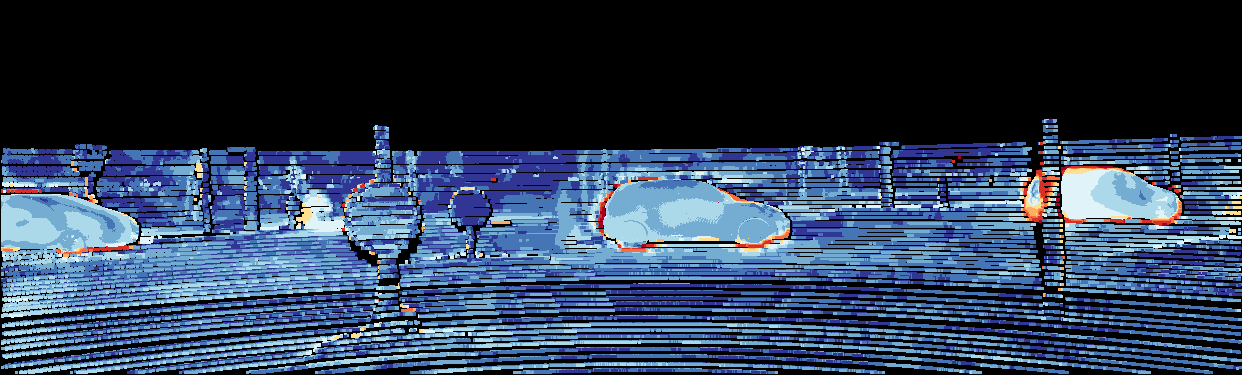}}
    \put(1,23){\scriptsize \textcolor{white}{SF-all: 2.63}}
  \end{picture}}
  \subfloat{\begin{picture}(0.2\linewidth,30)
    \put(0,0){\includegraphics[width=0.2\linewidth]{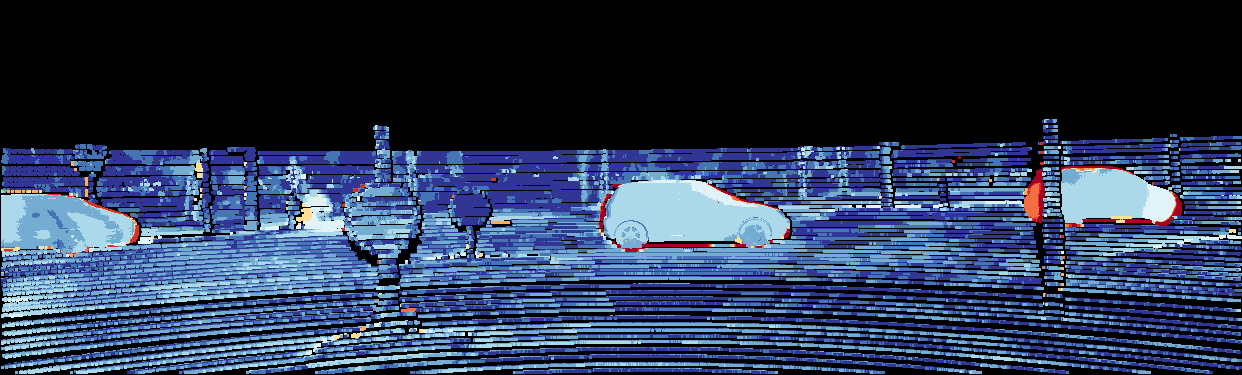}}
    \put(1,23){\scriptsize \textcolor{white}{SF-all: 2.36}}
  \end{picture}}\\
  \vspace{-9pt}

  \subfloat{\begin{picture}(0.2\linewidth,30)
    \includegraphics[width=0.2\linewidth]{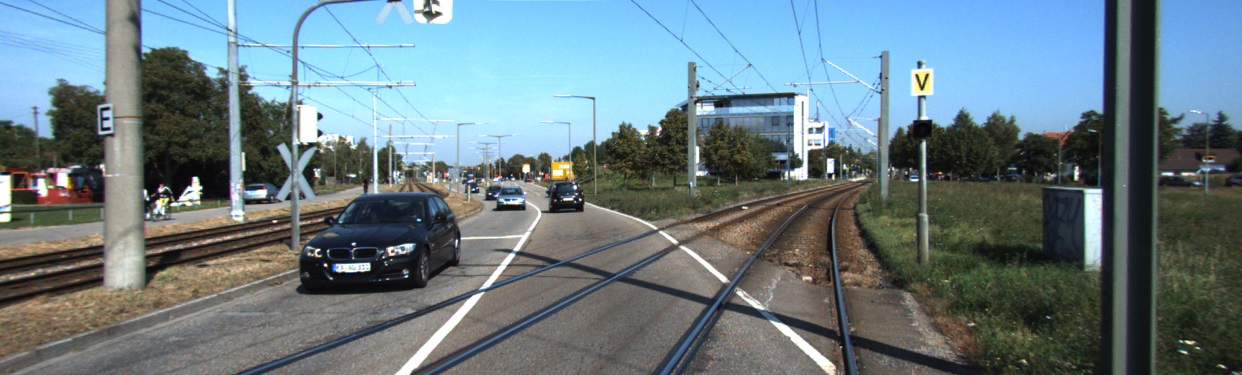}
  \end{picture}}
  \subfloat{\begin{picture}(0.2\linewidth,30)
    \put(0,0){\includegraphics[width=0.2\linewidth]{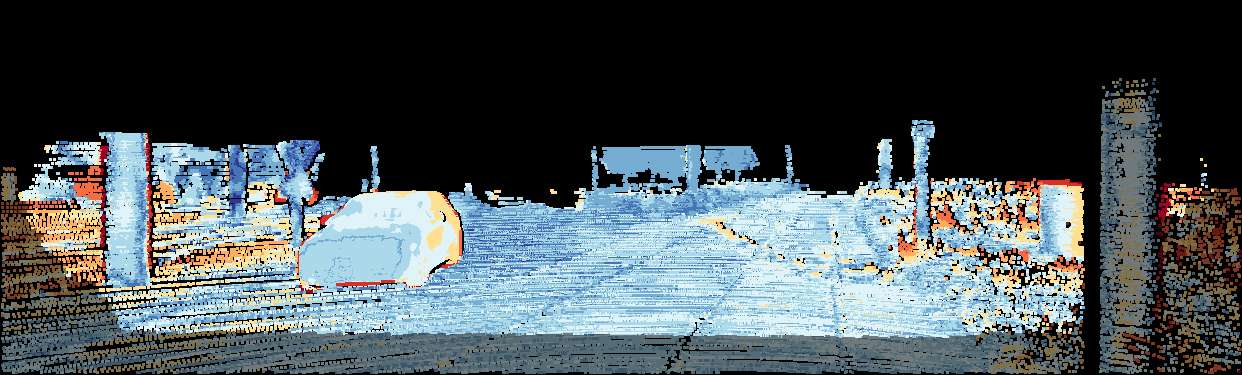}}
    \put(1,23){\scriptsize \textcolor{white}{SF-all: 9.91}}
  \end{picture}}
  \subfloat{\begin{picture}(0.2\linewidth,30)
    \put(0,0){\includegraphics[width=0.2\linewidth]{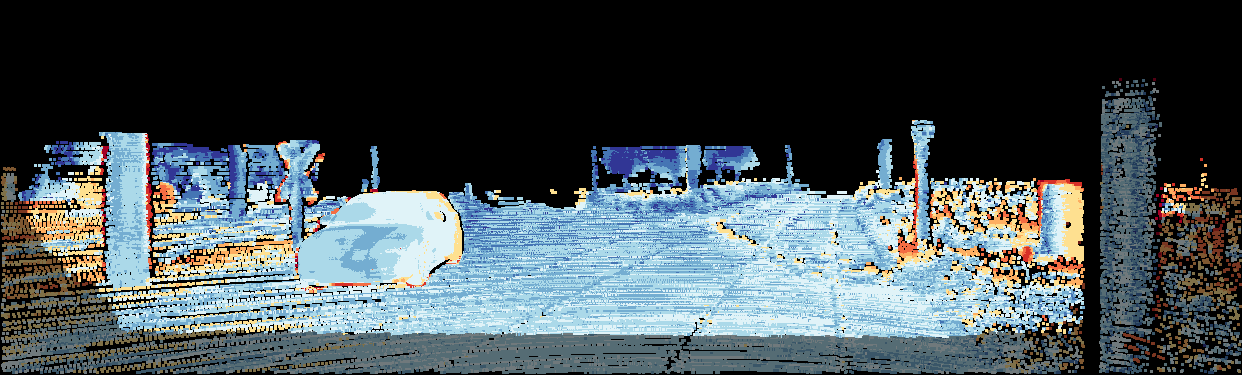}}
    \put(1,23){\scriptsize \textcolor{white}{SF-all: 8.14}}
  \end{picture}}
  \subfloat{\begin{picture}(0.2\linewidth,30)
    \put(0,0){\includegraphics[width=0.2\linewidth]{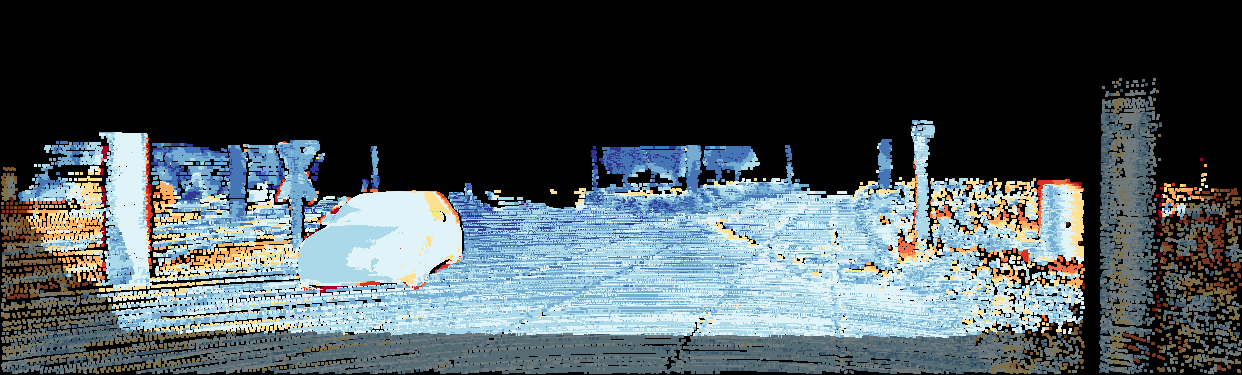}}
    \put(1,23){\scriptsize \textcolor{white}{SF-all: 8.01}}
  \end{picture}}
  \subfloat{\begin{picture}(0.2\linewidth,30)
    \put(0,0){\includegraphics[width=0.2\linewidth]{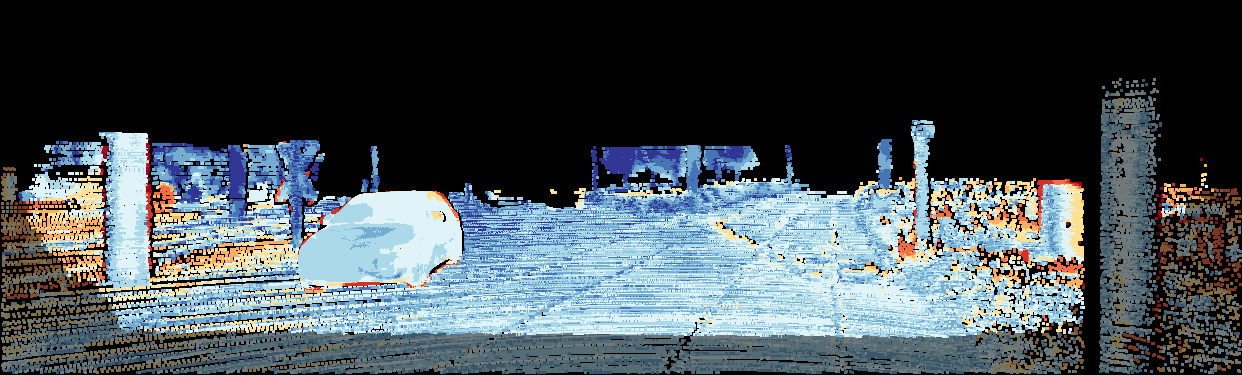}}
    \put(1,23){\scriptsize \textcolor{white}{SF-all: 7.47}}
  \end{picture}}\\
  \vspace{-9pt}

  \subfloat{\begin{picture}(0.2\linewidth,30)
    \includegraphics[width=0.2\linewidth]{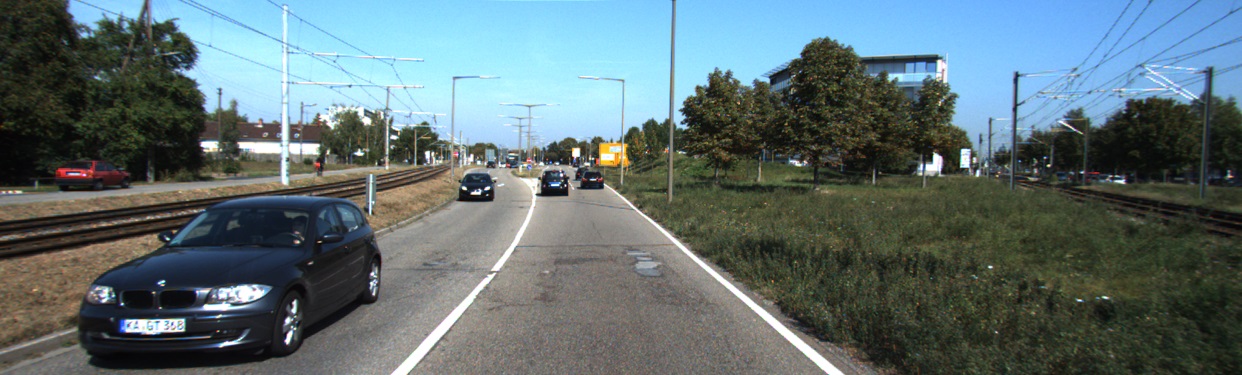}
  \end{picture}}
  \subfloat{\begin{picture}(0.2\linewidth,30)
    \put(0,0){\includegraphics[width=0.2\linewidth]{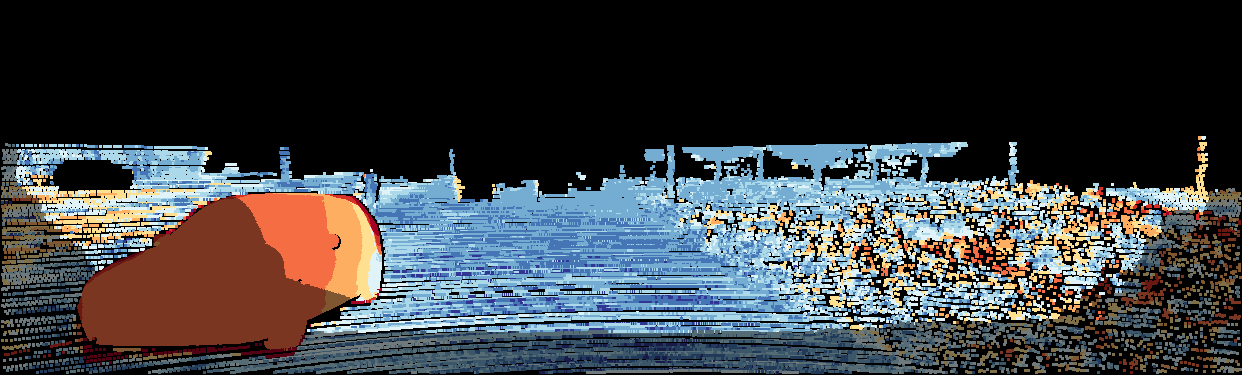}}
    \put(1,23){\scriptsize \textcolor{white}{SF-all: 52.75}}
  \end{picture}}
  \subfloat{\begin{picture}(0.2\linewidth,30)
    \put(0,0){\includegraphics[width=0.2\linewidth]{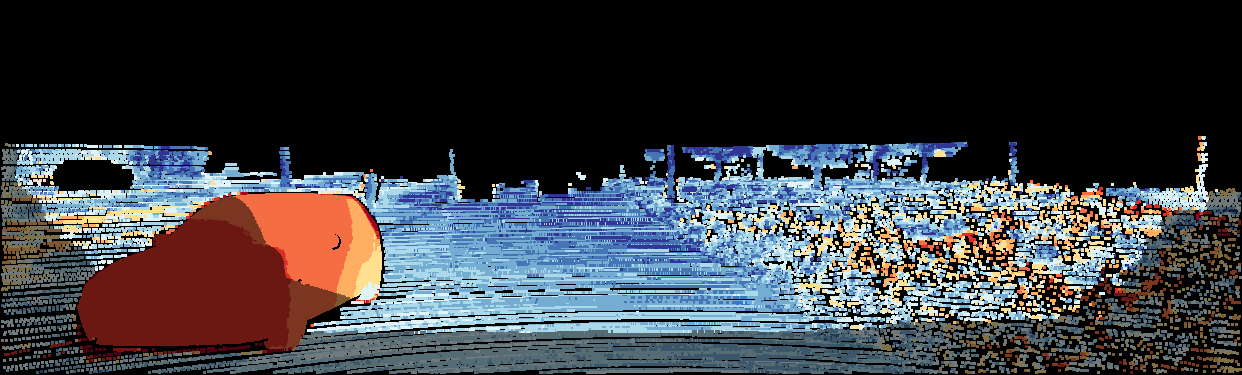}}
    \put(1,23){\scriptsize \textcolor{white}{SF-all: 52.14}}
  \end{picture}}
  \subfloat{\begin{picture}(0.2\linewidth,30)
    \put(0,0){\includegraphics[width=0.2\linewidth]{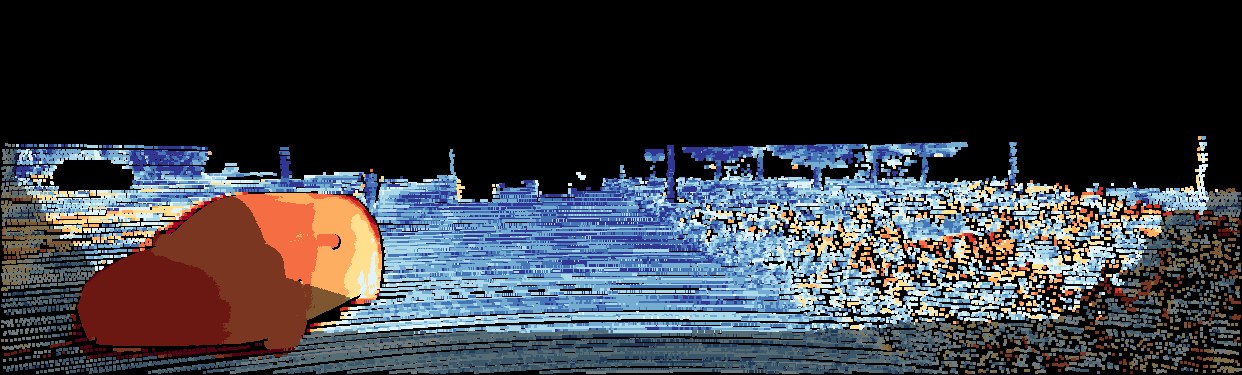}}
    \put(1,23){\scriptsize \textcolor{white}{SF-all: 51.70}}
  \end{picture}}
  \subfloat{\begin{picture}(0.2\linewidth,30)
    \put(0,0){\includegraphics[width=0.2\linewidth]{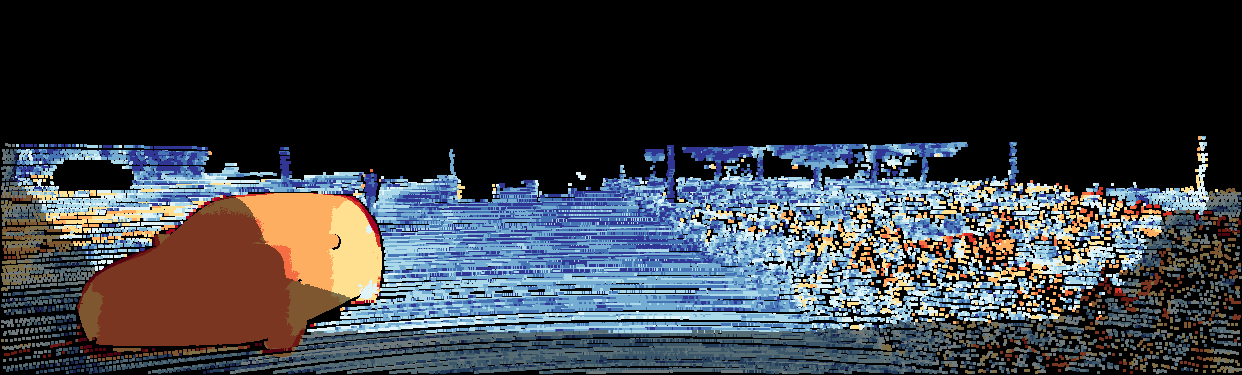}}
    \put(1,23){\scriptsize \textcolor{white}{SF-all: 51.71}}
  \end{picture}}\\

  \includegraphics[width=\linewidth]{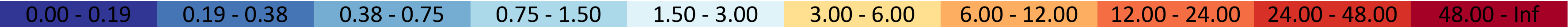}

  \vspace{-6pt}
  
  \caption{Qualitative results on the KITTI Scene Flow test set. Blue indicates a lower error, red indicates a higher error. Our approach improves accuracy near object boundaries. The bottom row shows a failure case: large occlusions confuse all methods.}
  \label{fig:main-kitti}
  \vspace{-5pt}

\end{figure*}

\vspace{3pt} \noindent
\textbf{Quantitative Results.} In Tab. \ref{tab:main-things}, we compare to several state-of-the-art methods which utilize different input modalities. Remarkably, our CamLiPWC surpasses all existing methods including RAFT-3D, which is built on top of the more advanced architecture RAFT. By contrast, our methods have four advantages. \textit{First}, we have better performance, especially for the 3D metrics (EPE\textsubscript{3D}: 0.057m vs. 0.094m). \textit{Second}, RAFT-3D takes dense RGB-D frames as input, while we only require sparse depth measurements. \textit{Third}, RAFT-3D exploits scene rigidity by rigid-motion embeddings, while we do not leverage such assumptions and can handle both rigid and non-rigid motion. \textit{Fourth}, Our methods are much more lightweight with less than 10M parameters. This demonstrates the superiority of multi-stage bidirectional fusion over single-stage fusion.

By extending our fusion pipeline to the RAFT architecture, CamLiRAFT further boosts performance in all aspects. We also add a variant of CamLiRAFT into comparison, which uses the first two stages of ImageNet1k-pretrained \cite{krizhevsky2012imagenet} ResNet-50 \cite{he2015resnet} as the 2D feature encoder. Pretraining on ImageNet accelerates the convergence and slightly improves the final performance.

\vspace{5pt} \noindent
\textbf{Qualitative Results.} The visual comparison of optical flow and scene flow estimation is shown in Fig. \ref{fig:main-things}. Our results are predicted by the best-performing model, CamLiRAFT. As we can see, unimodal methods struggle to achieve satisfactory performance due to either lack of texture or lack of geometric information. Our method can better handle objects with repetitive structures and complex scenes with overlapping objects.

\subsubsection{KITTI}

Since the KITTI leaderboard has a limit on the number of submissions, we only evaluate our best-performing model CamLiRAFT. We follow \cite{yang2021rigidmask, yang2020opticalexp} and divide the 200 training images into \textit{train}, \textit{val} splits based on the 4:1 ratio. The ground-truth disparity maps are lifted into point clouds using the provided calibration parameters.

\vspace{5pt} \noindent
\textbf{Training.} Using the weight pre-trained on FlyingThings3D, we finetune CamLiRAFT on KITTI for 800 epochs. During finetuning, we freeze all BatchNorm \cite{ioffe2015batchnorm} layers as suggested by RAFT. Basic data augmentation strategies including color jitter, random horizontal flipping, random scaling, and random cropping are applied. We use the \texttt{ColorJitter} from Torchvision \cite{paszke2019pytorch} with brightness 0.4, contrast 0.4, saturation 0.2, and hue $0.4/\pi$. Images are randomly rescaled by the factor in the range $[1.0, 1.5]$.

\vspace{5pt} \noindent
\textbf{Testing.} During testing, since neither disparity maps nor point clouds are provided, we employ GA-Net \cite{zhang2019ganet} to estimate the disparity from stereo images, and generate pseudo-LiDAR point clouds with depth $<$ 90m and height $<$ 2m. The sparse output of the point branch is interpolated to create a dense prediction. 

\begin{figure*}
    \centering
    \captionsetup[subfigure]{labelformat=empty,position=top}
    \captionsetup[subfloat]{captionskip=1pt}

    \subfloat[Reference Frame]{\includegraphics[width=0.198\linewidth]{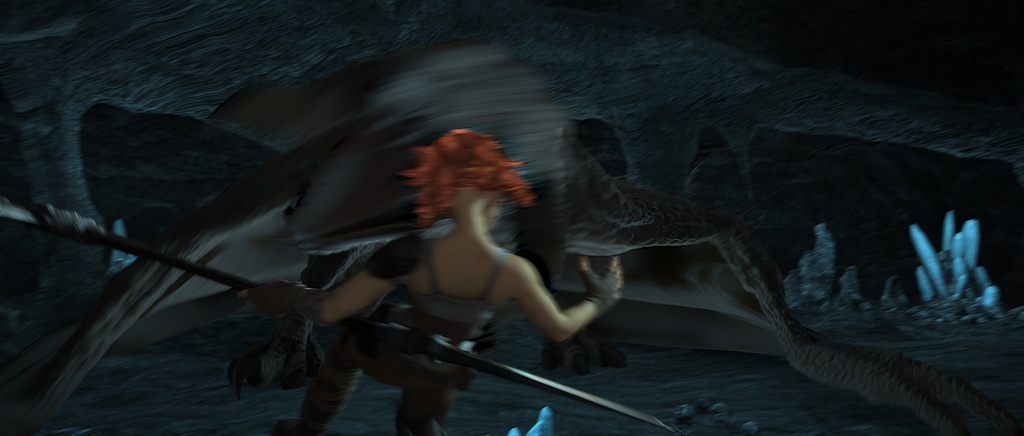}}
    \hfill
    \subfloat[RAFT]{\includegraphics[width=0.198\linewidth]{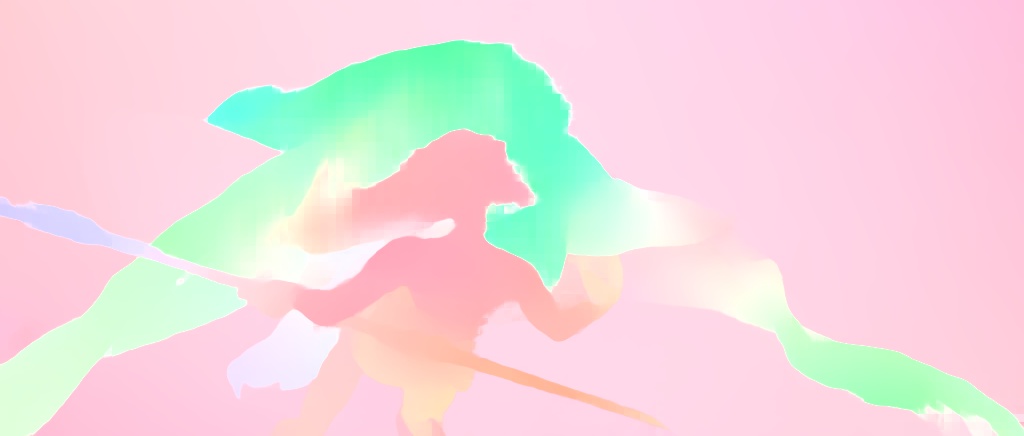}}
    \hfill
    \subfloat[RAFT-3D]{\includegraphics[width=0.198\linewidth]{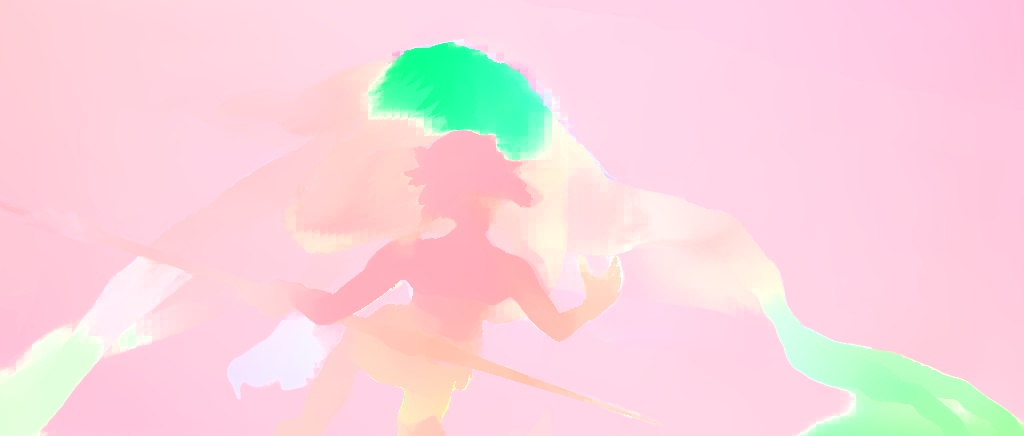}}
    \hfill
    \subfloat[CamLiRAFT (Ours)]{\includegraphics[width=0.198\linewidth]{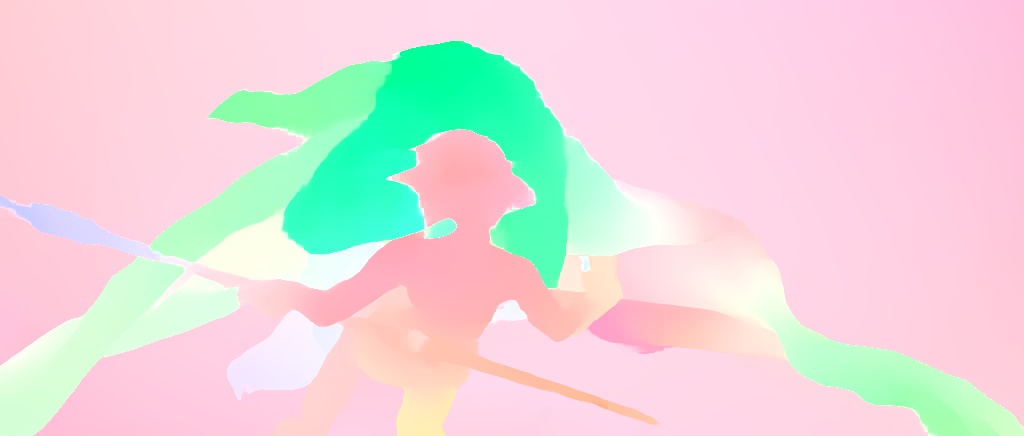}}
    \hfill
    \subfloat[Ground Truth]{\includegraphics[width=0.198\linewidth]{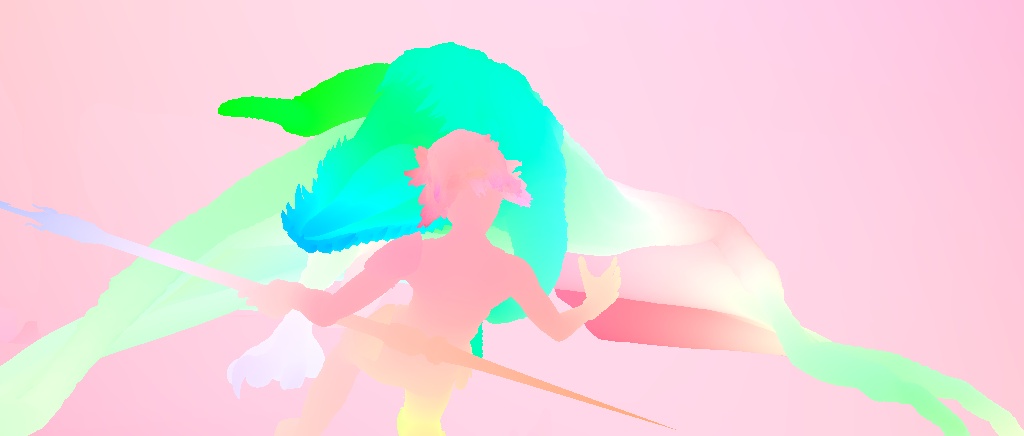}}\\
    \vspace{-9pt}
    
    \subfloat{\includegraphics[width=0.198\linewidth]{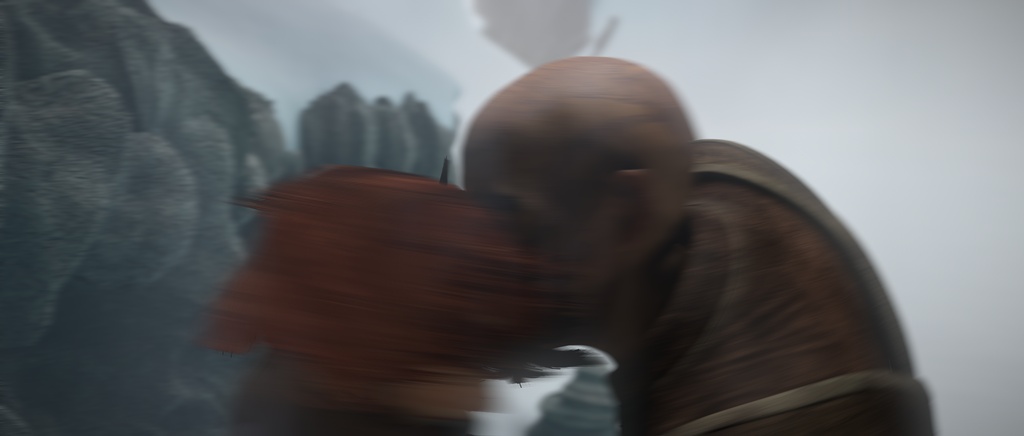}}
    \hfill
    \subfloat{\includegraphics[width=0.198\linewidth]{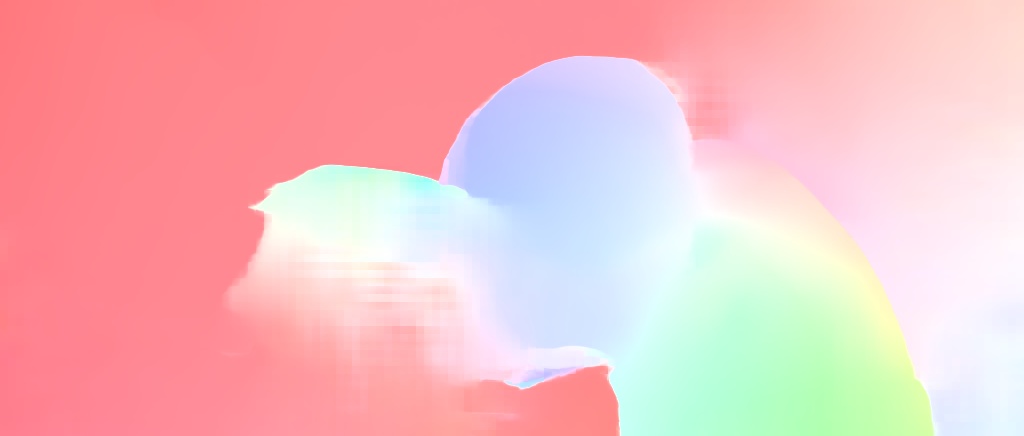}}
    \hfill
    \subfloat{\includegraphics[width=0.198\linewidth]{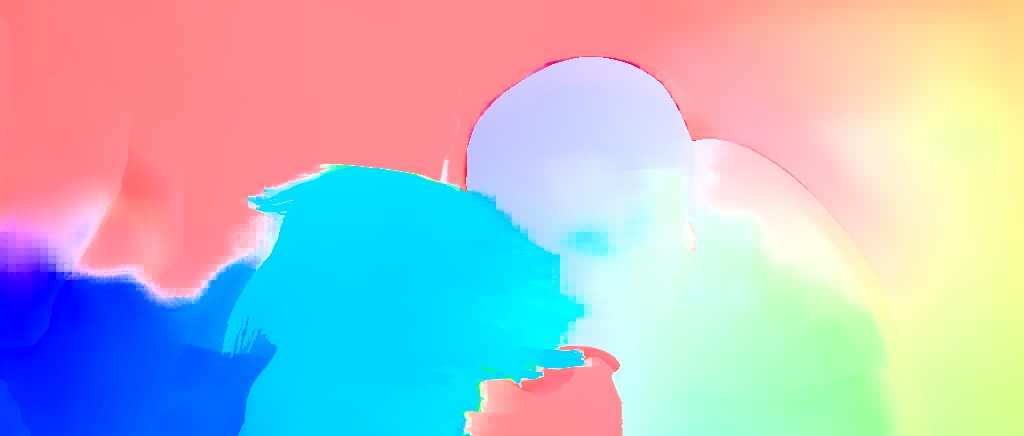}}
    \hfill
    \subfloat{\includegraphics[width=0.198\linewidth]{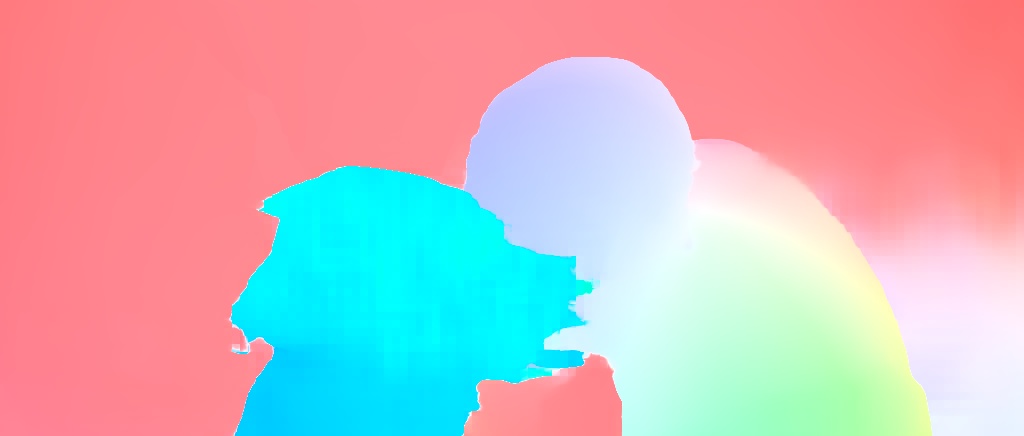}}
    \hfill
    \subfloat{\includegraphics[width=0.198\linewidth]{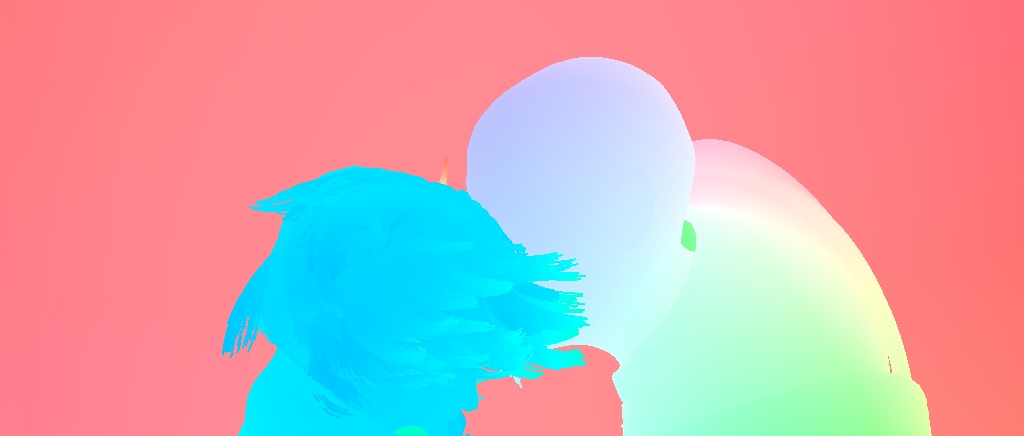}}\\
    \vspace{-9pt}
    
    \subfloat{\includegraphics[width=0.198\linewidth]{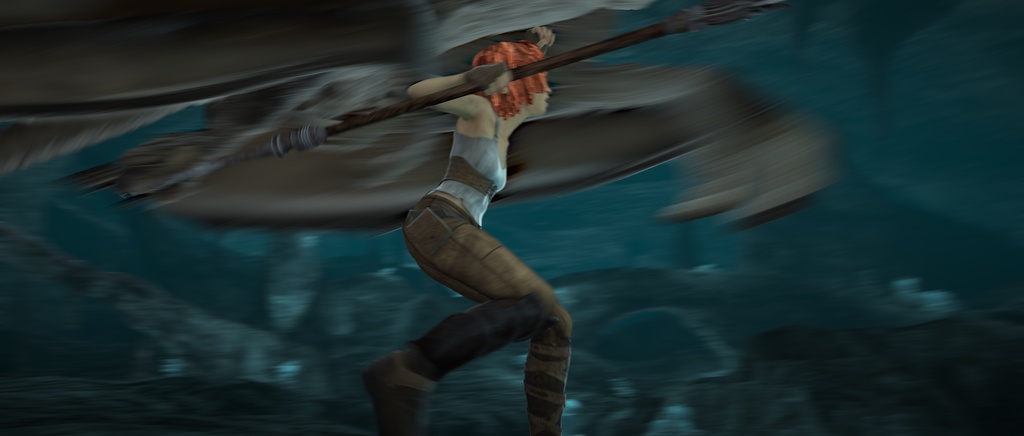}}
    \hfill
    \subfloat{\includegraphics[width=0.198\linewidth]{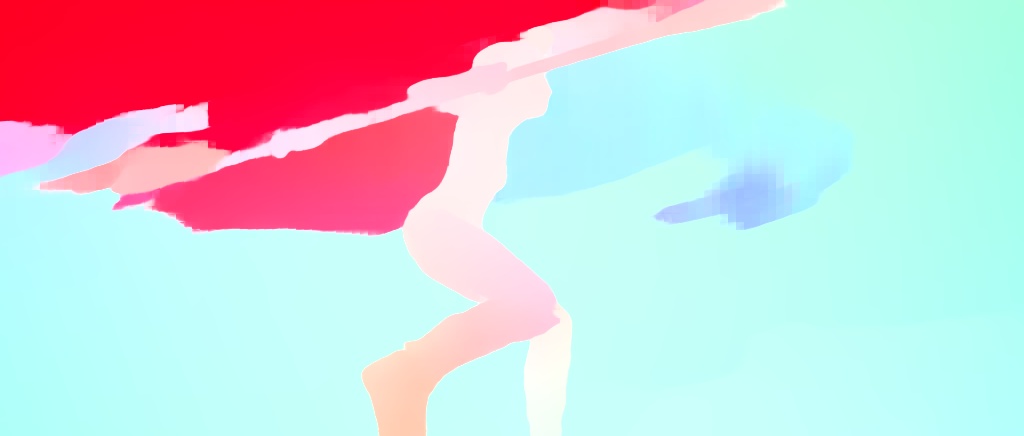}}
    \hfill
    \subfloat{\includegraphics[width=0.198\linewidth]{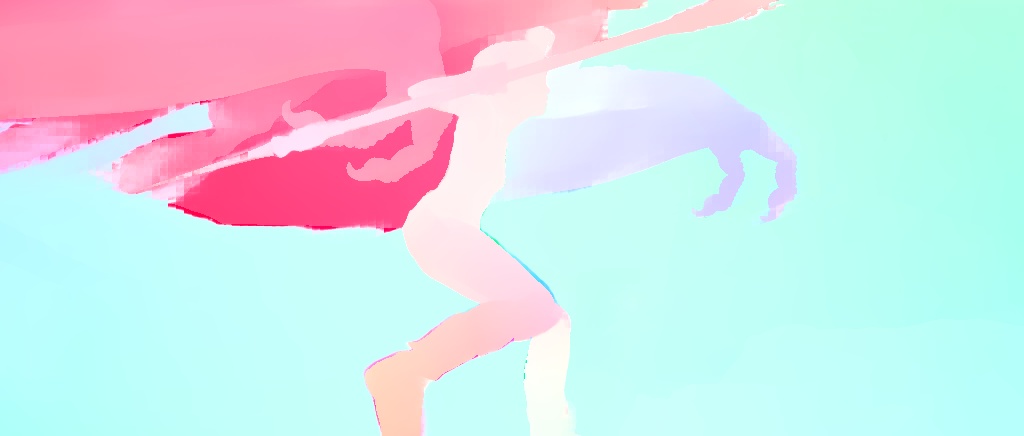}}
    \hfill
    \subfloat{\includegraphics[width=0.198\linewidth]{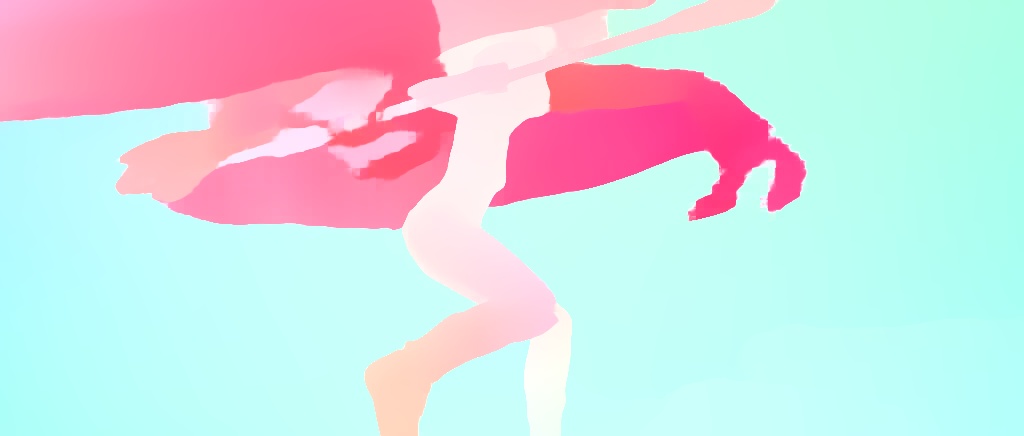}}
    \hfill
    \subfloat{\includegraphics[width=0.198\linewidth]{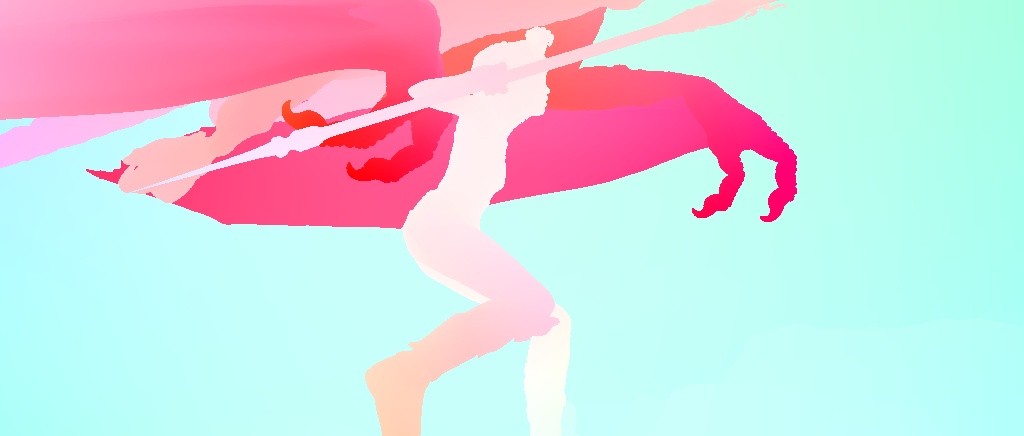}}\\
    \vspace{-9pt}
    
    \subfloat{\includegraphics[width=0.198\linewidth]{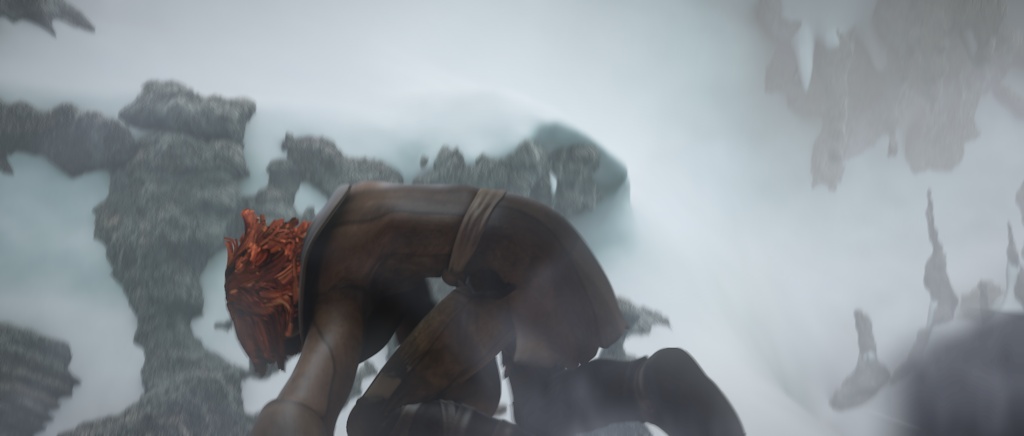}}
    \hfill
    \subfloat{\includegraphics[width=0.198\linewidth]{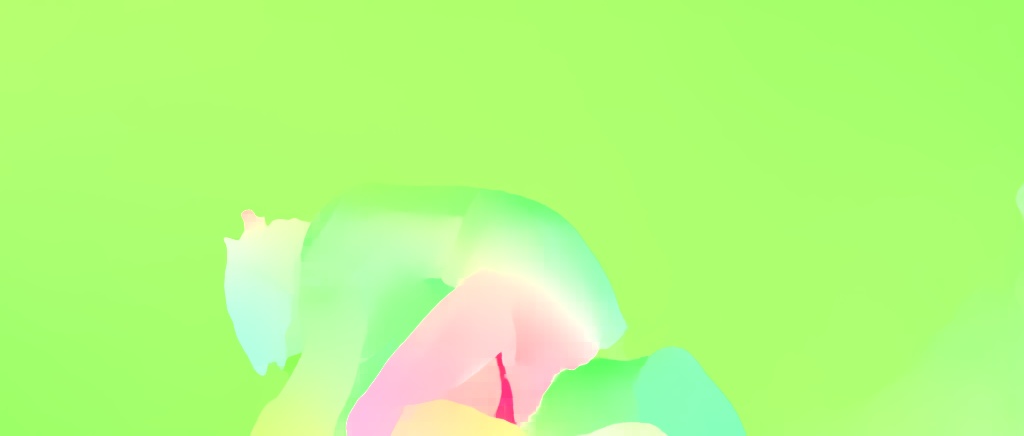}}
    \hfill
    \subfloat{\includegraphics[width=0.198\linewidth]{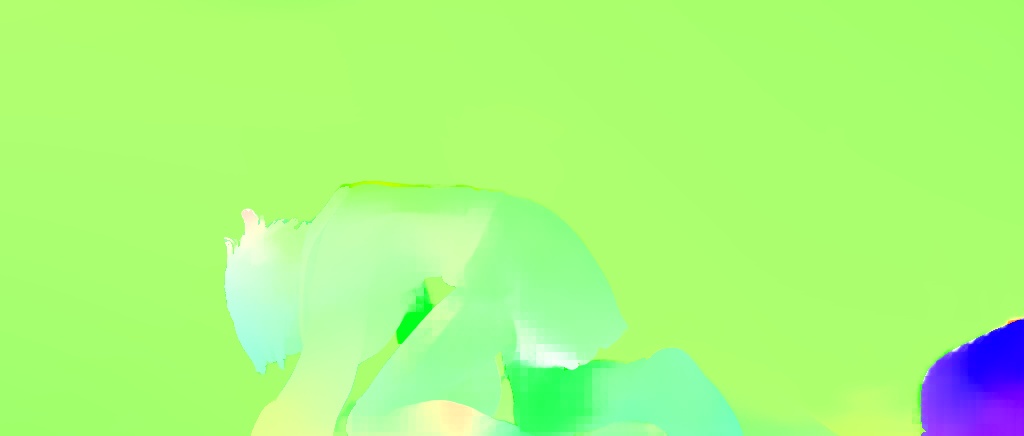}}
    \hfill
    \subfloat{\includegraphics[width=0.198\linewidth]{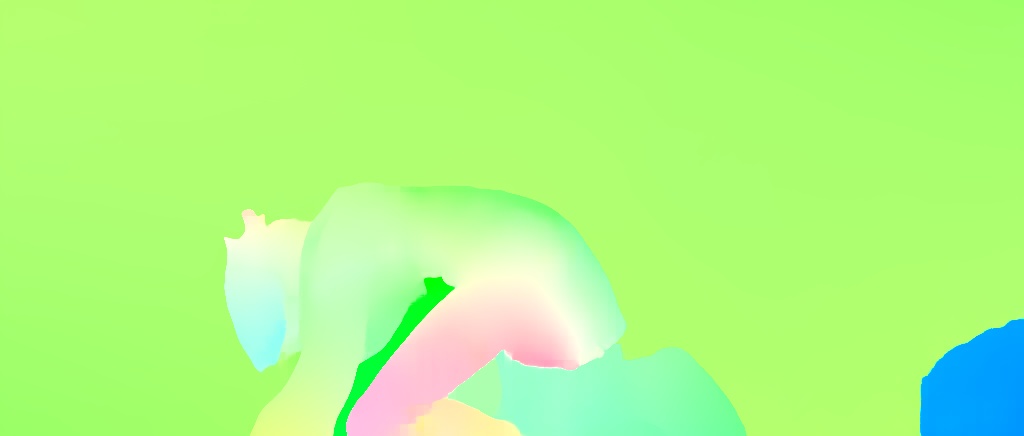}}
    \hfill
    \subfloat{\includegraphics[width=0.198\linewidth]{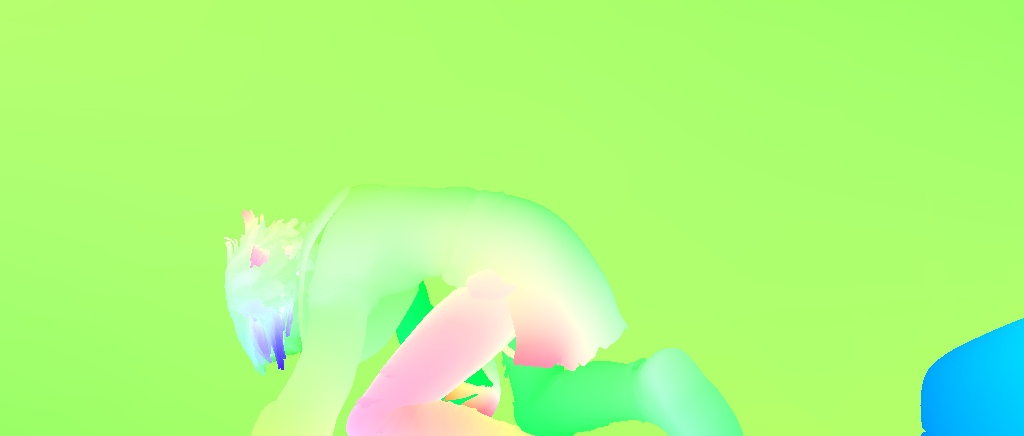}}\\
    \vspace{-9pt}

    \subfloat{\includegraphics[width=0.198\linewidth]{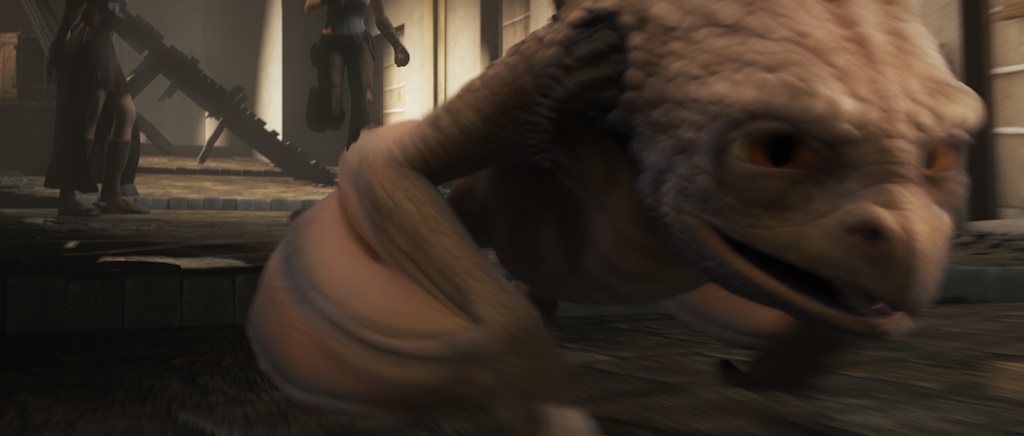}}
    \hfill
    \subfloat{\includegraphics[width=0.198\linewidth]{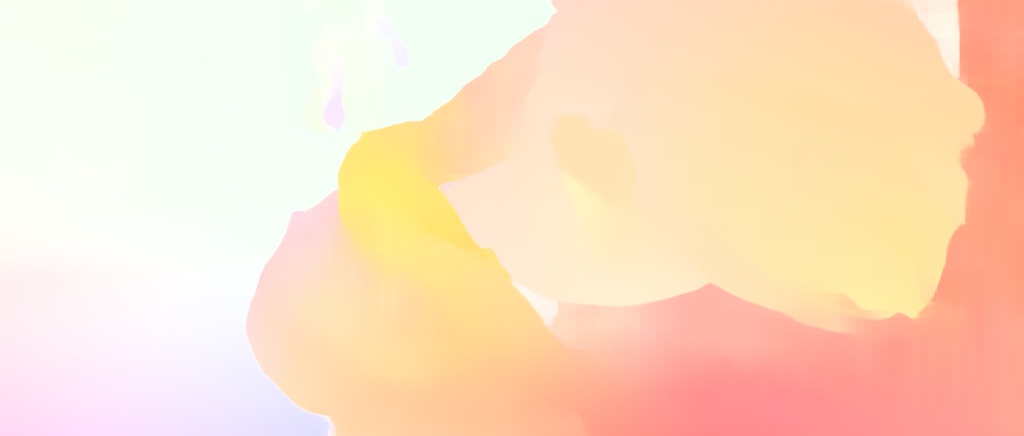}}
    \hfill
    \subfloat{\includegraphics[width=0.198\linewidth]{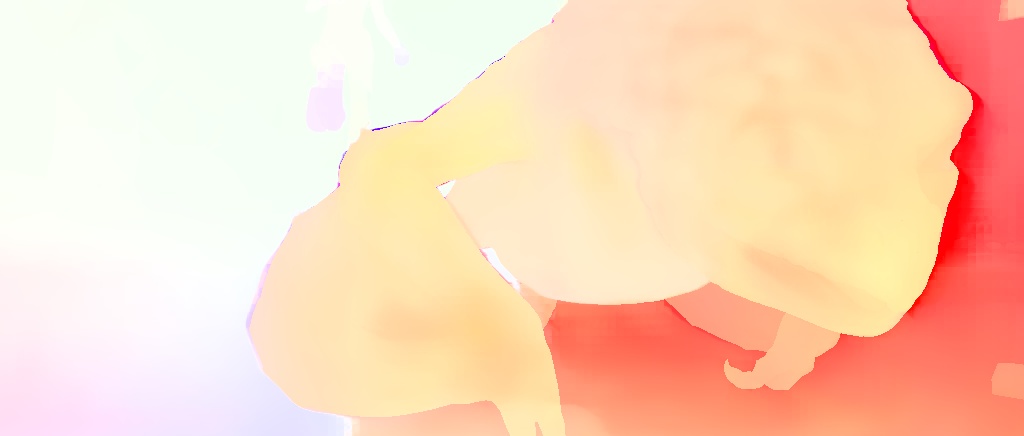}}
    \hfill
    \subfloat{\includegraphics[width=0.198\linewidth]{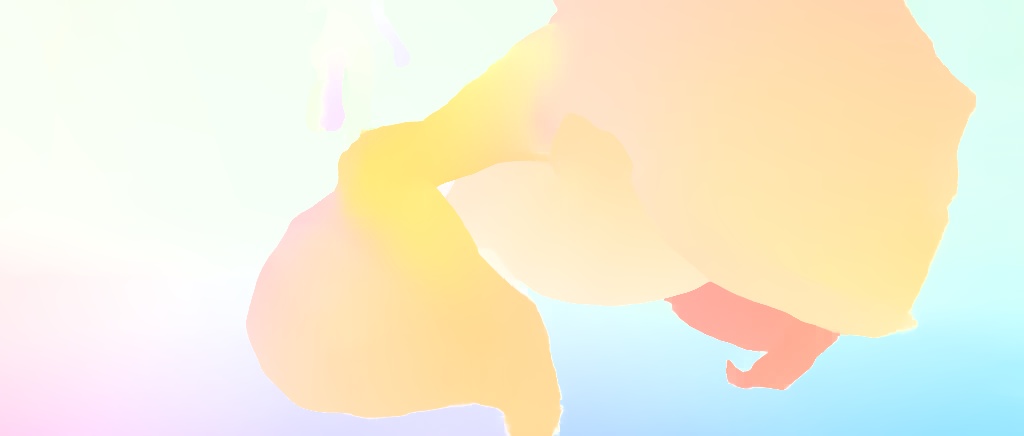}}
    \hfill
    \subfloat{\includegraphics[width=0.198\linewidth]{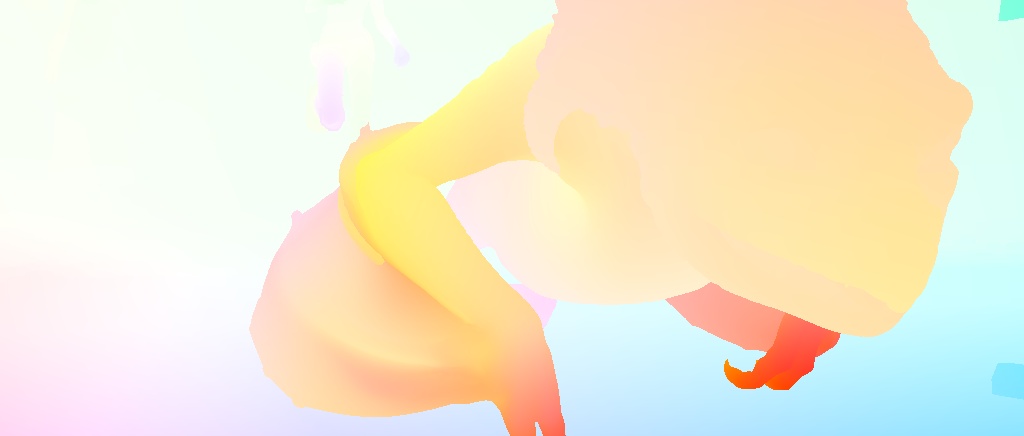}}\\
    
    \caption{Visualized comparison on the non-rigid Sintel dataset. All methods are only trained on synthetic data and validated on the \textit{training} split of Sintel. Our CamLiRAFT has good generalization performance and can handle generic non-rigid motions.}
    \label{fig:main-sintel}
    \vspace{-5pt}
    
    \end{figure*}

\vspace{5pt} \noindent
\textbf{Refinement of Background Scene Flow.} Since most background objects in KITTI are rigid (e.g. ground, buildings, etc), we can refine the background scene flow using a rigidity refinement step. Specifically, we employ DDRNet-Slim \cite{hong2021ddrnet}, a lightweight 2D semantic segmentation network, to determine the rigid background. DDRNet-Slim is pre-trained on Cityscapes \cite{cordts2016cityscapes} and then fine-tuned on KITTI. Next, we estimate the ego-motion by fitting and decomposing essential matrices from the background flow map using a neural-guided RANSAC \cite{brachmann2019ransac}. Finally, the background scene flow is generated using the ego motion and replaces the original flow predicted by the model.

\vspace{5pt} \noindent
\textbf{Comparison with State-of-the-art Methods.} We submit our approach to the website of the KITTI Scene Flow benchmark and report the leaderboard in Tab. \ref{tab:main-kitti}. A visualized comparison is shown in Fig. \ref{fig:main-kitti}. Notably, \textit{without} leveraging any rigid body assumption, CamLiRAFT performs on par with the previous state-of-the-art method RigidMask \cite{yang2021rigidmask} (SF-all: 4.97\% vs. 4.89\%). In contrast, RigidMask leverages more strict rigid-body assumptions by assigning rigid motion to all objects and employs 10 $\times$ more parameters. By refining the background scene flow with the priors of rigidity, CamLiRAFT achieves an error of {\bf4.26\%}, outperforming all previous methods by a large margin. Moreover, our method is much more lightweight with only 20.4M parameters (6.3M GA-Net + 8.4M CamLiRAFT + 5.7M DDRNet-Slim) and can handle general non-rigid motion since we only apply rigid motion refinement to the static background.

Moreover, we would like to emphasize that the power of CamLiRAFT is not fully unlocked since the pseudo-LiDAR point clouds generated by stereo matching methods (e.g. GA-Net) are not perfect. In practice, when deploying to a car equipped with a real LiDAR sensor, CamLiRAFT can achieve much better performance.

\subsubsection{Sintel}

\begin{table}[t]
  \small
  \renewcommand{\arraystretch}{1.15}
  \caption{Performance comparison on the training set of Sintel without finetuning. C and T denote FlyingChairs and FlyingThings3D respectively. }
  \label{tab:main-sintel}
  \centering
  \begin{tabular}{l|c|c|cc}
  \hline
  \multirow{2}{*}{Method} & Training & \multirow{2}{*}{Input} & \multicolumn{2}{c}{Sintel} \\
  & Data & & Clean & Final \\
  \hline
  PWC-Net \cite{sun2018pwc} & C + T & RGB & 2.55 & 3.93 \\
  FlowNet2 \cite{ilg2017flownet2} & C + T & RGB & 2.02 & 3.54 \\
  RAFT \cite{teed2020raft} & C + T & RGB & 1.43 & 2.71 \\
  RAFT-3D \cite{teed2021raft3d} & T & RGB+Depth & 1.75 & 2.91 \\
  \hline
  \textbf{CamLiPWC} & T & RGB+XYZ & 1.79 & 2.55 \\
  \textbf{CamLiRAFT} & T & RGB+XYZ & \textbf{1.27} & \textbf{2.38} \\
  \hline
  \end{tabular}
  \vspace{-5pt}
\end{table}

Previous scene flow methods \cite{yang2021rigidmask,teed2021raft3d,ma2019drisf,behl2017isf,menze2015osf} typically only focus on rigid scenarios. In this section, we test the performance of our method on MPI Sintel \cite{sintel}, which mainly consists of non-rigid motion. Since the depth maps are not available on the test set, we evaluate the generalization performance: models are trained on synthetic data and validated on the training split of Sintel \textit{without} finetuning. Similarly, we lift the ground truth depth maps to point clouds with depth $<$ 30m.

\vspace{5pt} \noindent
\textbf{Quantitative Results.} In Tab. \ref{tab:main-sintel}, we compare our methods to the unimodal baselines and multimodal competitor RAFT-3D. Due to the lack of ground-truth 3D scene flow, we only evaluate optical flow performance. Since RAFT-3D introduces rigid-motion embeddings into the model, it is not suitable for non-rigid scenarios and its performance is even worse than RAFT (2.91 vs. 2.71 on the final pass). In contrast, our methods can handle generic non-rigid motion and have good generalization performance. Even CamLiPWC can achieve 2.55 on the last pass of Sintel, surpassing all previous work. This exciting result demonstrates the importance of 3D geometry and multi-modal fusion. CamLiRAFT further achieves 1.27 and 2.38 on the clean and final pass respectively, reducing the error by 11.2\% and 12.2\% compared with its unimodal baseline RAFT.

\begin{table*}
    \small
    \renewcommand{\arraystretch}{1.15}
    \caption{Comparison with existing LiDAR-only scene flow methods on FlyingThings3D. ``-L'' denotes the LiDAR-only variant of our method which removes the image branch. $\mathtt{T}\times$ denotes the number of iterative updates.}
    \vspace{-5pt}
    \label{tab:things-lidar-only}
    \centering
    \begin{tabular}{l|cccc|cccc|c}
      \hline
      \multirow{2}{*}{Methods} & \multicolumn{4}{c|}{Non-occluded FlyingThings3D} & \multicolumn{4}{c|}{Occluded FlyingThings3D} & Throughput \\
      & EPE\textsubscript{3D} & ACC\textsubscript{S} & ACC\textsubscript{R} & Outliers & EPE\textsubscript{3D} & ACC\textsubscript{S} & ACC\textsubscript{R} & Outliers & (pair/s) \\
      \hline
      \multicolumn{10}{l}{\emph{Coarse-to-fine Methods}} \\
      \hline
      FlowNet3D \cite{liu2019flownet3d}          & 0.113 & 41.2\% & 77.1\% & 60.2\% & 0.158 & 22.9\% & 58.2\% & 80.4\% & 94.4 \\
      HPLFlowNet \cite{gu2019hplflownet}         & 0.080 & 61.4\% & 85.5\% & 42.9\% & 0.169 & 26.3\% & 57.5\% & 81.2\% & - \\
      PointPWC \cite{wu2019pointpwc}             & 0.059 & 73.8\% & 92.8\% & 34.2\% & 0.155 & 41.6\% & 69.9\% & 63.9\% & 12.6 \\
      FLOT \cite{puy2020flot}                    & 0.052 & 73.2\% & 92.7\% & 35.7\% & 0.153 & 39.7\% & 66.1\% & 66.3\% & 8.6 \\
      OGSF-Net \cite{ouyang2021ogsf}             & - & - & - & - & 0.122 & 55.2\% & 77.7\% & 51.8\% & 14.1 \\
      \textbf{CamLiPWC-L} & \textbf{0.032} & \textbf{92.5\%} & \textbf{97.9\%} & \textbf{15.6\%} & \textbf{0.092} & \textbf{71.5\%} & \textbf{87.1\%} & \textbf{37.2\%} & 61.6 \\
      \hline
      \multicolumn{10}{l}{\emph{Recurrent Methods}} \\
      \hline
      PV-RAFT \cite{wei2021pvraft}               & 0.046 & 81.7\% & 95.7\% & 29.2\% & - & - & - & - & 1.4 \\
      FlowStep3D \cite{kittenplon2021flowstep3d}  & 0.045 & 81.6\% & 96.1\% & 21.6\% & - & - & - & - & 3.6\\
      \textbf{CamLiRAFT-L} ($8 \times$) & \textbf{0.029} & \textbf{93.0\%} & \textbf{98.0\%} & \textbf{13.4\%} & 0.078 & 78.4\% & 89.6\% & 28.3\% & 20.8 \\
      \textbf{CamLiRAFT-L} ($20 \times$) & \textbf{0.029} & \textbf{93.0\%} & \textbf{98.0\%} & 13.6\% & \textbf{0.076} & \textbf{79.4\%} & \textbf{90.4\%} & \textbf{27.9\%} & 9.5 \\
      \hline
    \end{tabular}
  \end{table*}
  
  \begin{table*}
    \small
    \renewcommand{\arraystretch}{1.15}
    \caption{Comparison with existing LiDAR-only scene flow methods on KITTI. ``-L'' and $\mathtt{T}\times$ have the same meaning as above. All methods are trained on FlyingThings3D and tested on KITTI without finetuning.}
    \vspace{-5pt}
    \label{tab:kitti-lidar-only}
    \centering
    \begin{tabular}{l|cccc|cccc|c}
      \hline
      \multirow{2}{*}{Methods} & \multicolumn{4}{c|}{Non-occluded KITTI} & \multicolumn{4}{c|}{Occluded KITTI} & Throughput \\
      & EPE\textsubscript{3D} & ACC\textsubscript{S} & ACC\textsubscript{R} & Outliers & EPE\textsubscript{3D} & ACC\textsubscript{S} & ACC\textsubscript{R} & Outliers & (pair/s) \\
      \hline
      FlowNet3D \cite{liu2019flownet3d}          & 0.177 & 37.4\% & 66.8\% & 52.7\%    & 0.183 &  9.8\% & 39.4\% & 79.9\% & 94.4 \\
      HPLFlowNet \cite{gu2019hplflownet}         & 0.117 & 47.8\% & 77.8\% & 41.0\%    & 0.343 & 10.3\% & 38.6\% & 81.4\% & - \\
      PointPWC \cite{wu2019pointpwc}             & 0.069 & 72.8\% & 88.8\% & 26.5\%    & 0.118 & 40.3\% & 75.7\% & 49.6\% & 12.6 \\
      FLOT \cite{puy2020flot}                    & 0.056 & 75.5\% & 90.8\% & 24.2\%    & 0.130 & 27.8\% & 66.7\% & 52.9\% & 8.6 \\
      OGSF-Net \cite{ouyang2021ogsf}             & - & - & - & -                       & 0.075 & 70.6\% & 86.9\% & 32.7\% & 14.1 \\
      PV-RAFT \cite{wei2021pvraft}               & 0.056 & 82.3\% & 93.7\% & 21.6\%    & - & - & - & - & 1.4 \\
      FlowStep3D \cite{kittenplon2021flowstep3d} & 0.055 & 80.5\% & 92.5\% & \textbf{14.9\%}    & - & - & - & - & 3.6 \\
      \hline
      \textbf{CamLiRAFT-L} ($8 \times$)          & 0.046 & 88.1\% & 95.7\% & 16.6\% & 0.058 & 78.4\% & 92.5\% & 23.2\% & 20.8 \\
      \textbf{CamLiRAFT-L} ($20 \times$)         & \textbf{0.038} & \textbf{90.3\%} & \textbf{97.3\%} & 15.0\% & \textbf{0.055} & \textbf{78.9\%} & \textbf{92.9\%} & \textbf{22.9\%} & 9.5 \\
      \hline
    \end{tabular}
  \end{table*}

\vspace{5pt} \noindent
\textbf{Qualitative Results.} We further select some challenging samples in the final pass of Sintel and provide a visualized comparison in Fig. \ref{fig:main-sintel}. As we can see, the rigid-motion embedding of RAFT-3D leads to large areas of error, whereas our method does not rely on any rigid prior and has robust performance in various scenarios. With the help of 3D information, we can overcome the challenges such as low light, motion blur, etc.

\subsection{Comparison with LiDAR-only Methods}
\label{sec:lidar-only-cmp}

Building a strong and efficient point branch is an important prerequisite for bidirectional multi-modal fusion. In this section, we compare the point branch of our two models (dubbed CamLiPWC-L and CamLiRAFT-L) with existing LiDAR-only scene flow methods on FlyingThings3D and KITTI respectively.

\vspace{5pt} \noindent
\textbf{Evaluation Settings.} There are two different ways of data preprocessing. The first setting is the one proposed by HPLFlowNet \cite{gu2019hplflownet}, which only keeps non-occluded points during the preprocessing. The second setting, proposed by FlowNet3D \cite{liu2019flownet3d}, remains the occluded points. We report results on both occluded/non-occluded settings for a more comprehensive comparison.

\vspace{5pt} \noindent
\textbf{Quantitative Results.} In Tab. \ref{tab:things-lidar-only}, we report results on FlyingThings3D. Without the assistance of cameras, our point branches still outperform previous work by a large margin. Moreover, the throughput of our methods (measured on a Tesla V100 GPU with a batch size of 8) is much higher than the competitors. On the non-occluded data, CamLiPWC-L reduces the EPE\textsubscript{3D} by 46\% compared with PointPWC-Net, while running $5\times$ faster. CamLiRAFT-L (8 iterations) reduces the error by 36\% and runs $6\times$ faster than FlowStep3D. 
In Tab. \ref{tab:kitti-lidar-only}, we test the generalization ability of our best model CamLiRAFT-L on KITTI. Here, we follow GMSF \cite{zhang2023gmsf} to adjust the KITTI data to cover the same domain range using the mean and standard deviation. Once again, CamLiRAFT-L outperforms all previous methods under both occluded/non-occluded settings.
We hope that our point branches can serve as strong baselines for LiDAR-only scene flow estimation.

\vspace{5pt} \noindent
\textbf{Discussion.} Although CamLiRAFT-L seems to have something in common with existing methods such as FlowStep3D \cite{kittenplon2021flowstep3d} and PV-RAFT \cite{wei2021pvraft}, it is difficult to adopt their methods as our point branch due to the heterogenous architecture, complicated training schedule, and heavy computation. (1) FlowStep3D only unrolls for 4 iterations since the performance starts to decrease after that. The reason might be that the large errors of the initial coarse flow fail to be corrected by the iterative updates, since the correlation unit of the update module only has a local receptive field. In contrast, the point-based correlation pyramid of CamLiRAFT-L can capture both small and large motion, and the performance continues to improve as the number of iterations increases, which is consistent with RAFT. This is crucial for our fusion pipeline, as the iteration number of the point and image branch should match to enable the fusion inside the update block. (2) PV-RAFT relies on a flow refinement module which makes the prediction smoother. Due to memory constraints, the refinement module is not trained end-to-end with the main network. CamLiRAFT-L, however, does not need extra networks for refinement and can be trained in an end-to-end manner. (3) Both FlowStep3D and PV-RAFT are not time-efficient due to complicated design and slow implementation. Specifically, training them takes more than 4 days, which is not applicable with limited budges. On the contrary, equipped with better designs and optimized implementation, CamLiRAFT-L can be trained in less than 24 hours while still achieving superior performance over the two competitors.

\subsection{Ablation Study}
\label{sec:ablation}

\begin{figure*}
  \centering
  \captionsetup[subfigure]{labelformat=empty,position=top}
  \captionsetup[subfloat]{captionskip=2pt}

  \subfloat[Reference Frame]{%
    \includegraphics[width=0.166\linewidth]{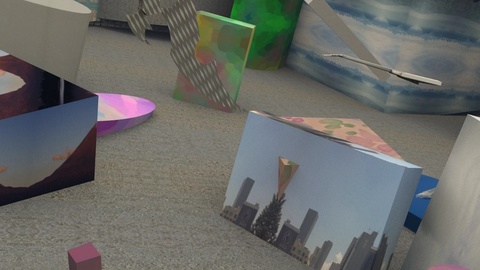}}
  \subfloat[W/o Fusion]{%
    \includegraphics[width=0.166\linewidth]{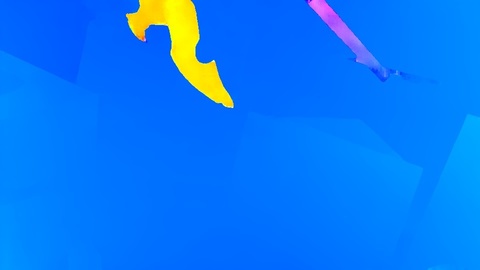}}
  \subfloat[+ Encoder Fusion]{%
    \includegraphics[width=0.166\linewidth]{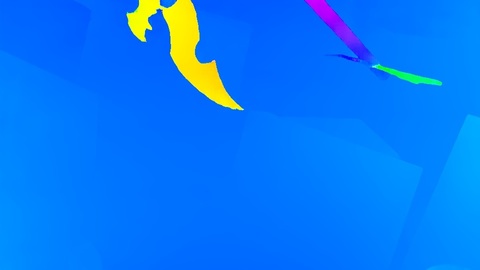}}
  \subfloat[+ Correlation Fusion]{%
    \includegraphics[width=0.166\linewidth]{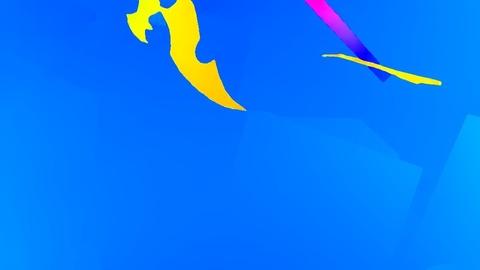}}
  \subfloat[+ Decoder Fusion]{%
    \includegraphics[width=0.166\linewidth]{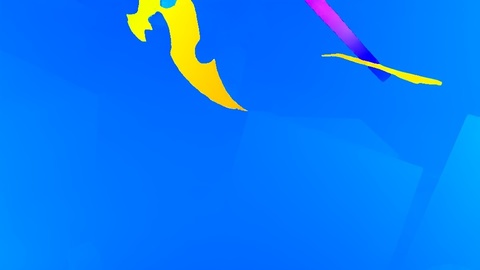}}
  \subfloat[Ground Truth]{%
    \includegraphics[width=0.166\linewidth]{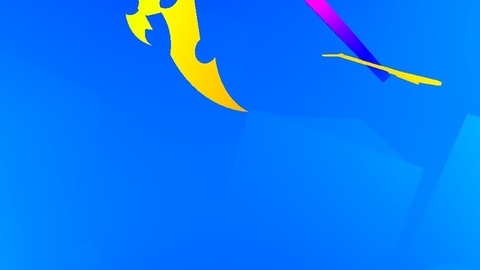}}\\
  \vspace{-10pt}

  \subfloat{\includegraphics[width=0.166\linewidth]{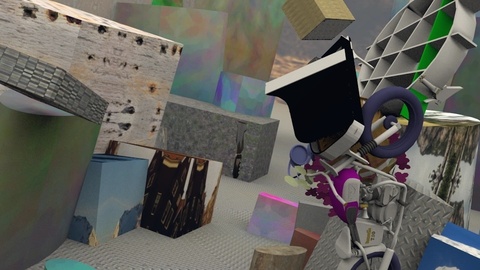}}
  \subfloat{\includegraphics[width=0.166\linewidth]{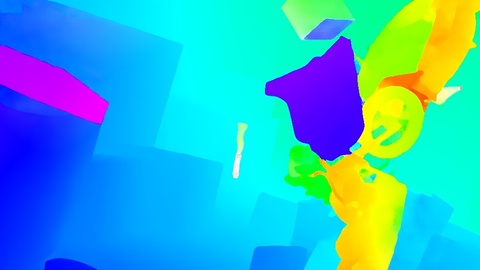}}
  \subfloat{\includegraphics[width=0.166\linewidth]{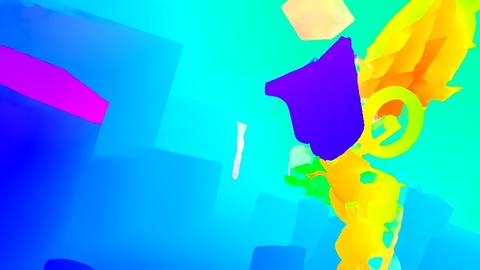}}
  \subfloat{\includegraphics[width=0.166\linewidth]{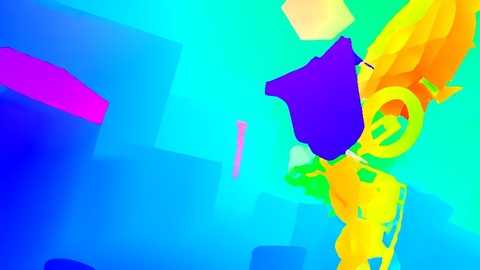}}
  \subfloat{\includegraphics[width=0.166\linewidth]{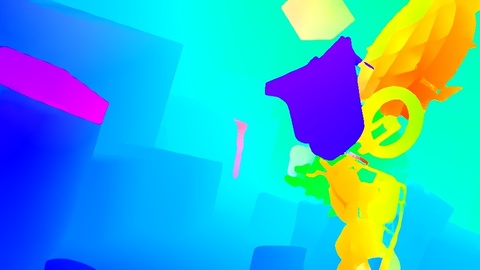}}
  \subfloat{\includegraphics[width=0.166\linewidth]{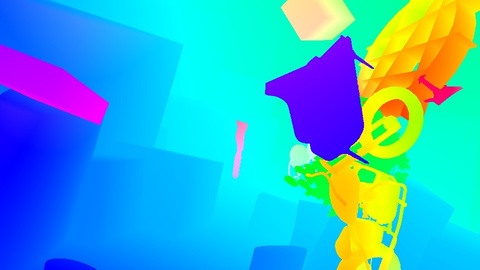}}\\
  \vspace{-10pt}

  \subfloat{\includegraphics[width=0.166\linewidth]{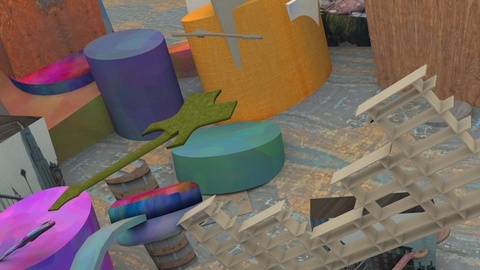}}
  \subfloat{\includegraphics[width=0.166\linewidth]{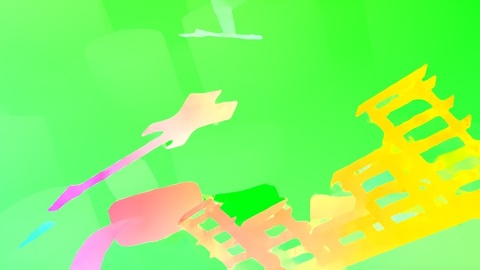}}
  \subfloat{\includegraphics[width=0.166\linewidth]{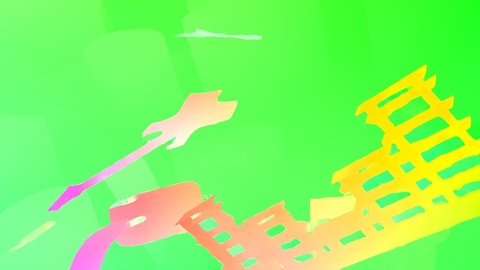}}
  \subfloat{\includegraphics[width=0.166\linewidth]{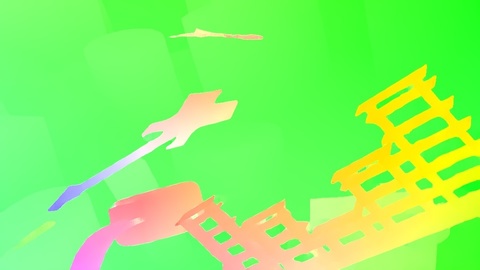}}
  \subfloat{\includegraphics[width=0.166\linewidth]{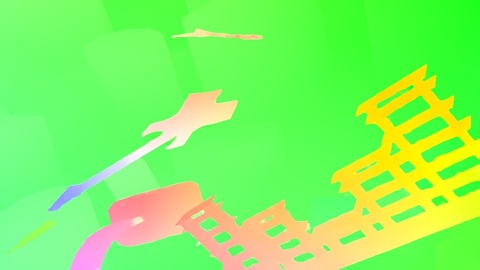}}
  \subfloat{\includegraphics[width=0.166\linewidth]{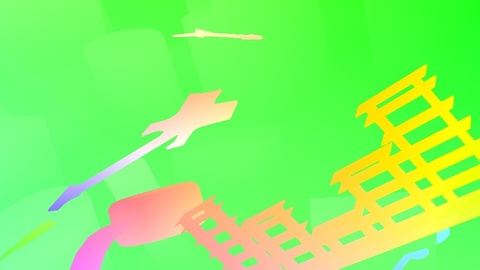}}\\
  
  \caption{A visual ablation on the multi-stage fusion pipeline of CamLiPWC. Fusing feature encoder helps to recover the structure and boundary of objects. Fusing the correlation feature enables the network to capture small and fast-moving objects. Fusing the flow decoder improves the performance on complex scenes where objects overlap.}
  \label{fig:multi-stage-ablation}
  \vspace{-5pt}
\end{figure*}

In this section, we conduct ablation studies on FlyingThings3D to confirm the effectiveness of each module. Due to the limited computational resource, we halve the number of training epochs, i.e. 100 epochs for CamLiRAFT and 300 epochs for CamLiPWC. Unless otherwise stated, we use CamLiPWC as the reference model. The default choice for our model is colored \colorbox{Gray}{gray}.

\begin{table}[t]
  \small
  \renewcommand{\arraystretch}{1.15}
  \caption{Unidirectional Fusion vs. Bidirectional Fusion. Bidirectional fusion provides the best results for all metrics.}
  \label{tab:ablation-fusion-direction}
  \centering
  \begin{tabular}{c|cc|cc}
  \hline
  \multirow{2}{*}{Fusion Direction} & \multicolumn{2}{c|}{2D Metrics} & \multicolumn{2}{c}{3D Metrics} \\
  & EPE\textsubscript{2D} & ACC\textsubscript{1px} & EPE\textsubscript{3D} & ACC\textsubscript{.05} \\
  \hline
  \multicolumn{5}{l}{\emph{Reference Model: CamLiPWC}} \\
  \hline
  2D $\Rightarrow$ 3D     & 3.63 & 78.6\% & 0.072 & 80.3\% \\
  2D $\Leftarrow$ 3D      & 2.58 & 82.0\% & 0.107 & 65.4\% \\
  \rowcolor{Gray}
  2D $\Leftrightarrow$ 3D & \textbf{2.52} & \textbf{82.2\%} & \textbf{0.070} & \textbf{82.3\%} \\
  \hline
  \multicolumn{5}{l}{\emph{Reference Model: CamLiRAFT}} \\
  \hline
  2D $\Rightarrow$ 3D     & 2.46 & 84.2\% & 0.056 & 85.3\% \\
  2D $\Leftarrow$ 3D      & 1.88 & \textbf{85.9\%} & 0.098 & 71.2\% \\
  \rowcolor{Gray}
  2D $\Leftrightarrow$ 3D & \textbf{1.86} & \textbf{85.9\%} & \textbf{0.053} & \textbf{86.2\%} \\
  \hline
  \end{tabular}
\end{table}

\vspace{5pt} \noindent
\textbf{Unidirectional Fusion vs. Bidirectional Fusion.} In our pipeline, features are fused in a bidirectional manner. Here, we train two variants for CamLiPWC and CamLiRAFT where features are fused in a unidirectional manner (2D $\Rightarrow$ 3D or 2D $\Leftarrow$ 3D). As shown in Tab. \ref{tab:ablation-fusion-direction}, unidirectional fusion either improves the 2D or 3D metrics, while bidirectional fusion provides better results for both modalities. The improvement is consistent across different architectures. Moreover, compared with unidirectional fusion, bidirectional fusion improves the best EPE\textsubscript{2D} from 1.88 to 1.86 and EPE\textsubscript{3D} from 0.056 to 0.053, suggesting that the improvement of one modality can also benefit the other one.

\begin{table}[t]
  \small
  \renewcommand{\arraystretch}{1.15}
  \caption{Single-stage fusion vs. multi-stage fusion. Multi-stage fusion performs much better than single-stage fusion.}
  \label{tab:ablation-multi-stage}
  \centering
  \begin{tabular}{l|cc|cc}
  \hline
  \multirow{2}{*}{Fusion Stages} & \multicolumn{2}{c|}{2D Metrics} & \multicolumn{2}{c}{3D Metrics} \\
  & EPE\textsubscript{2D} & ACC\textsubscript{1px} & EPE\textsubscript{3D} & ACC\textsubscript{.05} \\
  \hline
  \multicolumn{5}{l}{\emph{Reference Model: CamLiPWC}} \\
  \hline
  -                    & 3.64 & 78.8\% & 0.108 & 65.2\% \\
  $S_F$                & 2.87 & 81.3\% & 0.088 & 78.5\% \\
  $S_F$, $S_C$         & 2.57 & 81.9\% & 0.075 & 80.8\% \\
  \rowcolor{Gray}
  $S_F$, $S_C$, $S_D$  & \textbf{2.52} & \textbf{82.2\%} & \textbf{0.070} & \textbf{82.3\%} \\
  \hline
  \multicolumn{5}{l}{\emph{Reference Model: CamLiRAFT}} \\
  \hline
  -                                   & 2.45 & 84.0\% & 0.100 & 71.1\% \\
  $S_F$                               & 2.31 & 84.2\% & 0.074 & 81.3\% \\
  $S_F$, $S_X$                        & 2.14 & 85.5\% & 0.071 & 82.0\% \\
  $S_F$, $S_X$, $S_C$                 & 1.96 & \textbf{86.0\%} & 0.061 & 82.8\% \\
  \rowcolor{Gray}
  $S_F$, $S_X$, $S_C$, $S_M$          & \textbf{1.86} & \textbf{85.9\%} & \textbf{0.053} & \textbf{86.2\%} \\
  $S_F$, $S_X$, $S_C$, $S_M$, $S_H$   & 2.30 & 82.1\% & 0.065 & 82.6\% \\
  \hline
  \end{tabular}
\end{table}

\vspace{5pt} \noindent
\textbf{Early/Late-Fusion vs. Multi-Stage Fusion.} Another highlight of our fusion pipeline is multi-stage fusion. In Tab. \ref{tab:ablation-multi-stage}, we verify the effectiveness of each fusion stage.

For CamLiPWC, we denote its three stages (feature encoder, correlation, and flow estimator) as $S_F$, $S_C$, and $S_D$ respectively. The top row denotes the version where no fusion connection exists between the two branches. As we can see from the table, single-stage fusion can only provide sub-optimal results. In contrast, fusing features at all three stages brings significant improvements. We also provide a more detailed qualitative analysis in Fig. \ref{fig:multi-stage-ablation}. As we can see, fusing encoder features makes the structure of the objects clearer, since the point pyramid encodes geometric information which helps to recover the shape of complex objects. Fusing the correlation feature enables the model to capture small and fast-moving objects, since the point-based 3D correlation searches for a dynamic range of neighborhoods and can be complementary to the 2D correlation which searches for a fixed range. Fusing the features of the flow decoder improves overall performance, especially in complex scenes where objects overlap.

For CamLiRAFT, we use $S_F$, $S_X$, $S_C$, and $S_M$ to denote each of the four fusion locations: feature encoder, context encoder, correlation lookup, and motion encoder. As above, multi-stage fusion leads to better performance than single-stage fusion. However, we find that fusing the hidden status of the GRU (denoted by $S_H$ in the table) can degrade the performance. Thus, only the first four stages $S_F$, $S_X$, $S_C$, and $S_M$ are fused.

\begin{table}[t]
  \small
  \renewcommand{\arraystretch}{1.15}
  \caption{Ablations on the interpolation module of Bi-CLFM. We only report 2D metrics here since 3D metrics are all similar.}
  \label{tab:ablation-clfm-interpolation}
  \centering
  \begin{tabular}{ccc|cc}
  \hline
  \multicolumn{3}{c|}{Configurations} & \multicolumn{2}{c}{2D Metrics} \\
  $k$-NN & ScoreNet & SOP & EPE\textsubscript{2D} & ACC\textsubscript{1px} \\
  \hline
  $k=0$ & - & -                   & 2.61 & 81.6\% \\
  $k=1$ & - & -                   & 2.60 & 81.8\% \\
  \rowcolor{Gray}
  $k=1$ & $\surd$ & -             & \textbf{2.52} & \textbf{82.2\%} \\
  $k=3$ & $\surd$ & \texttt{MEAN} & 2.54 & 82.0\% \\
  $k=3$ & $\surd$ & \texttt{MAX}  & 2.54 & \textbf{82.2\%} \\
  \hline
  \end{tabular}
\end{table}

\vspace{5pt} \noindent
\textbf{Learnable Interpolation.} In Tab. \ref{tab:ablation-clfm-interpolation}, we test the interpolation module of Bi-CLFM with different configurations. The top row denotes a naive implementation that simply projects 3D features onto the image plane without interpolation (empty locations are filled with zeros). Directly densifying the feature map with nearest-neighbor interpolation will not bring significant improvement (see the second row). By introducing a learnable ScoreNet into the nearest-neighbor interpolation (the third row), we reduce EPE\textsubscript{2D} from 2.60 to 2.52, and improve ACC\textsubscript{1px} from 81.8\% to 82.2\%, demonstrating that learnable interpolation is a key design. We also conduct two experiments by increasing $k$ (the number of nearest neighbors) from 1 to 3, followed by a symmetric operation such as \texttt{MEAN} and \texttt{MAX}. However, no significant improvement is observed, suggesting that $k=1$ is enough for interpolation.

Since the improvement of learnable interpolation over the naive baseline looks marginal on FlyingThings3D, we further compare the generalization performance using CamLiRAFT in Tab. \ref{tab:ablation-kitti}. All variants are trained on FlyingThings3D and tested on the training set of KITTI without finetuning. As we can see, learnable interpolation significantly outperforms the baseline, especially for the 2D metrics (EPE\textsubscript{2D}: 3.64 $\rightarrow$ 3.03), demonstrating its good generalization performance.

\begin{table}[t]
  \small
  \renewcommand{\arraystretch}{1.15}
  \caption{Ablations on the feature fusion module of Bi-CLFM. Selective feature fusion significantly improves the 3D metrics.}
  \label{tab:ablation-sk-fusion}
  \centering
  \begin{tabular}{c|cc|cc}
  \hline
  \multirow{2}{*}{Methods} & \multicolumn{2}{c|}{2D Metrics} & \multicolumn{2}{c}{3D Metrics} \\
  & EPE\textsubscript{2D} & ACC\textsubscript{1px} & EPE\textsubscript{3D} & ACC\textsubscript{.05} \\
  \hline
  Addition    & 2.53 & 82.0\% & 0.074 & 80.9\% \\
  Concatenation & \textbf{2.52} & \textbf{82.4\%} & 0.075 & 80.5\% \\
  Gated  & 2.53 & 82.1\% & 0.075 & 80.4\% \\
  \hline
  Selective ($r$=4) & 2.53 & 82.0\% & 0.071 & 81.1\% \\
  \rowcolor{Gray}
  Selective ($r$=2) & \textbf{2.52} & 82.2\% & \textbf{0.070} & \textbf{82.3\%} \\
  \hline
  \end{tabular}
\end{table}

\vspace{5pt} \noindent
\textbf{Feature Fusion.} In Sec. \ref{sec:feature-fusion}, we propose to fuse features from different modalities adaptively based on the Selective Kernel Networks (SKNet). Here, we compare selective fusion with other common feature fusion operations including addition, concatenation, and gated fusion \cite{xu2021rpvnet}. Gated fusion is also an attention-based fusion method. The difference is that gated fusion leverages spatial attention to adaptively merge the features for each position, while selective fusion pays its attention to the channels. As we can see from Tab. \ref{tab:ablation-sk-fusion}, compared with addition and concatenation, gated fusion does not bring any improvement. Selective fusion is slightly better for 3D metrics (EPE\textsubscript{3D}: 0.075 $\rightarrow$ 0.070), but tied with concatenation for 2D metrics. In Fig. \ref{tab:ablation-kitti}, we further test the generalization performance among these fusion methods using CamLiRAFT. Under this setting, selective fusion achieves absolute advantages compared to concatenation and other methods. This result shows that channel attention brings better generalization performance, which is more important for flow tasks.

\begin{table}[t]
  \small
  \renewcommand{\arraystretch}{1.15}
  \setlength{\tabcolsep}{4pt}
  \caption{Additional ablations on the feature fusion and learnable interpolation (``LearnInterp'' for short) module. We test the generalization performance on KITTI (train) after training on FlyingThings3D.}
  \label{tab:ablation-kitti}
  \centering
  \begin{tabular}{c|c|cc|cc}
  \hline
  \multirow{2}{*}{Fusion} & Learn- & \multicolumn{2}{c|}{2D Metrics (K)} & \multicolumn{2}{c}{3D Metrics (K)} \\
  & Interp & EPE\textsubscript{2D} & ACC\textsubscript{1px} & EPE\textsubscript{3D} & ACC\textsubscript{.05} \\
  \hline
  Addition      & $\surd$ & 3.92 & 64.8\% & 0.108 & 53.9\% \\
  Concatenation & $\surd$ & 3.25 & 65.9\% & 0.091 & 58.5\% \\
  Gated         & $\surd$ & 3.51 & 64.1\% & 0.101 & 55.5\% \\
  Selective     & -       & 3.64 & 62.6\% & 0.102 & 55.6\% \\
  \rowcolor{Gray}
  Selective     & $\surd$ & \textbf{3.03} & \textbf{67.0\%} & \textbf{0.087} & \textbf{59.8\%} \\
  \hline
  \end{tabular}
\end{table}

\begin{table}[t]
  \small
  \renewcommand{\arraystretch}{1.15}
  \caption{Ablations on inverse depth scaling (IDS). ``-L'' denotes the LiDAR-only variant which removes the image branch.}
  \label{tab:ablation-ids}
  \centering
  \begin{tabular}{l|c|cc}
  \hline
  \multirow{2}{*}{Models} & \multirow{2}{*}{IDS} & \multicolumn{2}{c}{3D Metrics} \\
  & & EPE\textsubscript{3D} & ACC\textsubscript{.05} \\
  \hline
  \multirow{2}{*}{CamLiPWC-L} & - & 0.123 & 58.1\% \\
  & $\surd$ & \textbf{0.116} & \textbf{61.6\%} \\
  \hline
  \multirow{2}{*}{CamLiRAFT-L} & - & 0.105 & 67.6\% \\
  & $\surd$ & \textbf{0.096} & \textbf{71.7\%} \\
  \hline
  \end{tabular}
\end{table}

\begin{figure}[t]
  \centering{
    \includegraphics[width=0.49\linewidth]{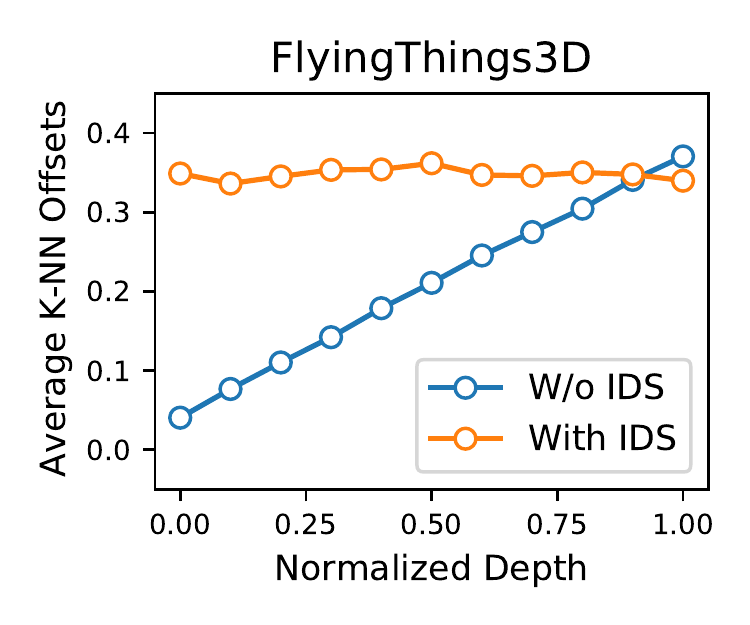}
    \includegraphics[width=0.49\linewidth]{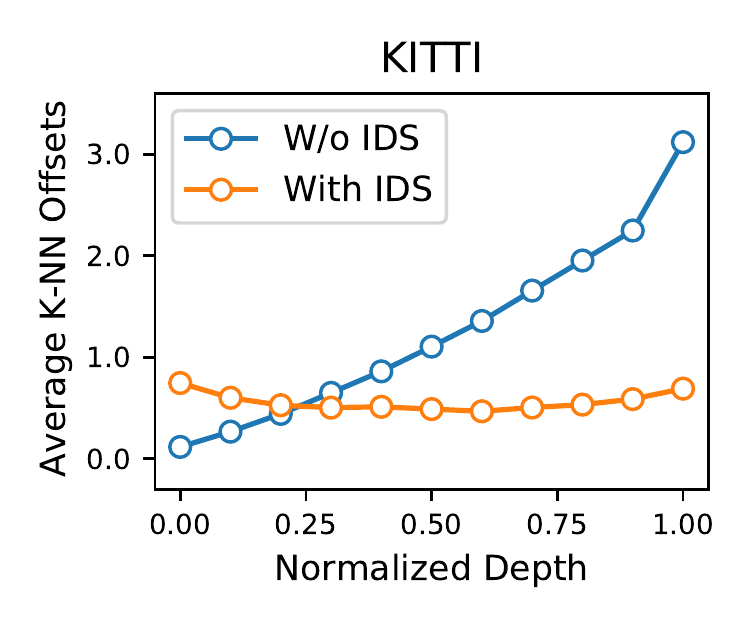}
  }
  \vspace{-15pt}
  \caption{The density of point clouds across different distances with/ without IDS. We measure the local density around a point by averaging the offsets of its $k$ nearest neighbors. As shown by the orange line, IDS makes the distribution of points more even across different regions.}
  \label{fig:ids-stat}
\end{figure}

\vspace{6pt} \noindent
\textbf{Inverse Depth Scaling.} In Fig. \ref{fig:ids-stat}, we carry out a statistical analysis on FlyingThings3D and the raw Velodyne data of KITTI to illustrate the density of points across varying distances with and without IDS. The local density around a point is gauged by averaging the offsets of its $k$ nearest neighbors ($k=16$). As can be seen, IDS leads to a more uniform distribution of points across different regions. We also conduct several comparison experiments on FlyingThings3D to verify the effects of IDS. Since IDS does not require input images, we also test it on CamLiPWC-L and CamLiRAFT-L where the image branch is removed. As shown in Tab. \ref{tab:ablation-ids}, the performance is improved under the LiDAR-only setup, suggesting that a more even distribution of points can facilitate the learning.

\begin{figure}[t]
  \centering
  \vspace{-5pt}
  \subfloat{\includegraphics[width=0.5\linewidth]{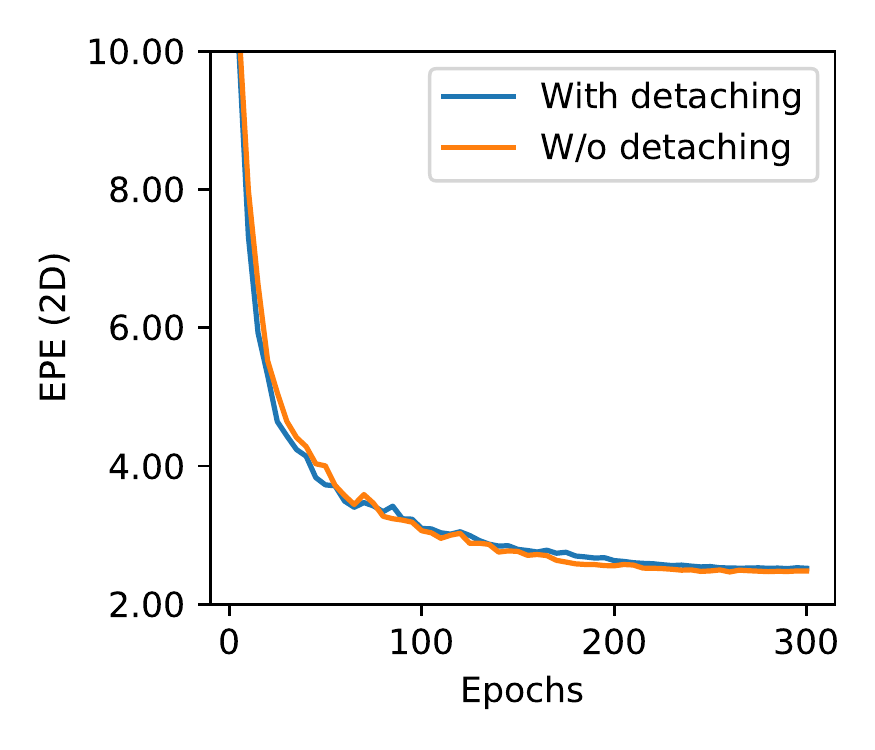}}%
  \hfill
  \subfloat{\includegraphics[width=0.5\linewidth]{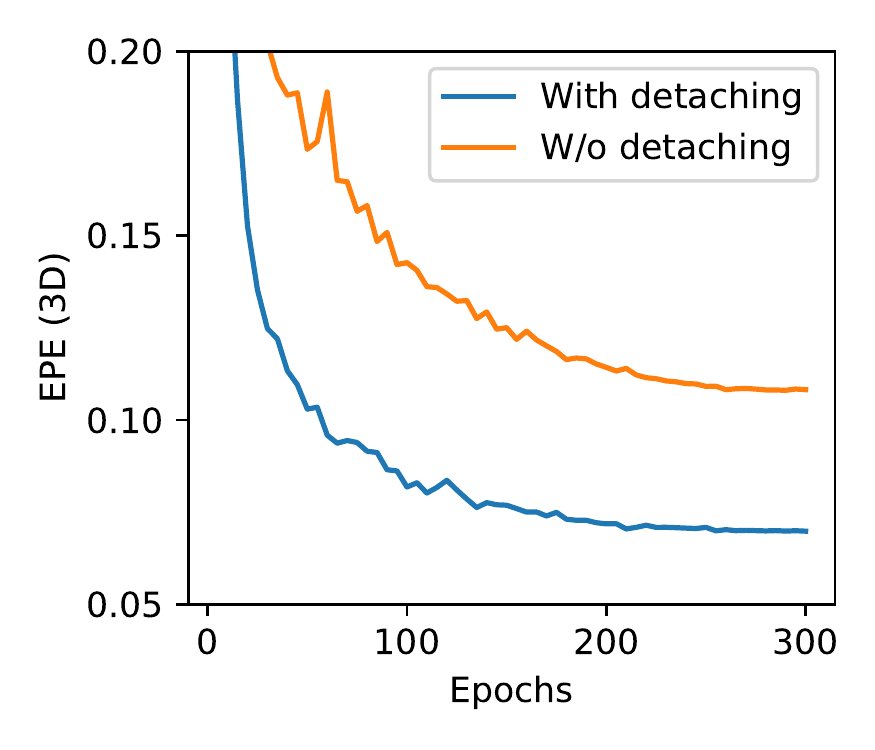}}%
  \vspace{-5pt}
  \caption{Performance of CamLiPWC on FlyingThings3D with/without our gradient detaching strategy. Detaching the gradient from one branch to the other can prevent one modality from dominating so that each branch focuses on its task.}
  \label{fig:grad_detach}
\end{figure}

\vspace{6pt} \noindent
\textbf{Gradient Detaching.} In Fig. \ref{fig:grad_detach}, we conduct an ablation on CamLiPWC to demonstrate the effect of gradient detaching. The 2D branch is not affected by the 3D gradient, but the 3D branch suffers a lot from the 2D gradient. This is probably because the 2D gradient is so large that it dominates the training (see Fig. \ref{fig:grad_norm}). By detaching the gradient from one branch to the other, we prevent one modality from dominating so that each branch focuses on its task.

\begin{table}[t]
  \small
  \renewcommand{\arraystretch}{1.15}
  \caption{Breakdown of the inference latency. Results are computed on 960x540 images with a Tesla V100 GPU (\texttt{fp32} PyTorch backend).}
  \label{tab:runtime}
  \centering
  \begin{tabular}{l|l|c|c}
    \hline
    \multirow{2}{*}{Models} & \multirow{2}{*}{Component} & Individual & Total \\
    & & Latency & Latency \\
    \hline
    \multirow{4}{*}{CamLiPWC} & Image branch & 42ms & \multirow{4}{*}{110ms} \\
    & Point branch & 31ms & \\
    & Bi-CLFM & 33ms & \\
    & Others & 4ms & \\
    \hline
    \multirow{4}{*}{CamLiRAFT} & Image branch & 106ms & \multirow{4}{*}{216ms} \\
    & Point branch & 79ms & \\
    & Bi-CLFM & 23ms & \\
    & Others & 8ms & \\
    \hline
  \end{tabular}
\end{table}

\subsection{Latency and Limitations}

\vspace{5pt} \noindent
\textbf{Inference Latency.} In Tab. \ref{tab:runtime}, we provide a breakdown of the inference latency on FlyingThings3D (960x540 images and 8192 points). Timing results are measured on a single Tesla V100 GPU with a \texttt{fp32} PyTorch \cite{paszke2019pytorch} backend. For CamLiRAFT, the number of iterative updates is set to 8 for efficiency. As we can see, the computation cost introduced by Bi-CLFM only accounts for 10.6\% of the total. The inference takes 110ms for CamLiPWC and 216ms for CamLiRAFT. Note that we do \textit{not} use any inference acceleration technique like TensorRT. Thus, there is potential room for further improvements.

\vspace{5pt} \noindent
\textbf{Limitations.} Since Bi-CLFM makes a hard association of LiDAR point clouds and camera images, the performance will be damaged when sensor misalignment appears. In the future, we will try to adaptively align the two modalities without establishing deterministic correspondence using the projection matrix. Another limitation is that CamLiRAFT is not very robust to objects at a greater distance, as the network has only been trained on data with a depth of less than 35m.

\section{Conclusion}

In this paper, we propose a bidirectional and multi-stage fusion pipeline that utilizes the complementarity between the camera and LiDAR. To fuse dense image features and sparse point features, we introduce the bidirectional camera-LiDAR fusion module (Bi-CLFM), a learnable operator which aligns and fuses the features in a bidirectional manner via learnable interpolation and bilinear sampling. With the proposed Bi-CLFM, we build CamLiPWC based on the pyramidal coarse-to-fine strategy and CamLiRAFT based on the recurrent all-pairs field transforms. Experiments show that our fusion pipeline is general and achieve consistent improvements across different architectures. Moreover, we achieve new state-of-the-art performance on FlyingThings3D and KITTI, which demonstrates the effectiveness of our fusion pipeline. Our methods also have good generalization performance and can handle generic non-rigid motion. We hope that this exciting result draws more attention to the bidirectional fusion paradigm. 


%

\ifCLASSOPTIONcompsoc
  \section*{Acknowledgments}
\else
  \section*{Acknowledgment}
\fi

This work is supported by the National Key R$\&$D Program of China (No. 2022ZD0160900), the National Natural Science Foundation of China (No. 62076119, No. 61921006), the Fundamental Research Funds for the Central Universities (No. 020214380091), and Collaborative Innovation Center of Novel Software Technology and Industrialization. The authors would like to thank the anonymous reviewers for their valuable suggestions and Hanxiao Xie for her help in promoting this work.

\ifCLASSOPTIONcaptionsoff
  \newpage
\fi



%
\bibliographystyle{IEEEtranS}
\bibliography{egbib.bib}

\begin{thebibliography}{10}
\providecommand{\url}[1]{#1}
\csname url@samestyle\endcsname
\providecommand{\newblock}{\relax}
\providecommand{\bibinfo}[2]{#2}
\providecommand{\BIBentrySTDinterwordspacing}{\spaceskip=0pt\relax}
\providecommand{\BIBentryALTinterwordstretchfactor}{4}
\providecommand{\BIBentryALTinterwordspacing}{\spaceskip=\fontdimen2\font plus
\BIBentryALTinterwordstretchfactor\fontdimen3\font minus
  \fontdimen4\font\relax}
\providecommand{\BIBforeignlanguage}[2]{{%
\expandafter\ifx\csname l@#1\endcsname\relax
\typeout{** WARNING: IEEEtranS.bst: No hyphenation pattern has been}%
\typeout{** loaded for the language `#1'. Using the pattern for}%
\typeout{** the default language instead.}%
\else
\language=\csname l@#1\endcsname
\fi
#2}}
\providecommand{\BIBdecl}{\relax}
\BIBdecl

\bibitem{aleotti2020learning}
F.~Aleotti, M.~Poggi, F.~Tosi, and S.~Mattoccia, ``Learning end-to-end scene
  flow by distilling single tasks knowledge,'' in \emph{Proceedings of the AAAI
  Conference on Artificial Intelligence}, vol.~34, no.~07, 2020, pp.
  10\,435--10\,442.

\bibitem{badki2021binaryttc}
A.~Badki, O.~Gallo, J.~Kautz, and P.~Sen, ``Binary ttc: A temporal geofence for
  autonomous navigation,'' in \emph{Proceedings of the IEEE/CVF Conference on
  Computer Vision and Pattern Recognition}, 2021, pp. 12\,946--12\,955.

\bibitem{bai2022transfusion}
X.~Bai, Z.~Hu, X.~Zhu, Q.~Huang, Y.~Chen, H.~Fu, and C.-L. Tai, ``Transfusion:
  Robust lidar-camera fusion for 3d object detection with transformers,'' in
  \emph{Proceedings of the IEEE/CVF Conference on Computer Vision and Pattern
  Recognition}, 2022, pp. 1090--1099.

\bibitem{battrawy2019lidarflow}
R.~Battrawy, R.~Schuster, O.~Wasenm{\"u}ller, Q.~Rao, and D.~Stricker,
  ``Lidar-flow: Dense scene flow estimation from sparse lidar and stereo
  images,'' in \emph{2019 IEEE/RSJ International Conference on Intelligent
  Robots and Systems (IROS)}.\hskip 1em plus 0.5em minus 0.4em\relax IEEE,
  2019, pp. 7762--7769.

\bibitem{bayramli2023raft}
B.~Bayramli, J.~Hur, and H.~Lu, ``Raft-msf: Self-supervised monocular scene
  flow using recurrent optimizer,'' \emph{International Journal of Computer
  Vision}, pp. 1--13, 2023.

\bibitem{behl2017isf}
A.~Behl, O.~Hosseini~Jafari, S.~Karthik~Mustikovela, H.~Abu~Alhaija, C.~Rother,
  and A.~Geiger, ``Bounding boxes, segmentations and object coordinates: How
  important is recognition for 3d scene flow estimation in autonomous driving
  scenarios?'' in \emph{Proceedings of the IEEE International Conference on
  Computer Vision}, 2017, pp. 2574--2583.

\bibitem{black1996robust}
M.~J. Black and P.~Anandan, ``The robust estimation of multiple motions:
  Parametric and piecewise-smooth flow fields,'' \emph{Computer vision and
  image understanding}, vol.~63, no.~1, pp. 75--104, 1996.

\bibitem{brachmann2019ransac}
E.~Brachmann and C.~Rother, ``Neural-guided ransac: Learning where to sample
  model hypotheses,'' in \emph{Proceedings of the IEEE/CVF International
  Conference on Computer Vision}, 2019, pp. 4322--4331.

\bibitem{brox2009large}
T.~Brox, C.~Bregler, and J.~Malik, ``Large displacement optical flow,'' in
  \emph{2009 IEEE Conference on Computer Vision and Pattern Recognition}.\hskip
  1em plus 0.5em minus 0.4em\relax IEEE, 2009, pp. 41--48.

\bibitem{brox2004warping}
T.~Brox, A.~Bruhn, N.~Papenberg, and J.~Weickert, ``High accuracy optical flow
  estimation based on a theory for warping,'' in \emph{European conference on
  computer vision}.\hskip 1em plus 0.5em minus 0.4em\relax Springer, 2004, pp.
  25--36.

\bibitem{bruhn2005lucas}
A.~Bruhn, J.~Weickert, and C.~Schn{\"o}rr, ``Lucas/kanade meets horn/schunck:
  Combining local and global optic flow methods,'' \emph{International journal
  of computer vision}, vol.~61, no.~3, pp. 211--231, 2005.

\bibitem{sintel}
D.~J. Butler, J.~Wulff, G.~B. Stanley, and M.~J. Black, ``A naturalistic open
  source movie for optical flow evaluation,'' in \emph{Computer Vision--ECCV
  2012: 12th European Conference on Computer Vision, Florence, Italy, October
  7-13, 2012, Proceedings, Part VI 12}.\hskip 1em plus 0.5em minus 0.4em\relax
  Springer, 2012, pp. 611--625.

\bibitem{chang2015shapenet}
A.~X. Chang, T.~Funkhouser, L.~Guibas, P.~Hanrahan, Q.~Huang, Z.~Li,
  S.~Savarese, M.~Savva, S.~Song, H.~Su \emph{et~al.}, ``Shapenet: An
  information-rich 3d model repository,'' \emph{arXiv preprint
  arXiv:1512.03012}, 2015.

\bibitem{cordts2016cityscapes}
M.~Cordts, M.~Omran, S.~Ramos, T.~Rehfeld, M.~Enzweiler, R.~Benenson,
  U.~Franke, S.~Roth, and B.~Schiele, ``The cityscapes dataset for semantic
  urban scene understanding,'' in \emph{Proceedings of the IEEE conference on
  computer vision and pattern recognition}, 2016, pp. 3213--3223.

\bibitem{dosovitskiy2015flownet}
A.~Dosovitskiy, P.~Fischer, E.~Ilg, P.~Hausser, C.~Hazirbas, V.~Golkov, P.~Van
  Der~Smagt, D.~Cremers, and T.~Brox, ``Flownet: Learning optical flow with
  convolutional networks,'' in \emph{Proceedings of the IEEE international
  conference on computer vision}, 2015, pp. 2758--2766.

\bibitem{feng2021advancing}
\BIBentryALTinterwordspacing
Z.~Feng, L.~Jing, P.~Yin, Y.~Tian, and B.~Li, ``Advancing self-supervised
  monocular depth learning with sparse li{DAR},'' in \emph{5th Annual
  Conference on Robot Learning}, 2021. [Online]. Available:
  \url{https://openreview.net/forum?id=EdmeHU4WVjJ}
\BIBentrySTDinterwordspacing

\bibitem{gu2019hplflownet}
X.~Gu, Y.~Wang, C.~Wu, Y.~J. Lee, and P.~Wang, ``Hplflownet: Hierarchical
  permutohedral lattice flownet for scene flow estimation on large-scale point
  clouds,'' in \emph{Proceedings of the IEEE/CVF Conference on Computer Vision
  and Pattern Recognition}, 2019, pp. 3254--3263.

\bibitem{he2015resnet}
K.~He, X.~Zhang, S.~Ren, and J.~Sun, ``Deep residual learning for image
  recognition,'' 2015.

\bibitem{hong2021ddrnet}
Y.~Hong, H.~Pan, W.~Sun, Y.~Jia \emph{et~al.}, ``Deep dual-resolution networks
  for real-time and accurate semantic segmentation of road scenes,''
  \emph{arXiv preprint arXiv:2101.06085}, 2021.

\bibitem{horn1981determining}
B.~K. Horn and B.~G. Schunck, ``Determining optical flow,'' \emph{Artificial
  intelligence}, vol.~17, no. 1-3, pp. 185--203, 1981.

\bibitem{howard2017mobilenets}
A.~G. Howard, M.~Zhu, B.~Chen, D.~Kalenichenko, W.~Wang, T.~Weyand,
  M.~Andreetto, and H.~Adam, ``Mobilenets: Efficient convolutional neural
  networks for mobile vision applications,'' \emph{arXiv preprint
  arXiv:1704.04861}, 2017.

\bibitem{hui2018liteflownet}
T.-W. Hui, X.~Tang, and C.~C. Loy, ``Liteflownet: A lightweight convolutional
  neural network for optical flow estimation,'' in \emph{Proceedings of the
  IEEE conference on computer vision and pattern recognition}, 2018, pp.
  8981--8989.

\bibitem{hur2019iterative}
J.~Hur and S.~Roth, ``Iterative residual refinement for joint optical flow and
  occlusion estimation,'' in \emph{Proceedings of the IEEE/CVF Conference on
  Computer Vision and Pattern Recognition}, 2019, pp. 5754--5763.

\bibitem{hur2020self}
------, ``Self-supervised monocular scene flow estimation,'' in
  \emph{Proceedings of the IEEE/CVF Conference on Computer Vision and Pattern
  Recognition}, 2020, pp. 7396--7405.

\bibitem{hur2021self}
------, ``Self-supervised multi-frame monocular scene flow,'' in
  \emph{Proceedings of the IEEE/CVF Conference on Computer Vision and Pattern
  Recognition}, 2021, pp. 2684--2694.

\bibitem{ilg2017flownet2}
E.~Ilg, N.~Mayer, T.~Saikia, M.~Keuper, A.~Dosovitskiy, and T.~Brox, ``Flownet
  2.0: Evolution of optical flow estimation with deep networks,'' in
  \emph{Proceedings of the IEEE conference on computer vision and pattern
  recognition}, 2017, pp. 2462--2470.

\bibitem{ioffe2015batchnorm}
S.~Ioffe and C.~Szegedy, ``Batch normalization: Accelerating deep network
  training by reducing internal covariate shift,'' 2015.

\bibitem{jaimez2015primal}
M.~Jaimez, M.~Souiai, J.~Gonzalez-Jimenez, and D.~Cremers, ``A primal-dual
  framework for real-time dense rgb-d scene flow,'' in \emph{2015 IEEE
  international conference on robotics and automation (ICRA)}.\hskip 1em plus
  0.5em minus 0.4em\relax IEEE, 2015, pp. 98--104.

\bibitem{jaimez2015motion}
M.~Jaimez, M.~Souiai, J.~St{\"u}ckler, J.~Gonzalez-Jimenez, and D.~Cremers,
  ``Motion cooperation: Smooth piece-wise rigid scene flow from rgb-d images,''
  in \emph{2015 International Conference on 3D Vision}.\hskip 1em plus 0.5em
  minus 0.4em\relax IEEE, 2015, pp. 64--72.

\bibitem{jiang2019sense}
H.~Jiang, D.~Sun, V.~Jampani, Z.~Lv, E.~Learned-Miller, and J.~Kautz, ``Sense:
  A shared encoder network for scene-flow estimation,'' in \emph{Proceedings of
  the IEEE/CVF International Conference on Computer Vision}, 2019, pp.
  3195--3204.

\bibitem{kittenplon2021flowstep3d}
Y.~Kittenplon, Y.~C. Eldar, and D.~Raviv, ``Flowstep3d: Model unrolling for
  self-supervised scene flow estimation,'' in \emph{Proceedings of the IEEE/CVF
  Conference on Computer Vision and Pattern Recognition}, 2021, pp. 4114--4123.

\bibitem{krizhevsky2012imagenet}
A.~Krizhevsky, I.~Sutskever, and G.~E. Hinton, ``Imagenet classification with
  deep convolutional neural networks,'' \emph{Advances in neural information
  processing systems}, vol.~25, 2012.

\bibitem{li2021acosf}
C.~Li, H.~Ma, and Q.~Liao, ``Two-stage adaptive object scene flow using hybrid
  cnn-crf model,'' in \emph{2020 25th International Conference on Pattern
  Recognition (ICPR)}.\hskip 1em plus 0.5em minus 0.4em\relax IEEE, 2021, pp.
  3876--3883.

\bibitem{li2019sknet}
X.~Li, W.~Wang, X.~Hu, and J.~Yang, ``Selective kernel networks,'' in
  \emph{Proceedings of the IEEE/CVF Conference on Computer Vision and Pattern
  Recognition}, 2019, pp. 510--519.

\bibitem{liang2018continuous}
M.~Liang, B.~Yang, S.~Wang, and R.~Urtasun, ``Deep continuous fusion for
  multi-sensor 3d object detection,'' in \emph{Proceedings of the European
  Conference on Computer Vision (ECCV)}, 2018, pp. 641--656.

\bibitem{liu2022camliflow}
H.~Liu, T.~Lu, Y.~Xu, J.~Liu, W.~Li, and L.~Chen, ``Camliflow: bidirectional
  camera-lidar fusion for joint optical flow and scene flow estimation,'' in
  \emph{Proceedings of the IEEE/CVF Conference on Computer Vision and Pattern
  Recognition}, 2022, pp. 5791--5801.

\bibitem{liu2019flownet3d}
X.~Liu, C.~R. Qi, and L.~J. Guibas, ``Flownet3d: Learning scene flow in 3d
  point clouds,'' in \emph{Proceedings of the IEEE/CVF Conference on Computer
  Vision and Pattern Recognition}, 2019, pp. 529--537.

\bibitem{liu2019meteornet}
X.~Liu, M.~Yan, and J.~Bohg, ``Meteornet: Deep learning on dynamic 3d point
  cloud sequences,'' in \emph{Proceedings of the IEEE/CVF International
  Conference on Computer Vision}, 2019, pp. 9246--9255.

\bibitem{liu2022bevfusion}
Z.~Liu, H.~Tang, A.~Amini, X.~Yang, H.~Mao, D.~Rus, and S.~Han, ``Bevfusion:
  Multi-task multi-sensor fusion with unified bird's-eye view representation,''
  \emph{arXiv}, 2022.

\bibitem{ma2018sparse}
F.~Ma and S.~Karaman, ``Sparse-to-dense: Depth prediction from sparse depth
  samples and a single image,'' in \emph{2018 IEEE international conference on
  robotics and automation (ICRA)}.\hskip 1em plus 0.5em minus 0.4em\relax IEEE,
  2018, pp. 4796--4803.

\bibitem{ma2019drisf}
W.-C. Ma, S.~Wang, R.~Hu, Y.~Xiong, and R.~Urtasun, ``Deep rigid instance scene
  flow,'' in \emph{Proceedings of the IEEE/CVF Conference on Computer Vision
  and Pattern Recognition}, 2019, pp. 3614--3622.

\bibitem{mayer2016things3d}
N.~Mayer, E.~Ilg, P.~Hausser, P.~Fischer, D.~Cremers, A.~Dosovitskiy, and
  T.~Brox, ``A large dataset to train convolutional networks for disparity,
  optical flow, and scene flow estimation,'' in \emph{Proceedings of the IEEE
  conference on computer vision and pattern recognition}, 2016, pp. 4040--4048.

\bibitem{menze2015osf}
M.~Menze and A.~Geiger, ``Object scene flow for autonomous vehicles,'' in
  \emph{Proceedings of the IEEE conference on computer vision and pattern
  recognition}, 2015, pp. 3061--3070.

\bibitem{ouyang2021ogsf}
B.~Ouyang and D.~Raviv, ``Occlusion guided scene flow estimation on 3d point
  clouds,'' in \emph{Proceedings of the IEEE/CVF Conference on Computer Vision
  and Pattern Recognition (CVPR) Workshops}, June 2021, pp. 2805--2814.

\bibitem{paszke2019pytorch}
A.~Paszke, S.~Gross, F.~Massa, A.~Lerer, J.~Bradbury, G.~Chanan, T.~Killeen,
  Z.~Lin, N.~Gimelshein, L.~Antiga \emph{et~al.}, ``Pytorch: An imperative
  style, high-performance deep learning library,'' \emph{Advances in neural
  information processing systems}, vol.~32, pp. 8026--8037, 2019.

\bibitem{poggi2021sensor}
M.~Poggi, F.~Aleotti, and S.~Mattoccia, ``Sensor-guided optical flow,'' in
  \emph{Proceedings of the IEEE/CVF International Conference on Computer
  Vision}, 2021, pp. 7908--7918.

\bibitem{puy2020flot}
G.~Puy, A.~Boulch, and R.~Marlet, ``Flot: Scene flow on point clouds guided by
  optimal transport,'' in \emph{Computer Vision--ECCV 2020: 16th European
  Conference, Glasgow, UK, August 23--28, 2020, Proceedings, Part XXVIII
  16}.\hskip 1em plus 0.5em minus 0.4em\relax Springer, 2020, pp. 527--544.

\bibitem{qi2018frustum}
C.~R. Qi, W.~Liu, C.~Wu, H.~Su, and L.~J. Guibas, ``Frustum pointnets for 3d
  object detection from rgb-d data,'' in \emph{Proceedings of the IEEE
  conference on computer vision and pattern recognition}, 2018, pp. 918--927.

\bibitem{qi2017pointnet}
C.~R. Qi, H.~Su, K.~Mo, and L.~J. Guibas, ``Pointnet: Deep learning on point
  sets for 3d classification and segmentation,'' in \emph{Proceedings of the
  IEEE conference on computer vision and pattern recognition}, 2017, pp.
  652--660.

\bibitem{qi2017pointnet++}
C.~R. Qi, L.~Yi, H.~Su, and L.~J. Guibas, ``Pointnet++: Deep hierarchical
  feature learning on point sets in a metric space,'' \emph{arXiv preprint
  arXiv:1706.02413}, 2017.

\bibitem{quiroga2014dense}
J.~Quiroga, T.~Brox, F.~Devernay, and J.~Crowley, ``Dense semi-rigid scene flow
  estimation from rgbd images,'' in \emph{European Conference on Computer
  Vision}.\hskip 1em plus 0.5em minus 0.4em\relax Springer, 2014, pp. 567--582.

\bibitem{ranjan2017spynet}
A.~Ranjan and M.~J. Black, ``Optical flow estimation using a spatial pyramid
  network,'' in \emph{Proceedings of the IEEE conference on computer vision and
  pattern recognition}, 2017, pp. 4161--4170.

\bibitem{ren2017ssf}
Z.~Ren, D.~Sun, J.~Kautz, and E.~Sudderth, ``Cascaded scene flow prediction
  using semantic segmentation,'' in \emph{2017 International Conference on 3D
  Vision (3DV)}.\hskip 1em plus 0.5em minus 0.4em\relax IEEE, 2017, pp.
  225--233.

\bibitem{rishav2020deeplidarflow}
R.~Rishav, R.~Battrawy, R.~Schuster, O.~Wasenm{\"u}ller, and D.~Stricker,
  ``Deeplidarflow: A deep learning architecture for scene flow estimation using
  monocular camera and sparse lidar,'' in \emph{2020 IEEE/RSJ International
  Conference on Intelligent Robots and Systems (IROS)}.\hskip 1em plus 0.5em
  minus 0.4em\relax IEEE, 2020, pp. 10\,460--10\,467.

\bibitem{sindagi2019mvx}
V.~A. Sindagi, Y.~Zhou, and O.~Tuzel, ``Mvx-net: Multimodal voxelnet for 3d
  object detection,'' in \emph{2019 International Conference on Robotics and
  Automation (ICRA)}.\hskip 1em plus 0.5em minus 0.4em\relax IEEE, 2019, pp.
  7276--7282.

\bibitem{sun2018pwc}
D.~Sun, X.~Yang, M.-Y. Liu, and J.~Kautz, ``Pwc-net: Cnns for optical flow
  using pyramid, warping, and cost volume,'' in \emph{Proceedings of the IEEE
  conference on computer vision and pattern recognition}, 2018, pp. 8934--8943.

\bibitem{teed2020raft}
Z.~Teed and J.~Deng, ``Raft: Recurrent all-pairs field transforms for optical
  flow,'' in \emph{European conference on computer vision}.\hskip 1em plus
  0.5em minus 0.4em\relax Springer, 2020, pp. 402--419.

\bibitem{teed2021raft3d}
------, ``Raft-3d: Scene flow using rigid-motion embeddings,'' in
  \emph{Proceedings of the IEEE/CVF Conference on Computer Vision and Pattern
  Recognition}, 2021, pp. 8375--8384.

\bibitem{vogel2015prsm}
C.~Vogel, K.~Schindler, and S.~Roth, ``3d scene flow estimation with a
  piecewise rigid scene model,'' \emph{International Journal of Computer
  Vision}, vol. 115, no.~1, pp. 1--28, 2015.

\bibitem{vora2020pointpainting}
S.~Vora, A.~H. Lang, B.~Helou, and O.~Beijbom, ``Pointpainting: Sequential
  fusion for 3d object detection,'' in \emph{Proceedings of the IEEE/CVF
  conference on computer vision and pattern recognition}, 2020, pp. 4604--4612.

\bibitem{wang2020flownet3d++}
Z.~Wang, S.~Li, H.~Howard-Jenkins, V.~Prisacariu, and M.~Chen, ``Flownet3d++:
  Geometric losses for deep scene flow estimation,'' in \emph{Proceedings of
  the IEEE/CVF Winter Conference on Applications of Computer Vision}, 2020, pp.
  91--98.

\bibitem{wei2021pvraft}
Y.~Wei, Z.~Wang, Y.~Rao, J.~Lu, and J.~Zhou, ``Pv-raft: Point-voxel correlation
  fields for scene flow estimation of point clouds,'' in \emph{Proceedings of
  the IEEE/CVF Conference on Computer Vision and Pattern Recognition}, 2021,
  pp. 6954--6963.

\bibitem{weinzaepfel2013deepflow}
P.~Weinzaepfel, J.~Revaud, Z.~Harchaoui, and C.~Schmid, ``Deepflow: Large
  displacement optical flow with deep matching,'' in \emph{Proceedings of the
  IEEE international conference on computer vision}, 2013, pp. 1385--1392.

\bibitem{wu2019pointconv}
W.~Wu, Z.~Qi, and L.~Fuxin, ``Pointconv: Deep convolutional networks on 3d
  point clouds,'' in \emph{Proceedings of the IEEE/CVF Conference on Computer
  Vision and Pattern Recognition}, 2019, pp. 9621--9630.

\bibitem{wu2019pointpwc}
W.~Wu, Z.~Wang, Z.~Li, W.~Liu, and L.~Fuxin, ``Pointpwc-net: A coarse-to-fine
  network for supervised and self-supervised scene flow estimation on 3d point
  clouds,'' \emph{arXiv preprint arXiv:1911.12408}, 2019.

\bibitem{xu2018pointfusion}
D.~Xu, D.~Anguelov, and A.~Jain, ``Pointfusion: Deep sensor fusion for 3d
  bounding box estimation,'' in \emph{Proceedings of the IEEE conference on
  computer vision and pattern recognition}, 2018, pp. 244--253.

\bibitem{xu2021rpvnet}
J.~Xu, R.~Zhang, J.~Dou, Y.~Zhu, J.~Sun, and S.~Pu, ``Rpvnet: A deep and
  efficient range-point-voxel fusion network for lidar point cloud
  segmentation,'' in \emph{Proceedings of the IEEE/CVF International Conference
  on Computer Vision}, 2021, pp. 16\,024--16\,033.

\bibitem{yan2023cross}
J.~Yan, Y.~Liu, J.~Sun, F.~Jia, S.~Li, T.~Wang, and X.~Zhang, ``Cross modal
  transformer via coordinates encoding for 3d object dectection,'' \emph{arXiv
  preprint arXiv:2301.01283}, 2023.

\bibitem{yang2019volumetric}
G.~Yang and D.~Ramanan, ``Volumetric correspondence networks for optical
  flow,'' \emph{Advances in neural information processing systems}, vol.~32,
  pp. 794--805, 2019.

\bibitem{yang2020opticalexp}
------, ``Upgrading optical flow to 3d scene flow through optical expansion,''
  in \emph{Proceedings of the IEEE/CVF Conference on Computer Vision and
  Pattern Recognition}, 2020, pp. 1334--1343.

\bibitem{yang2021rigidmask}
------, ``Learning to segment rigid motions from two frames,'' in
  \emph{Proceedings of the IEEE/CVF Conference on Computer Vision and Pattern
  Recognition}, 2021, pp. 1266--1275.

\bibitem{yang2018ipod}
Z.~Yang, Y.~Sun, S.~Liu, X.~Shen, and J.~Jia, ``Ipod: Intensive point-based
  object detector for point cloud,'' \emph{arXiv preprint arXiv:1812.05276},
  2018.

\bibitem{you2019pseudo}
Y.~You, Y.~Wang, W.-L. Chao, D.~Garg, G.~Pleiss, B.~Hariharan, M.~Campbell, and
  K.~Q. Weinberger, ``Pseudo-lidar++: Accurate depth for 3d object detection in
  autonomous driving,'' \emph{arXiv preprint arXiv:1906.06310}, 2019.

\bibitem{zach2007duality}
C.~Zach, T.~Pock, and H.~Bischof, ``A duality based approach for realtime tv-l
  1 optical flow,'' in \emph{Joint pattern recognition symposium}.\hskip 1em
  plus 0.5em minus 0.4em\relax Springer, 2007, pp. 214--223.

\bibitem{zhang2019ganet}
F.~Zhang, V.~Prisacariu, R.~Yang, and P.~H. Torr, ``Ga-net: Guided aggregation
  net for end-to-end stereo matching,'' in \emph{Proceedings of the IEEE/CVF
  Conference on Computer Vision and Pattern Recognition}, 2019, pp. 185--194.

\bibitem{zhang2023gmsf}
Y.~Zhang, J.~Edstedt, B.~Wandt, P.-E. Forss{\'e}n, M.~Magnusson, and
  M.~Felsberg, ``Gmsf: Global matching scene flow,'' \emph{arXiv preprint
  arXiv:2305.17432}, 2023.

\end{thebibliography}

%

\begin{IEEEbiography}[{\includegraphics[width=1in,height=1.25in,clip,keepaspectratio]{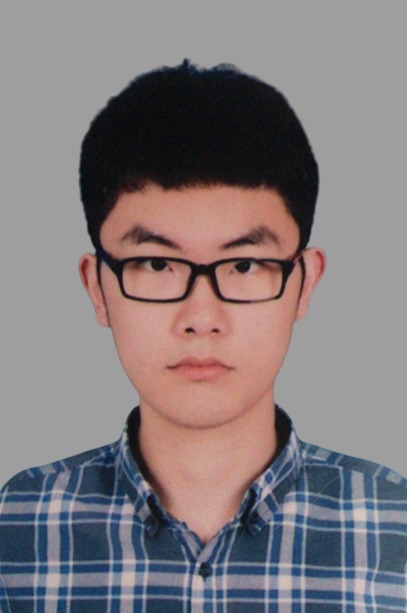}}]{Haisong Liu}
Haisong Liu is a Ph.D. student at the Department of Computer Science and Technology, Nanjing University. Before that, he obtained his B.E. degree in computer science from Xi'an Jiaotong University, Xi'an, China. His research interests lie in 3D computer vision, point cloud processing, and motion analysis.
\end{IEEEbiography}
  
\begin{IEEEbiography}[{\includegraphics[width=1in,height=1.25in,clip,keepaspectratio]{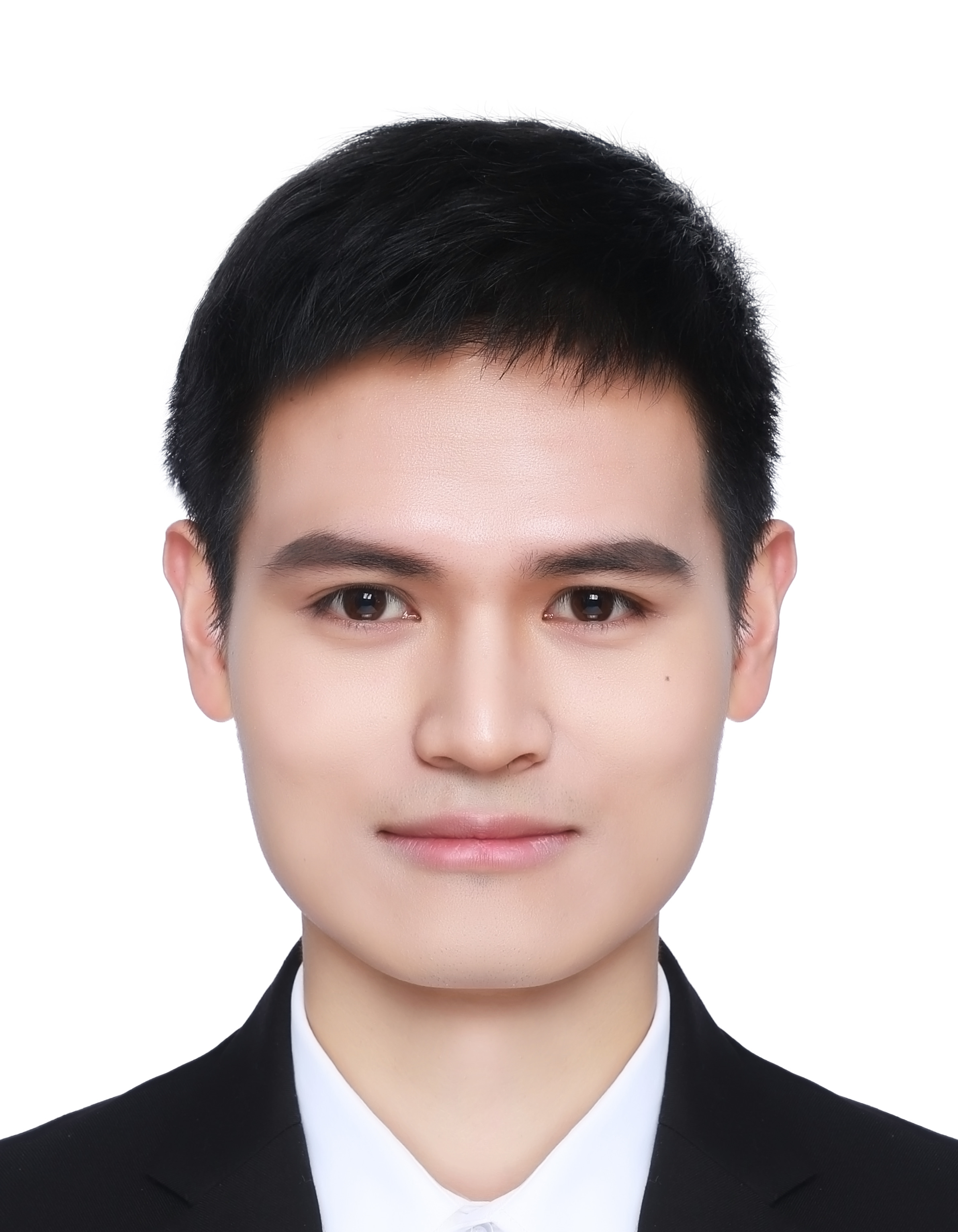}}]{Tao Lu}
Tao Lu received the BSc degree from Southwest Jiaotong University, Chengdu, China in 2016. Now he is a Ph.D. candidate at Nanjing University, supervised by Professor Gangshan Wu and Professor Limin Wang. His research interests include 3D computer vision and deep learning.
\end{IEEEbiography}

\begin{IEEEbiography}[{\includegraphics[width=1in,height=1.25in,clip,keepaspectratio]{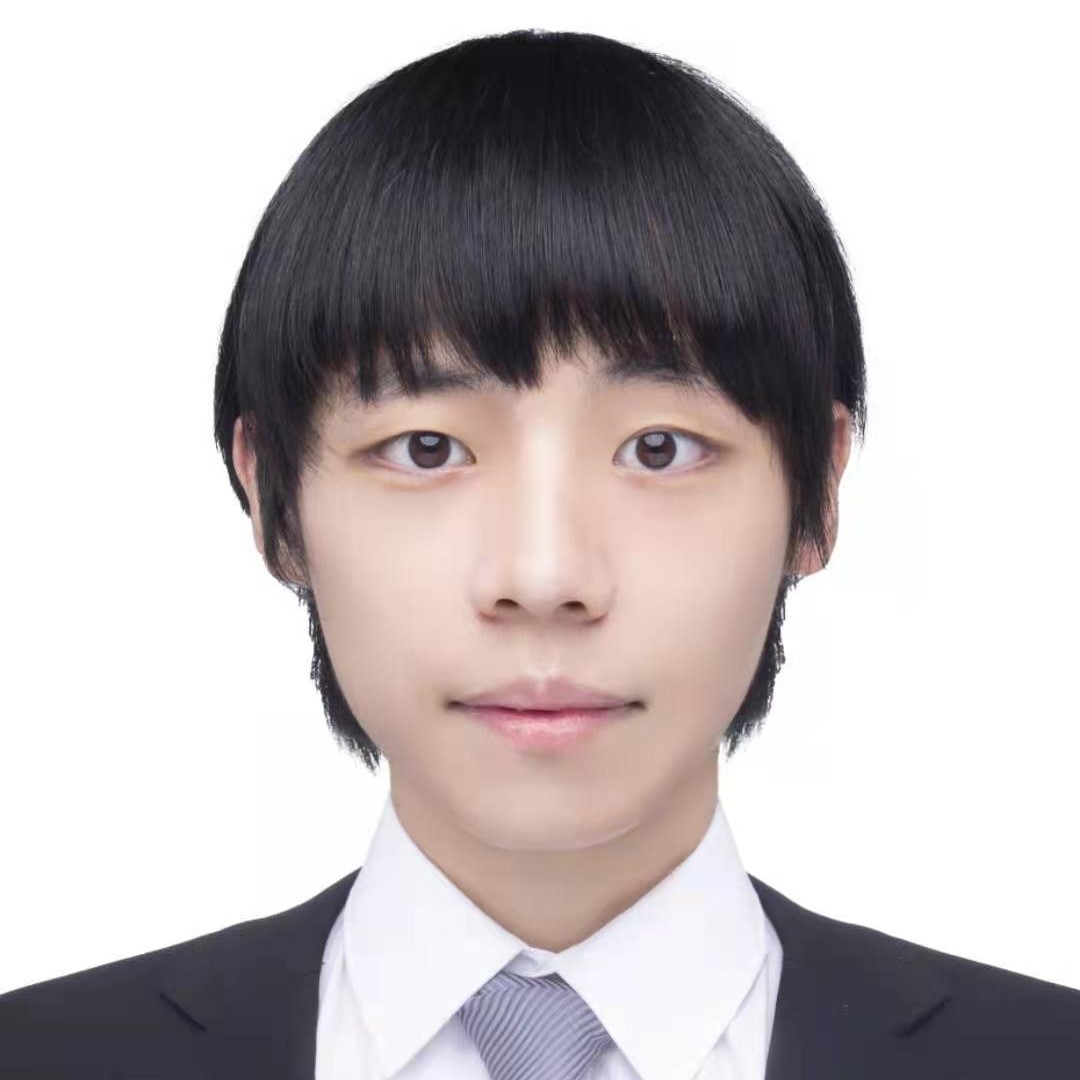}}]{Yihui Xu}
Yihui Xu received the B.E. degree in software engineering from the Northeastern University of China in 2019 and the M.E. degree in computer technology from Nanjing University in 2022, respectively. His research interests include optical flow, object tracking, big data, and data mining.
\end{IEEEbiography}

\begin{IEEEbiography}[{\includegraphics[width=1in,height=1.25in,clip,keepaspectratio]{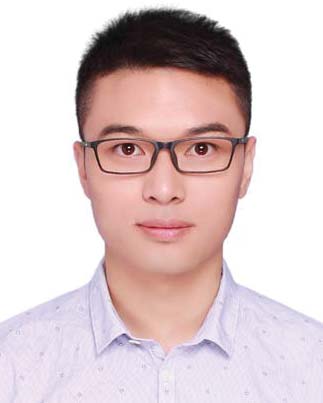}}]{Jia Liu}
Jia Liu (Member, IEEE) received the B.E. degree in software engineering from Xidian University, Xi'an, China, in 2010, and the Ph.D. degree in computer science and technology from Nanjing University, Nanjing, China, in 2016. He is currently a Research Associate Professor with the Department of Computer Science and Technology, Nanjing University. His research interest includes RFID systems. He is a member of the ACM.
\end{IEEEbiography}

\begin{IEEEbiography}[{\includegraphics[width=1in,height=1.25in,clip,keepaspectratio]{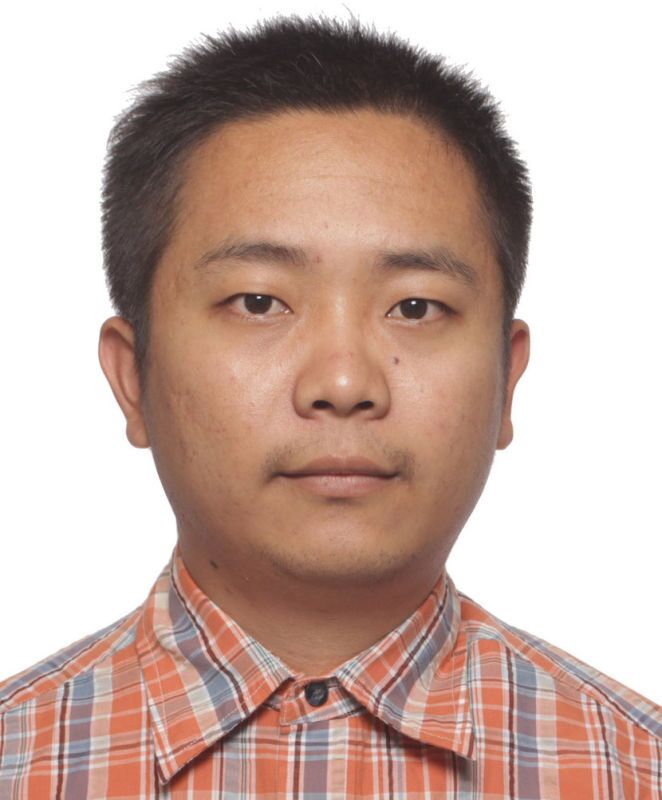}}]{Limin Wang}
Limin Wang received the B.Sc. degree from Nanjing University, Nanjing, China, in 2011, and the Ph.D. degree from the Chinese University of Hong Kong, Hong Kong, in 2015. From 2015 to 2018, he was a Post-Doctoral Researcher with the Computer Vision Laboratory, ETH Zurich. He is currently a Professor with the Department of Computer Science and Technology, Nanjing University. His research interests include computer vision and deep learning. He was the first runner-up at the ImageNet Large Scale Visual Recognition Challenge 2015 in scene recognition, and the winner at the ActivityNet Large Scale Activity Recognition Challenge 2016 in video classification. He has served as a Senior PC or Area Chair for NeurIPS, CVPR, ICCV, NeurIPS, and is on the Editorial Board of IJCV. He is a member of the IEEE.
\end{IEEEbiography}




\end{document}